\documentclass[10pt,a4paper]{article}
\usepackage[a4paper,top=2.65cm,bottom=2.65cm,left=2.15cm,right=2.15cm,headheight=19pt,headsep=14pt,footskip=26pt]{geometry}
\usepackage[utf8]{inputenc}
\usepackage[T1]{fontenc}
\usepackage{lmodern}
\usepackage{amsmath,amssymb,textcomp}
\usepackage{microtype}
\usepackage{graphicx,longtable,booktabs,array,calc,float,xcolor}
\usepackage[font=footnotesize,labelformat=empty]{caption}
\usepackage{fancyhdr}
\usepackage{titlesec}
\definecolor{biorxivnavy}{HTML}{1B365D}
\definecolor{biorxivslate}{HTML}{4A5568}
\definecolor{abstractbg}{HTML}{F7FAFC}
\definecolor{abstractrule}{HTML}{CBD5E0}
\usepackage[colorlinks=true,linkcolor=biorxivnavy,citecolor=biorxivnavy,urlcolor=biorxivnavy,pdfborder={0 0 0}]{hyperref}
\hypersetup{pdftitle={Do single-cell foundation models need 512 dimensions for one number?},pdfauthor={Lecheng Zhang, Yuhao Yang, Stan Z. Li},pdfsubject={Preprint}}
\IfFileExists{xurl.sty}{\usepackage{xurl}}{}

\titleformat{\section}{\large\sffamily\bfseries\color{biorxivnavy}}{}{0pt}{}[\vspace{2pt}\titlerule]
\titlespacing*{\section}{0pt}{14pt}{6pt}
\titleformat{\subsection}{\normalsize\sffamily\bfseries\color{black}}{}{0pt}{}
\titlespacing*{\subsection}{0pt}{10pt}{3pt}

\renewcommand{\headrulewidth}{0.3pt}
\renewcommand{\headrule}{\hbox to\headwidth{\color{abstractrule}\leaders\hrule height \headrulewidth\hfill}}
\fancypagestyle{plain}{%
  \fancyhf{}%
  \fancyfoot[C]{\footnotesize\sffamily\color{biorxivslate} \thepage}%
  \renewcommand{\headrulewidth}{0pt}%
}
\begin{document}

\thispagestyle{plain}
\begin{flushleft}
{\footnotesize\sffamily\bfseries\color{biorxivslate} PREPRINT  |  BIOINFORMATICS}\\[6pt]
{\LARGE\sffamily\bfseries\color{biorxivnavy} Do single-cell foundation models need 512 dimensions for one number?}\\[10pt]
{\normalsize\sffamily Lecheng Zhang$^{1,\dagger}$, Yuhao Yang$^{1,2,\dagger}$ and Stan Z. Li$^{1,*}$}\\[6pt]
{\footnotesize\color{biorxivslate} $^{1}$\,AI Research and Innovation Laboratory, School of Engineering, Westlake University, Hangzhou, Zhejiang 310030, China\\ $^{2}$\,SDU-ANU Joint Science College, Shandong University, Weihai, Shandong 264209, China\\[3pt]$^{\dagger}$\,These authors contributed equally to this work. \quad $^{*}$\,Correspondence: \href{mailto:stan.zq.li@westlake.edu.cn}{\texttt{stan.zq.li@westlake.edu.cn}}}
\end{flushleft}
\vspace{4pt}

\noindent\colorbox{abstractbg}{\parbox{\dimexpr\textwidth-2\fboxsep\relax}{%
\vspace{5pt}{\small\sffamily\bfseries\color{biorxivnavy} Abstract}\par\vspace{3pt}
{\small Single-cell foundation models represent counts through high-dimensional value embeddings or ranked feature tokens. We asked how many dimensions counts need and what the network computes. In a controlled Transformer, a four-dimensional value track separated from a wide gene-identity backbone retains nearly all masked-prediction net gain, and an affine value encoder replaces pretrained value curves with little transfer loss. Across cohorts, clamping detected-gene values retains most gain in annotation, surface-protein prediction and perturbation-shift consistency, consistent with prevalent single-molecule counts. A two-feature encoder combining tie-free rank and within-cell standardised magnitude improves controlled retraining and preserves annotation when grafted onto pretrained scGPT. Gene offsets and three linear cell-level scalars approximate masked prediction and predict responses to count thinning. Depth leads in the controlled model; cell type leads in scGPT. Matched controls show that attribution, attention, dictionary and probe readouts can reflect depth, gene frequency, hub degree or directions that transfer little output.\par}\vspace{5pt}}}

\vspace{6pt}

\section*{Introduction}

Single-cell foundation models are pretrained on tens of millions of cells to annotate cell types, integrate atlases and predict responses to perturbations {[}\hyperlink{ref:1}{1}--\hyperlink{ref:5}{5}{]}. Although each gene or peak contributes one scalar count, these Transformers represent each feature with hundreds of coordinates. Learned value encoders embed expression magnitudes in scGPT and scFoundation {[}\hyperlink{ref:1}{1}, \hyperlink{ref:2}{2}{]}, gene rank order conveys expression in Geneformer {[}\hyperlink{ref:3}{3}, \hyperlink{ref:4}{4}{]}, and soft-binned count embeddings are added to protein-derived gene features in State {[}\hyperlink{ref:5}{5}{]}. Shrinking an entire Transformer narrows the scalar value pathway and the gene-combining backbone together, leaving open how many dimensions the count itself needs and what the network computes from it.

Independent benchmarks have reported modest or task-dependent advantages of pretrained models over linear and task-specific baselines {[}\hyperlink{ref:6}{6}--\hyperlink{ref:9}{9}{]}, with clearer gains emerging on calibrated metrics that account for technical noise {[}\hyperlink{ref:10}{10}--\hyperlink{ref:12}{12}{]}. Explaining when value representations help requires examining the measurements they encode. Droplet transcriptomes are sparse and contain many low positive counts, making feature detection an important component of their variation {[}\hyperlink{ref:13}{13}, \hyperlink{ref:14}{14}{]}, while scGPT's within-cell quantile binning splits tied counts across bins at random. Recent studies have compared learned gene tables with protein-language-model embeddings {[}\hyperlink{ref:15}{15}{]}, explored raw-count inputs {[}\hyperlink{ref:16}{16}{]} and tested how pretraining data scale and diversity affect performance {[}\hyperlink{ref:17}{17}{]}. Yet it remains unclear how much downstream tasks rely on expression magnitudes beyond detection, or whether a compact encoder built around tie-free ranks and relative magnitudes can replace wide MLP value encoders.

Input width alone does not settle what the network computes. Even when a model predicts held-out counts accurately, prediction error does not identify the computation behind its score; distinct predictors can achieve similar scores through different computations, and in causal inference, distinct structures can also be observationally equivalent {[}\hyperlink{ref:18}{18}--\hyperlink{ref:20}{20}{]}. Inside pretrained networks, interpretability studies have probed biological coherence {[}\hyperlink{ref:21}{21}{]}, scored attention weights against regulatory databases {[}\hyperlink{ref:22}{22}{]} and decomposed activations into sparse dictionaries and manifolds {[}\hyperlink{ref:23}{23}, \hyperlink{ref:24}{24}{]}. Sequencing depth and gene prevalence structure the input data, while hub degree can dominate regulatory benchmarks. Even in hidden layers, linearly decoding a state does not ensure that predictions route through the decoded directions. Separating learned regulation from these statistical effects therefore requires functional interventions and matched baselines.

Here we trace scalar measurements from input tokens to predictions and internal readouts in pretrained transcriptomic models (scGPT, scFoundation, Geneformer-V2 and State), chromatin-accessibility models (EpiAgent and ChromFound) and controlled Transformers trained from scratch. Evaluations span human tissue transcriptomes, paired surface-protein measurements, pooled genetic perturbations and single-cell chromatin-accessibility cohorts. Separating value transport from a wide gene-identity backbone allows a narrow value track to preserve most masked-prediction net gain, while a two-feature encoder based on tie-free rank and within-cell standardised magnitude improves controlled retraining and preserves annotation during pretrained adaptation with far fewer value-encoder parameters. Across annotation, surface-protein prediction, perturbation consistency and chromatin accessibility, transfer relies mainly on which features are detected. Beyond cohort means, gene offsets plus a readout of a few linear cell-level scalars led by sequencing depth or coarse cell type approximate masked prediction and predict responses to count thinning. Matched controls show that attribution, attention, dictionary and probing readouts can reflect embedding-path geometry, hub degree, gene frequency or decodable state directions that transfer little output when swapped.

\section*{Results}

\subsection*{The expression value needs a few dimensions}

\begin{figure}[p]
\phantomsection\label{fig:1}
\centering
\includegraphics[width=\textwidth,height=0.66\textheight,keepaspectratio]{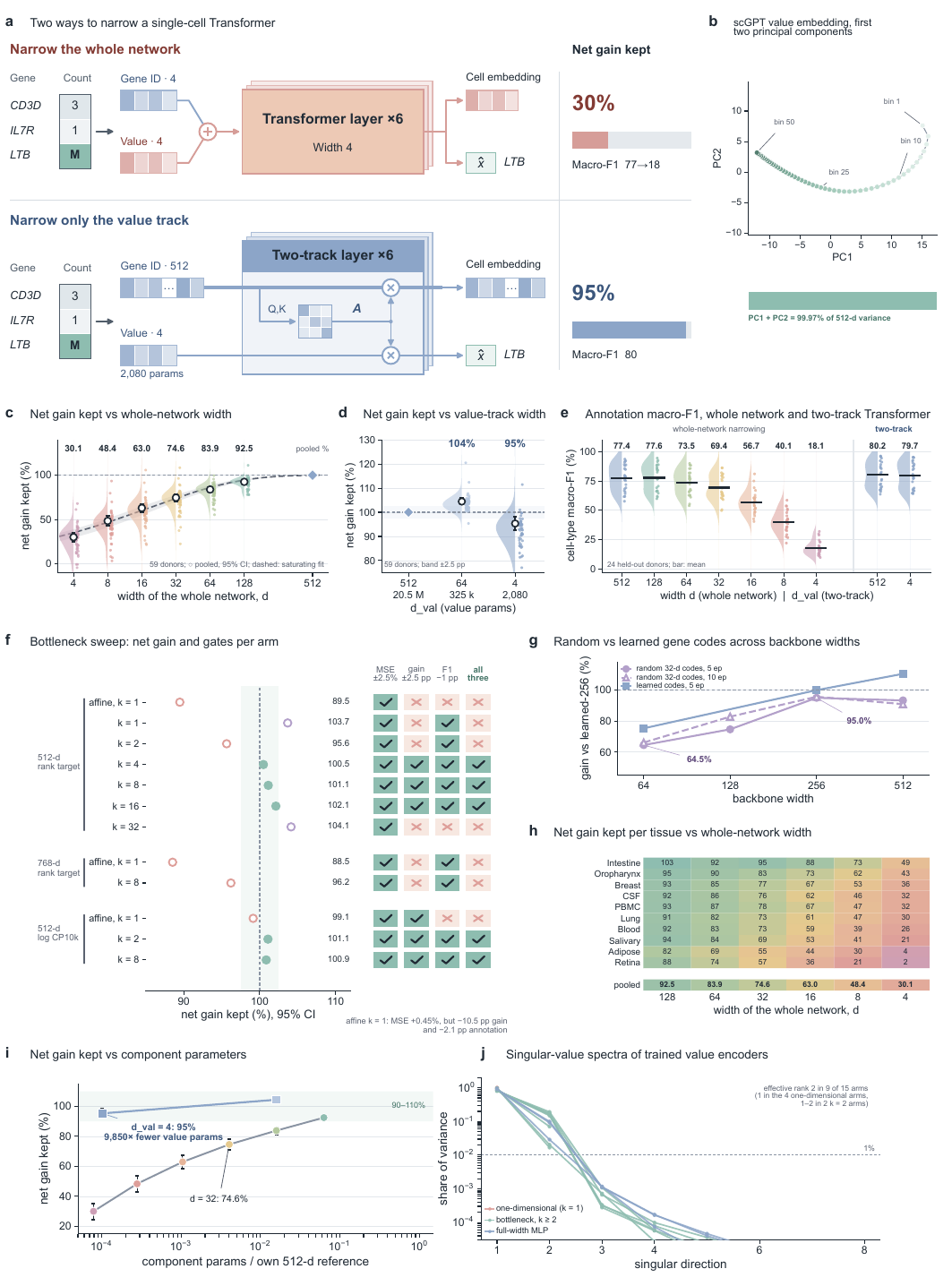}
\vspace{4pt}
\caption*{\footnotesize {\sffamily\bfseries Figure 1 \textbar{} Value-track narrowing preserves net gain; whole-network narrowing reduces it.} \textbf{a}, Schematic of whole-network and value-track narrowing in controlled Transformers (Transformers trained from scratch under controlled conditions), with colours separating gene identity, values and masked queries. \textbf{b}, Principal-component projection of pretrained scGPT value embeddings across 50 input bins, with colour marking bin rank and the bar partitioning projected and remaining variance. \textbf{c}, Masked-prediction net gain kept across whole-network widths. Dots denote 59 held-out donors averaged across three runs, rings and bars give pooled estimates with 95\% donor-bootstrap intervals, and the dashed fit carries a 95\% bootstrap band. \textbf{d}, Net gain kept under value-track narrowing on the same donors; rings and bars show pooled estimates with 95\% donor-bootstrap intervals, and shading marks $\pm$2.5 percentage points around the wide reference. \textbf{e}, Cell-type annotation macro-F1 for both architectures across 24 held-out donors, with horizontal bars marking means. \textbf{f}, Value-encoder bottleneck sweep across three runs per arm under rank-target and log counts per 10,000 (CP10k) objectives. Filled markers pass all gates; bars show 95\% bootstrap intervals, and tiles test error within $\pm$2.5\%, gain within $\pm$2.5 points and annotation no lower than $-$1 point (Methods). The one-dimensional MLP bottleneck exceeds the upper gain margin while passing the other criteria. \textbf{g}, Net gain with fixed random versus learned gene codes across backbone widths. Symbols distinguish code types and training durations, relative to learned codes at the reference width. \textbf{h}, Whole-network narrowing across ten tissue cohorts; heat-map colour and labels encode relative net gain, and the bottom row pools donors. \textbf{i}, Relative net gain across 59 held-out donors against the fraction of narrowed parameters. Circles denote whole networks and squares value tracks, with 95\% donor-bootstrap intervals and a 90--110\% reference band. Horizontal-axis denominators differ by scan, using backbone/value parameters for whole networks and value-stream parameters for two-track models, each normalised to its width-512 reference; total two-track parameters fall from 35.5 to 15.0 million at value width four. \textbf{j}, Singular-value spectra of trained value encoders across configurations. Curves show squared three-run mean singular values normalised over plotted directions on a logarithmic vertical axis, coloured by encoder type.}
\end{figure}

Narrowing the whole network and narrowing only the value path give opposite answers (\hyperref[fig:1]{Fig. 1a}). When every component of an scGPT-style Transformer {[}\hyperlink{ref:1}{1}{]} shares one width, the net gain over cohort means falls as the width shrinks, to 30\% of the 512-dimensional gain at four dimensions, and cell-type annotation from the cell embedding falls with it. The two-track Transformer separates the two roles: gene identity keeps a wide track that routes attention and forms the cell embedding, while the expression value travels through all six layers on its own narrow track. A four-dimensional value track with 2,080 parameters keeps 95\% of the net gain, and annotation changes by less than half a macro-F1 point (\hyperref[fig:1]{Fig. 1c--e}, \hyperref[edfig:3]{Extended Data Fig. 3} and Methods). Whole-network narrowing loses gain in every tissue, whereas the narrow value track keeps it with 9,850-fold fewer value-track parameters (\hyperref[fig:1]{Fig. 1h,i}). Because the identity track stays wide, the whole model shrinks from 35.5 million to 15.0 million parameters.

Pretrained value encoders already use few directions. Two principal components carry more than 99.9\% of the variance of scGPT's learned value curve and 97\% of scFoundation's {[}\hyperlink{ref:1}{1}, \hyperlink{ref:2}{2}{]} (\hyperref[fig:1]{Fig. 1b} and Supplementary Table 4), and replacing either curve with a straight line, an affine value encoder, changes downstream transfer little. The exception is masked prediction by scFoundation on K562 cells, where the affine value encoder retains 44.76\% of the masked-prediction gain (Supplementary Table 4). Chromatin-accessibility models that encode counts as values show the same geometry: ChromFound's value embedding has an effective rank of one, whereas EpiAgent, which encodes accessibility through peak ranks, has a high-dimensional rank embedding (\hyperref[edfig:2]{Extended Data Fig. 2k}).

Training from scratch gives the same answer. Across ten cohorts evaluated on held-out donors (\hyperref[edfig:1]{Extended Data Fig. 1} and Supplementary Table 1), an affine value encoder matches an MLP value encoder on masked prediction, annotation and surface-protein prediction, also at one million training cells (\hyperref[edfig:2]{Extended Data Fig. 2c,d}). Free to use many directions, the MLP value encoder collapses to an effective rank of two after training (\hyperref[fig:1]{Fig. 1j}), so the network keeps little of its nonlinearity.

A bottleneck sweep identifies which widths suffice (\hyperref[fig:1]{Fig. 1f}). In an scGPT-sized rank-target Transformer, bottlenecks of 4, 8 and 16 dimensions meet every pre-specified criterion on masked-prediction error, net gain and downstream probes, and a one-dimensional affine value encoder keeps 90\% of the net gain. A one-dimensional MLP bottleneck keeps 104\% of the full-width gain, above the upper margin of the two-sided gain criterion (Methods and Supplementary Table 4).

Width is needed where genes are combined. With fixed random gene codes, a 256-dimensional Transformer recovers 95\% of the net gain of learned codes, narrower backbones recover less, and learned codes pull ahead only at the widest setting (\hyperref[fig:1]{Fig. 1g}). The count itself needs a few dimensions; the backbone that combines genes needs many.

\subsection*{Downstream tasks read which genes are detected}

\begin{figure}[p]
\phantomsection\label{fig:2}
\centering
\includegraphics[width=\textwidth,height=0.66\textheight,keepaspectratio]{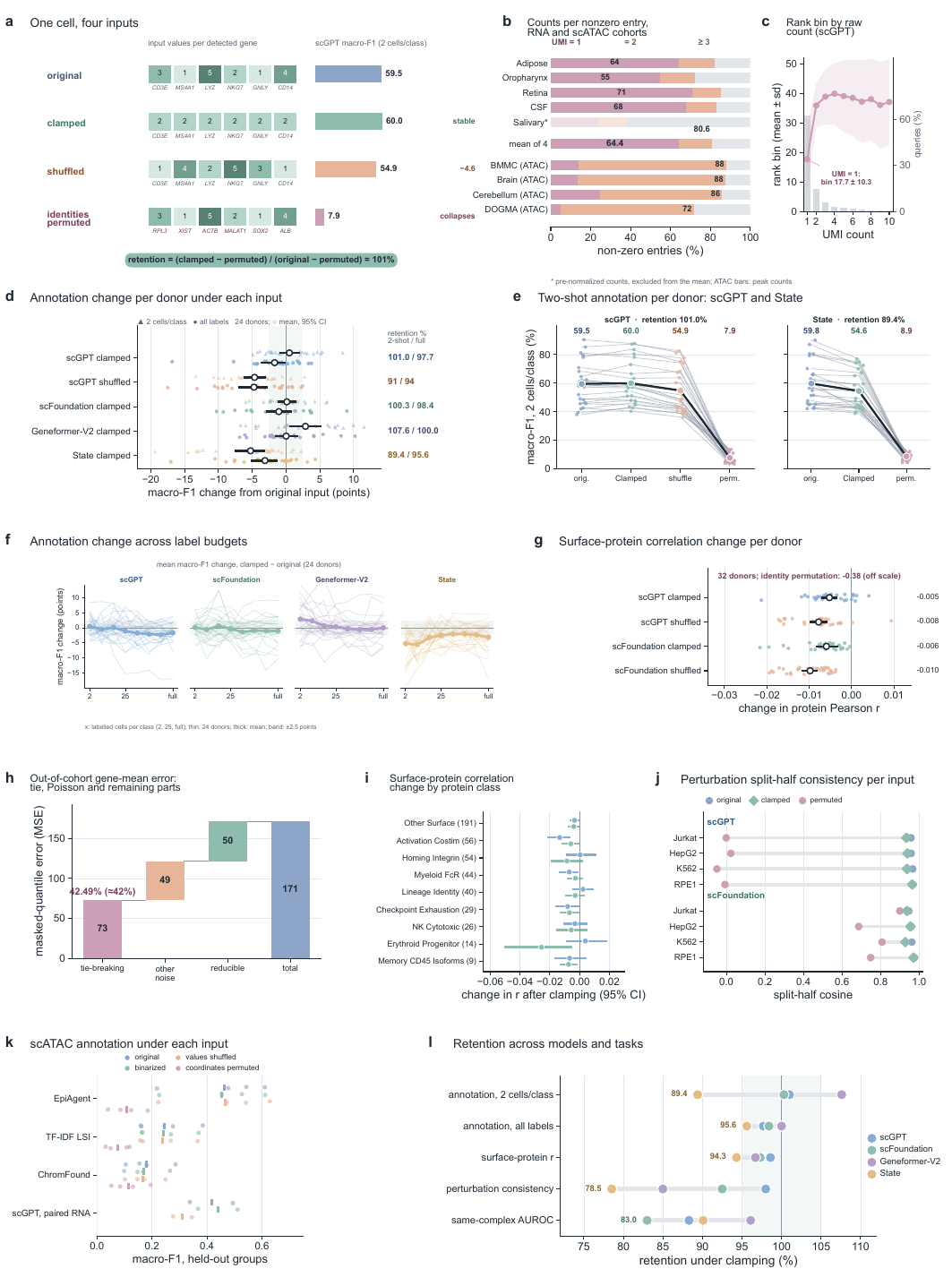}
\vspace{4pt}
\caption*{\footnotesize {\sffamily\bfseries Figure 2 \textbar{} Gene detection carries most of the downstream gain.} \textbf{a}, Schematic of original, clamped, shuffled-value and identity-permuted inputs, with colours distinguishing interventions and gene labels identifying detected transcripts. \textbf{b}, Nonzero count distributions across tissue and chromatin-accessibility cohorts. Stacks separate one, two and higher counts for RNA unique molecular identifiers (UMIs) and accessibility peaks; faded salivary counts are excluded from the RNA mean. \textbf{c}, Within-cell rank bins by UMI count, with the curve and band giving mean $\pm$ standard deviation and bars giving query shares. \textbf{d}, Donor-level cell-type annotation changes across 24 held-out donors. Triangles denote two labelled cells per class (two-shot), circles all labels, and rings means with 95\% donor-bootstrap intervals. \textbf{e}, Two-shot annotation across 24 held-out donors for pretrained scGPT and State; thin lines connect donors and thick lines connect means. \textbf{f}, Clamped-input annotation changes across label budgets. Thin lines denote donors, thick lines means, and shading $\pm$2.5 points. \textbf{g}, Surface-protein correlation changes under clamping and value shuffling with cognate transcripts masked across 32 donors (dots), alongside means and 95\% donor-bootstrap intervals (rings). \textbf{h}, Out-of-cohort gene-mean error across five cohorts. Waterfall bars separate tie variance, additional Poisson-model noise and remaining error (Methods). \textbf{i}, Clamping-induced correlation changes by protein class across 463 surface proteins in 32 held-out donors. Points show mean changes with 95\% protein-bootstrap intervals. \textbf{j}, Perturbation-response split-half consistency under each input across 320 CRISPRi perturbations in four cell lines, grouped by cell line and model. \textbf{k}, Chromatin-accessibility annotation for EpiAgent, ChromFound, term frequency--inverse document frequency (TF-IDF) with latent semantic indexing (LSI), and paired-RNA scGPT. Dots mark cohorts and ticks mark means; accessibility probes remain fixed after intact-input fitting, grouped by donor in three cohorts and by library in DOGMA. \textbf{l}, Retention across models and tasks. Colour identifies models, shading marks 95--105\% retention, and complex recovery uses area under the receiver operating characteristic curve (AUROC).}
\end{figure}

To compare tasks on one scale, we gave each pretrained model four versions of every cell: the original values, a clamped input with one constant value on every detected gene, values shuffled among the detected genes, and permuted gene identities as the lower reference (\hyperref[fig:2]{Fig. 2a}). Retention is the gain of an altered input above the permuted reference, as a share of the gain of the original input (Methods). Probes were refitted after each intervention. Across four models and five tasks, the three models that read expression through detected genes keep most of their gain under clamping (\hyperref[fig:2]{Fig. 2l}); scFoundation keeps its supplied depth tokens (Methods).

Cell-type annotation keeps its gain without the values. Across 24 held-out donors from five cohorts, clamped inputs retain 98--101\% of the gain in scGPT and scFoundation, from two labelled cells per class to full labels, and Geneformer-V2 retains at least as much. Donors vary: in scGPT, changes range from $-$16.8 to +8.4 macro-F1 points, and full-label mean annotation falls by 1.7 points. Shuffling values costs scGPT about five macro-F1 points, so the original encoder uses magnitudes even though refitted probes recover most average transfer from clamped inputs. By contrast, in State, which encodes magnitudes more directly as a positive control, clamping lowers its two-shot retention to 89\% (\hyperref[fig:2]{Fig. 2d--f}).

Surface-protein prediction and the consistency of perturbation-associated embedding shifts show similar retention. When 463 proteins are predicted from expression with their cognate, paralogous and subunit transcripts masked, clamped inputs retain 97--99\% of the gain in scGPT, scFoundation and Geneformer-V2, and the average effect remains small across protein classes (\hyperref[fig:2]{Fig. 2g,i}). Across 320 CRISPRi perturbations in four cell lines, clamped inputs keep the split-half consistency of each perturbation-associated embedding shift, whereas permuting gene identities removes it (\hyperref[fig:2]{Fig. 2j}). Recall of protein complexes keeps less of its gain, which shows that magnitudes contribute to complex structure. Clamping gaps stay small after partial fine-tuning (Supplementary Table 5; protein comparisons in \hyperref[edfig:4]{Extended Data Fig. 4c}). After fine-tuning, annotation changes by $-$0.61 and +1.23 macro-F1 points at full labels in scGPT and scFoundation, respectively, and falls by 1.79 and 1.94 points with ten labels per class.

Chromatin-accessibility models also read which features are present. In four scATAC-seq cohorts, binarising peak counts or shuffling nonzero values within a cell changes EpiAgent annotation by at most 4.3 macro-F1 points, whereas permuting peak coordinates removes about 70\% of its score (\hyperref[fig:2]{Fig. 2k}). TF-IDF with latent semantic indexing behaves the same way. Applied to the paired RNA of two of these cohorts, scGPT does lose annotation when its values are shuffled, so the same assay detects magnitude use where it exists.

The counts themselves explain why detection carries so much. In the four held-out tissue cohorts with raw counts, about two thirds of nonzero entries are single molecules, most nonzero scATAC entries are one or two, and detection alone accounts for 83\% of the per-gene variance of log-normalised expression. Under scGPT's within-cell quantile binning, genes with equal counts are split across bins at random, so a single molecule can land in almost any bin. This tie-breaking accounts for 42\% of the error of an out-of-cohort gene-mean predictor (\hyperref[fig:2]{Fig. 2b,c,h}). In scFoundation, raw predictions improve on out-of-cohort gene means only when zeros are among the queries, and calibrating detection rates and visible depth recovers the gain on positive-only queries (\hyperref[edfig:4]{Extended Data Fig. 4b}).

Tie noise and depth motivate an encoder that represents tie-free rank and relative magnitude explicitly. The two-feature encoder combines a tie-free mid-rank with a within-cell standardised log magnitude in 1,280 parameters, against 66,816 for the MLP value encoder. Trained on ten cohorts and scored on 59 held-out donors, it lowers masked-prediction error against the quantile encoder in every donor and raises three-run mean annotation from 0.66 to 0.77 macro-F1 (0.68 to 0.79 for the three-run ensemble). Either tie-free feature alone captures most of the error reduction and annotation gain, whereas reintroducing raw depth into the classification token degrades both. Grafted onto pretrained scGPT, the two-feature encoder replaces 264,704 value parameters with 2,560 and keeps annotation at the level of the original encoder (\hyperref[edfig:2]{Extended Data Fig. 2f,g}); its advantage persists in a 12-layer Transformer trained on one million cells (Supplementary Table 4).

\subsection*{A few cell-level scalars approximate much of masked prediction}

\begin{figure}[p]
\phantomsection\label{fig:3}
\centering
\includegraphics[width=\textwidth,height=0.66\textheight,keepaspectratio]{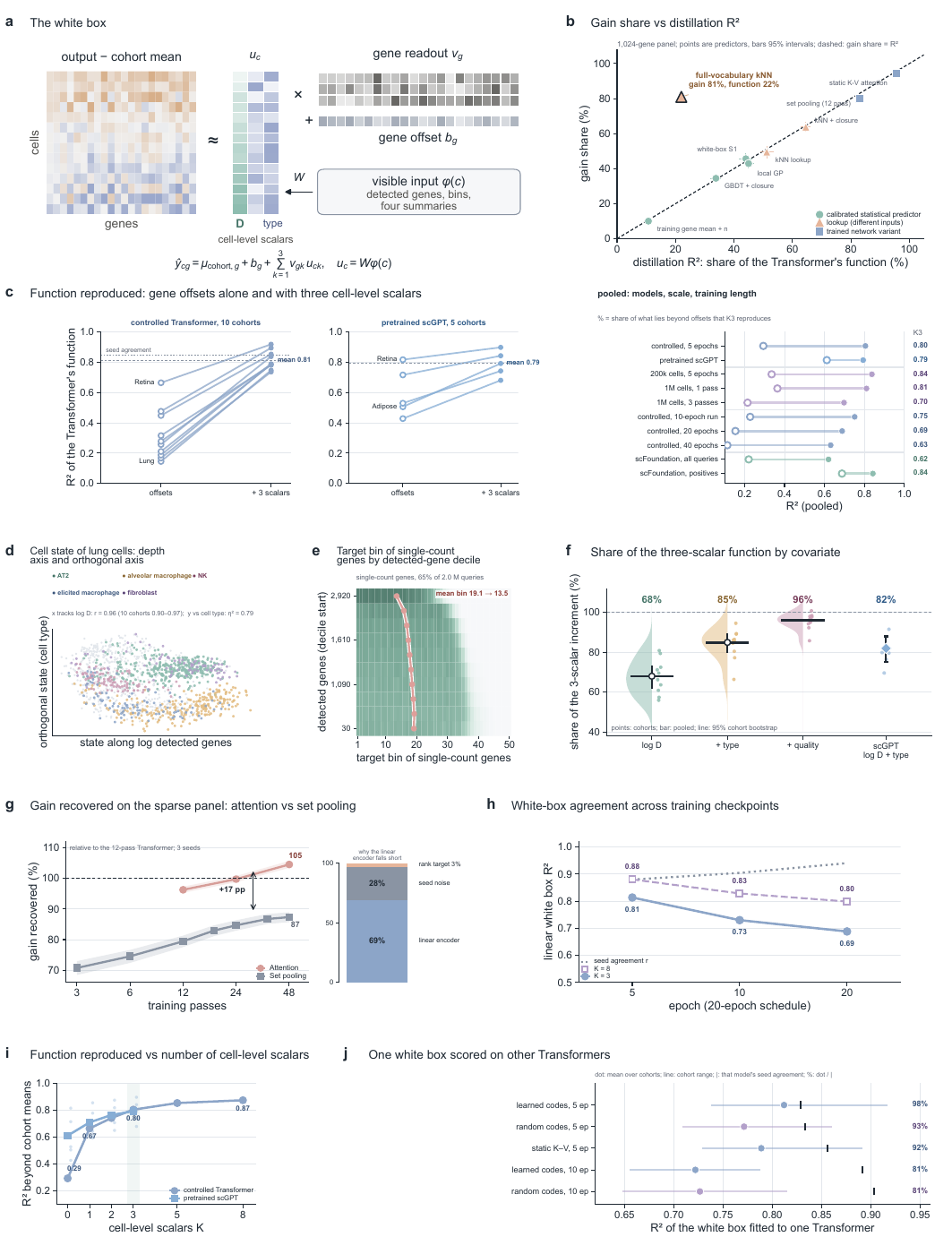}
\vspace{4pt}
\caption*{\footnotesize {\sffamily\bfseries Figure 3 \textbar{} Gene offsets and a few cell-level scalars approximate much of masked prediction.} \textbf{a}, Schematic of the white box comprising cohort means, gene offsets and a readout of cell-level scalars; coloured matrices separate cell state from gene readouts, and the arrow marks the visible-input map (Methods). \textbf{b}, Gain share versus distillation agreement across predictor families, with 95\% run-by-cell bootstrap intervals and the equality diagonal. \textbf{c}, Function reproduced by gene offsets alone and with three cell-level scalars across ten cohorts (five cohorts for pretrained scGPT). Lines join each cohort's open and filled symbols, dashed lines mark cohort means, dotted lines mark independent-model agreement, and right-hand dumbbells show pooled summaries. \textbf{d}, Lung cells in depth-aligned cell-state coordinates, coloured by cell type with the second axis orthogonal to depth. \textbf{e}, Single-count target bins across detected-gene-count deciles. Heat-map intensity shows frequency, and the overlaid curve traces mean bins. \textbf{f}, Share of the three-scalar function explained by depth, cell type and quality covariates across ten cohorts (five cohorts for pretrained scGPT). Dots denote cohorts; summary markers give pooled shares with 95\% cohort-bootstrap intervals. \textbf{g}, Gain recovered on the sparse panel by attention versus set pooling over training. Bands show 95\% cell-bootstrap intervals across three runs, and the inset partitions the sparse-context shortfall among encoder, model-noise and rank terms. \textbf{h}, White-box agreement across training checkpoints by number of cell-level scalars, relative to independent-model agreement (dotted line). \textbf{i}, Function agreement against the number of cell-level scalars across model families; small points mark scGPT cohorts, and shading marks the reference dimension. \textbf{j}, Cross-model evaluation across ten cohorts of a white box fitted to one Transformer. Dots show cohort means, bars cohort ranges, and ticks each model's independent-run agreement.}
\end{figure}

Although downstream transfer reads mainly gene detection, masked prediction still uses visible context to predict held-out expression, and what it computes can be measured directly. For calibrated predictors with nested information, the share of the Transformer's function a predictor reproduces, its distillation $R^2$, equals the share of the gain it recovers. The controlled Transformers are calibrated in this sense, and predictors built from the sparse panel's own statistics sit on this diagonal. Nearest-neighbour retrieval on the full vocabulary sits far below that diagonal, as it recovers 81\% of the gain while reproducing only 22\% of the function, so a high score can come from a different computation (\hyperref[fig:3]{Fig. 3b}). We therefore describe what the Transformer computes by measuring agreement with its own outputs alongside agreement with the targets.

Beyond cohort means, we fit the following white-box approximation to the controlled Transformer (\hyperref[fig:3]{Fig. 3a}):

\[
T_{cg} \approx \hat y_{cg} = \mu_{\text{cohort},g} + b_g + \sum_{k=1}^{3} v_{gk}\,u_{ck}, \qquad u_c = W\phi(c).
\]

Each gene has an offset $b_g$ and a readout $v_g$ of a cell state $u_c$ with three cell-level scalars, and the cell state is a linear map of the visible input $\phi(c)$: which genes are detected, their bins and four summary statistics. On 59 held-out donors, the offsets alone reproduce 0.29 of the Transformer's function beyond cohort means, and the three cell-level scalars raise this to 0.80, which is 95\% of the agreement between independently trained runs (\hyperref[fig:3]{Fig. 3c,i}). The same white box also describes Transformers trained with random gene codes, static key--value attention or longer schedules (\hyperref[fig:3]{Fig. 3j}). In pretrained scGPT it reaches a similar agreement on five held-out cohorts, but the gene offsets carry more of the function, and the cell-level scalars describe 47\% of what lies beyond them, against 76\% in the controlled Transformer on the same cohorts.

The leading cell-level scalar is sequencing depth, measured by the detected-gene count $D$. In the controlled Transformer, $\log D$ alone accounts for 68\% of what the three scalars add beyond the offsets, as much as total UMI, mitochondrial and ribosomal fractions combined, and coarse cell type raises the share to 85\% (\hyperref[fig:3]{Fig. 3d,f}). In pretrained scGPT, cell type and $D$ together account for a similar share, with cell type leading. Depth therefore sets much of what the model predicts for every masked gene of a cell, and the model needs no explicit depth input to read it. Although two thirds of its training cells arrive as a random window of 1,024 detected genes, the visible window predicts the whole-cell $\log D$ of held-out donors almost exactly (\hyperref[edfig:6]{Extended Data Fig. 6b}).

Under rank targets, the dependence on $D$ follows from ranking itself. Each target ranks a gene among the genes its cell detected, so a gene's expected bin moves with the number of genes that compete with it. A parameter-free multinomial model that uses only average gene abundances and the binning rule predicts how strongly each gene responds to $D$ (\hyperref[fig:3]{Fig. 3e} and \hyperref[edfig:6]{Extended Data Fig. 6c}). Ranking targets within the input window instead of the whole cell leaves the share of $D$ unchanged. Along the tested gene readouts, none of 1,801 Hallmark, Reactome and curated gene sets shows pathway enrichment after controlling for abundance and detection rate (\hyperref[edfig:6]{Extended Data Fig. 6e}).

Longer training adds cell-level scalars without replacing the leading ones. Training on five times more cells at the same number of optimisation steps leaves the three-scalar description almost unchanged, with $D$ still leading. Three passes over the larger corpus raise the net gain by 70\%, and the function then needs more cell-level scalars. Three scalars reproduce 0.70 of it and eight reproduce 0.79, the same values a free per-cell state reaches (\hyperref[fig:3]{Fig. 3h} and \hyperref[edfig:6]{Extended Data Figs. 6d} and 7a). Additional optimisation steps increase the dimensionality that the white box needs more than adding cells at fixed steps. This step-driven growth complements reports of rank growth in network weights and embeddings {[}\hyperlink{ref:25}{25}--\hyperlink{ref:28}{28}{]}.

Pairwise attention earns its cost when context is sparse. On a panel of 1,024 genes with about 18 visible genes per cell, attention outperforms an architecture-matched set model by 17 percentage points of gain recovered. With so few visible genes the cell state stays low-dimensional, but a linear encoder no longer reaches it, and most of that shortfall comes from the encoder itself (\hyperref[fig:3]{Fig. 3g} and \hyperref[edfig:6]{Extended Data Figs. 6a} and 7d).

\subsection*{Sequencing depth leads the cell state and predicts the response to thinning}

\begin{figure}[p]
\phantomsection\label{fig:4}
\centering
\includegraphics[width=\textwidth,height=0.66\textheight,keepaspectratio]{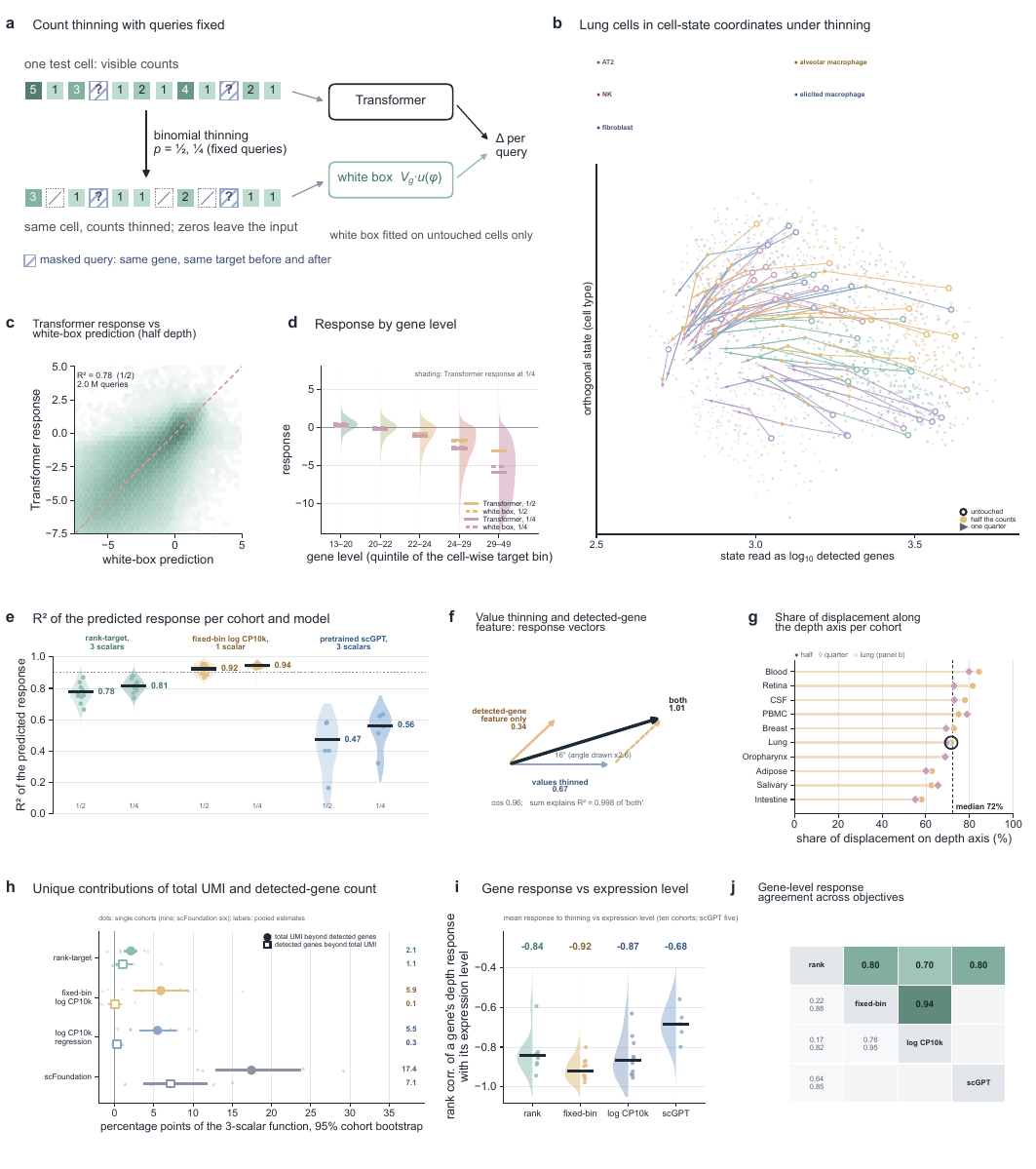}
\vspace{4pt}
\caption*{\footnotesize {\sffamily\bfseries Figure 4 \textbar{} The white box predicts responses to count thinning along the depth axis.} \textbf{a}, Schematic of count thinning, binomial thinning of visible counts to one half or one quarter with queries and targets fixed. Crossed boxes denote lost detections, and arrows trace rebuilt inputs to model and white-box responses (Methods). \textbf{b}, Lung-cell trajectories from untouched to thinned cell states for ten cells per type; colours mark cell types, rings untouched cells, yellow dots half-depth states and arrowheads quarter-depth states. \textbf{c}, Observed Transformer responses versus white-box predictions at half depth across 2.0 million queries across ten cohorts, with logarithmic hexagon density and the dashed equality diagonal. \textbf{d}, Query responses by target-bin quintile. Shading shows quarter-depth distributions, while solid and dashed ticks mark model and white-box means at each dose. \textbf{e}, Thinning-response agreement across ten cohorts (five cohorts for pretrained scGPT) for rank-target, fixed-bin log counts per 10,000 (CP10k) and pretrained scGPT models. Dots and densities show cohort distributions; black ticks mark pooled uncentred agreement. \textbf{f}, Response vectors for separate value thinning, detected-gene-feature change and their sum alongside the joint response, with vertical separation exaggerated as labelled. \textbf{g}, Share of cell-state displacement along the depth direction across ten cohorts for half (circles) and quarter (diamonds) depth; the dashed line marks the half-depth median and the ring marks lung. \textbf{h}, Unique depth-covariate shares across nine raw-count cohorts by objective and input. Circles give total unique molecular identifiers (UMIs) beyond detected-gene count, open squares the converse, and bars 95\% cohort-bootstrap intervals. \textbf{i}, Correlation of per-gene thinning responses with expression level across ten cohorts by objective (dots, cohorts; black ticks, medians). \textbf{j}, Pairwise gene-response correlations across objectives and pretrained scGPT over ten cohorts (five for scGPT). Upper-triangle shading gives cohort medians; lower entries list cohort minima and maxima.}
\end{figure}

Thinning visible counts binomially lowers a cell's depth, its leading cell-level scalar, while the cell, its masked queries and their targets stay fixed (\hyperref[fig:4]{Fig. 4a}). On held-out test cells thinned to one half or one quarter of their visible counts, we asked the white box, fitted on untouched training cells, to predict how the Transformer's outputs change. At half depth, thinning moves the rank-target Transformer's predictions by about three quarters of the spread of its whole contextual function. Across 2.0 million masked queries in all ten cohorts, the white box predicts these changes with $R^2 = 0.78$ at half depth and 0.81 at quarter depth (\hyperref[fig:4]{Fig. 4c,e}).

Thinning moves cells along the depth axis of the cell state. That axis carries 72\% of the squared displacement in the median cohort, and cells of every type move the same way. Genes respond according to their expression level, with lowly expressed genes rising and highly expressed genes falling. The white box follows both the ordering and the size of the mean response (\hyperref[fig:4]{Fig. 4b,d,g}).

Which depth statistic leads depends on the training objective and input normalisation (\hyperref[fig:4]{Fig. 4h}). The fixed-bin Transformer trains on global bins of log CP10k, the logarithm of one plus counts per 10,000. In this fixed-bin model, one cell-level scalar tracking total UMI carries 99\% of the white-box gain, against 75\% under rank targets, and a one-scalar white box predicts its response to thinning with $R^2$ above 0.9. Total UMI leads because full-cell totals normalise the fixed-bin inputs, whereas rebuilding the training inputs from visible genes alone makes detected-gene count lead just as closely. Under rank targets the two statistics are interchangeable, and in the one cohort whose counts were pre-normalised to a near-constant total, the function follows detected-gene count (\hyperref[edfig:8]{Extended Data Fig. 8a,b}). scFoundation orders the depth covariates in the same way, with cell type carrying as much. By contrast, the white box predicts pretrained scGPT's response to thinning less completely ($R^2=0.47$ and 0.56); 57--63\% of the response lies within the span of its gene readouts (\hyperref[fig:4]{Fig. 4e}).

Changing the values and changing a supplied depth feature move predictions along one direction. The log CP10k regression Transformer receives depth twice, through the visible values and through a detected-gene feature in its classification token. Thinning the values with that feature held fixed and changing the feature with the values held fixed give nearly parallel responses (uncentred cosine 0.96), and their sum predicts the joint change almost exactly (\hyperref[fig:4]{Fig. 4f}). Across objectives and in scGPT, a gene's response to thinning tracks how lowly it is expressed, and gene-level responses agree between objectives and with scGPT, even though the mean response changes sign between rank and log CP10k targets (\hyperref[fig:4]{Fig. 4i,j} and \hyperref[edfig:8]{Extended Data Fig. 8c}).

Both targets make depth the best single summary of a cell. Library-size normalisation places every single-count gene at $\log(1 + 10^4/\mathrm{UMI})$, and ranking moves a gene's target with the number of genes that compete with it. Either way, the best prediction of a masked gene depends on the cell's depth, which the model reads from its visible genes and from any supplied depth feature, and coarse lineage supplies most of the rest of the cell state.

\subsection*{Interpreting readouts requires matched functional controls}

\begin{figure}[p]
\phantomsection\label{fig:5}
\centering
\includegraphics[width=\textwidth,height=0.66\textheight,keepaspectratio]{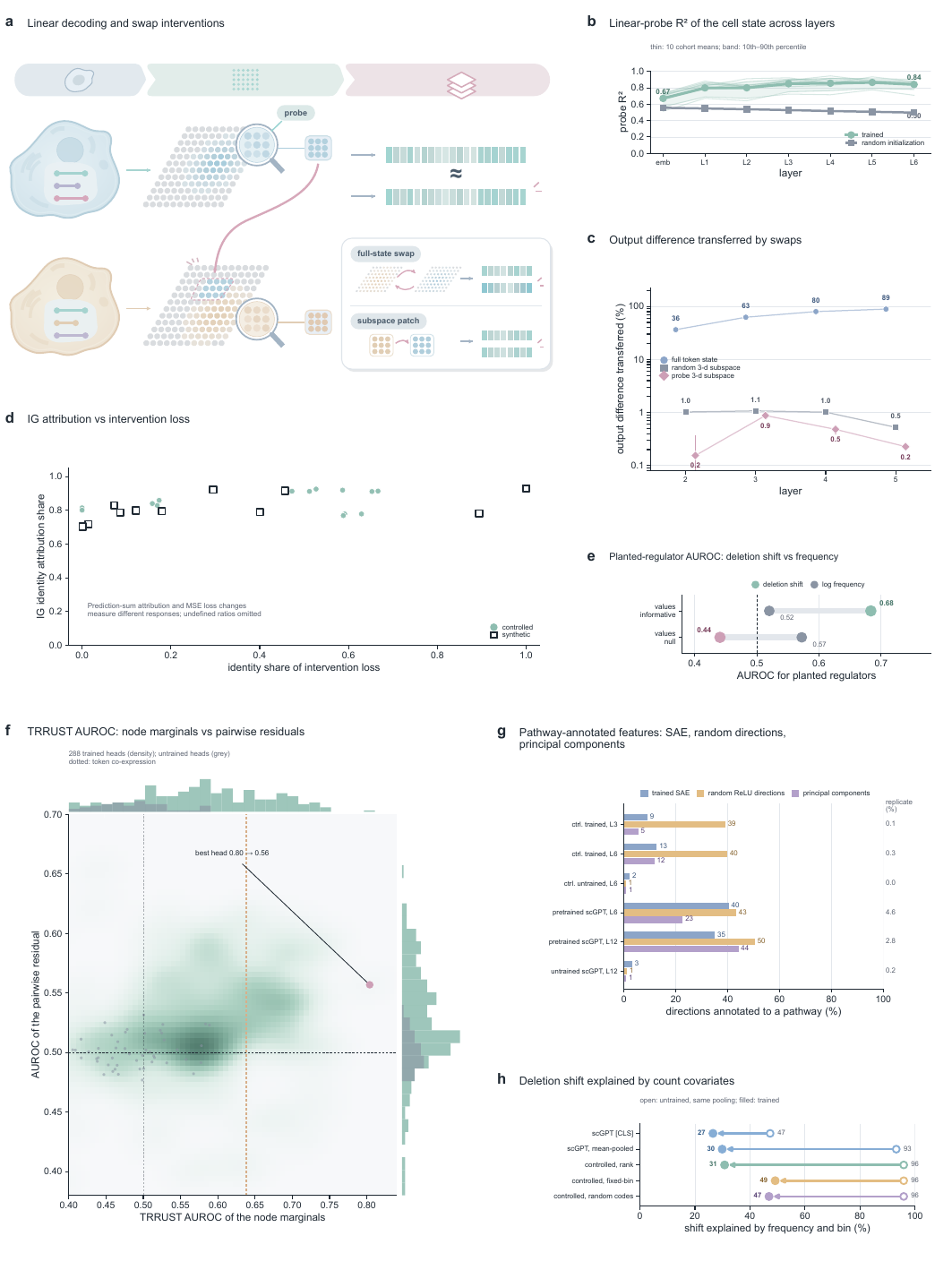}
\vspace{4pt}
\caption*{\footnotesize {\sffamily\bfseries Figure 5 \textbar{} Matched controls distinguish readable structure from intervention effects.} \textbf{a}, Schematic of linear decoding and swap interventions on the same layer state, separating covariate readout from coordinate replacement along probe axes. \textbf{b}, Linear-probe decoding of cell state across layers in trained and randomly initialised Transformers. Thin lines trace ten cohort means across runs, thick lines show stored macro summaries, and shading spans the tenth-to-ninetieth cohort percentiles. \textbf{c}, Subspace swap transfer across ten cohorts. Symbols mark full-token, random-subspace and probe-subspace swaps on a logarithmic axis with visible portions of 95\% bootstrap intervals (Methods). \textbf{d}, Integrated Gradients (IG) identity attribution versus intervention-based identity loss share. Circles denote controlled-model runs and open squares synthetic runs, each averaged over available cohorts. Horizontal positions give the positive-part MSE increase from within-cell identity permutation divided by its sum with the increase from clamped input; ratios with a zero denominator are omitted. IG attributes the sum of masked predictions along embedding paths. \textbf{e}, Recovery of 40 planted transcription factors across two runs by deletion shift versus gene frequency, scored by area under the receiver operating characteristic curve (AUROC) under value-informative and value-null conditions with connectors pairing readouts. \textbf{f}, Attention-head node marginals and pairwise residuals scored against TRRUST, a curated database of human transcriptional regulatory interactions. Density covers 288 trained heads, grey dots mark untrained heads, marginal histograms show their distributions, dashed lines mark chance, and the dotted line marks token co-expression. \textbf{g}, Pathway-annotated feature shares for sparse autoencoder (SAE) atoms (mean of two fits), random rectified directions and principal components; right-hand labels report cross-fit atom reproducibility above cosine similarity 0.9. \textbf{h}, Variance in deletion-induced embedding shifts explained by gene frequency, expression bin and visible-gene count across 19,200 controlled-model and 4,000 scGPT deletions. Open and filled circles distinguish untrained and trained models, and arrows pair matched pooling conditions.}
\end{figure}

If masked prediction is a readout of a few cell-level scalars led by depth, the representations that interpretability methods read should carry the same statistics. We applied six common readouts to pretrained scGPT, to the controlled Transformers, to control models in which one input channel carries no information by construction, and to a 100,000-cell synthetic benchmark with 12 planted modules and regulators, each against a matched control. The cell state shows the distinction most directly. A linear probe decodes it from every layer in every cohort ($R^2 = 0.87$ at layer 5), yet swapping it between cells along the probe directions transfers 0.23\% of the output difference, less than swapping a random three-dimensional subspace (\hyperref[fig:5]{Fig. 5a--c}). Learned subspaces can transfer more than the full state at late layers, but randomly initialised networks show the same effect (\hyperref[edfig:9]{Extended Data Fig. 9b,i}).

Cell embeddings carry depth. A polynomial in detected-gene count and total UMI explains about half of the embedding variance of the rank-target Transformer and most of it under fixed bins. A principal-component projection of the binary detection pattern, which uses no network at all, annotates cells better than these embeddings, and untrained Transformers match a random projection of the same pattern (\hyperref[edfig:9]{Extended Data Fig. 9a,d}). Pretrained scGPT is less dominated by depth and annotates better than the detection projection on the same cohorts. In the controlled models, annotation largely reflects summaries of gene detection and sequencing depth.

Integrated Gradients (IG) measures path-dependent attribution in the embedding space. For identity and value vectors added at the input, the path gradient is shared and IG multiplies it by each vector's displacement from the baseline. Controls that randomise gene identities on every forward pass still receive 75--95\% identity attribution. Across controlled and synthetic models, these attribution shares also differ from the relative loss increases caused by within-cell identity permutation and value clamping (\hyperref[fig:5]{Fig. 5d} and \hyperref[edfig:9]{Extended Data Fig. 9e}). The two quantities measure different objects: attribution to a sum of masked predictions along an embedding path versus changes in prediction error after discrete input interventions. IG channel shares therefore track embedding-path geometry, and a channel that carries no information can receive most of the attribution.

Attention benchmarks reward hub genes. When each head's attention is decomposed into node marginals and a pairwise residual, single-head marginals separate 715 TRRUST regulator--target pairs from frequency-matched negatives with an AUROC of 0.80, whereas the pairwise residual stays near chance (\hyperref[fig:5]{Fig. 5f}). Transformers with random gene codes and token co-expression match head-averaged attention, and the two genes' TRRUST degrees alone separate the pairs almost perfectly. Pathway benchmarks similarly reward annotation degree (\hyperref[edfig:9]{Extended Data Fig. 9c,h}). On 14,734 CRISPRi-tested enhancer--promoter pairs, inverse distance reaches an AUROC of 0.88, as high as the token similarity of a chromatin-accessibility model, and both fall to chance once negatives are matched for distance (Supplementary Table 3).

Deleting a gene moves a cell according to the gene's frequency. Gene frequency, expression bin and the number of visible genes explain almost all of the embedding shift caused by a deletion in untrained networks, and a quarter to a half after training. Deletion still carries signal beyond frequency on the synthetic benchmark, where it ranks planted regulators above frequency alone when expression values carry information and below it when they do not (\hyperref[fig:5]{Fig. 5e,h}).

Dictionary and gene-table readouts are matched by simple baselines. Random rectified directions enrich Gene Ontology and Reactome terms as often as trained TopK sparse autoencoders, or more often, and fewer than 5\% of autoencoder atoms reproduce across fits (\hyperref[fig:5]{Fig. 5g} and \hyperref[edfig:9]{Extended Data Fig. 9f}). The gene tables of the controlled Transformers stay close to their random initialisation (\hyperref[edfig:10]{Extended Data Fig. 10d}). Counterfactual decomposition and steering still isolate latents whose effect depends on expression magnitude (\hyperref[edfig:10]{Extended Data Fig. 10m,n}). Chromatin accessibility shows the same entanglement in controlled scATAC autoencoders, where depth can be decoded from the latent space and an autoencoder trained to remove it loses most of its annotation as well (\hyperref[edfig:10]{Extended Data Fig. 10p}).

\section*{Discussion}

A single count does not need hundreds of dimensions when a narrow value pathway is paired with a wide backbone that combines genes. Whole-network narrowing degrades performance because it constrains that backbone as well. Allocating width by computational role separates the two requirements. A compact two-feature encoder improves retraining and annotation on the value side, while on the sparse panel pairwise attention adds about 17 percentage points of gain recovered over set pooling.

Gene detection carries most downstream transfer gain in annotation and surface-protein prediction. This reliance on detection connects the models' input dependence to the discreteness of the measurements, because many detected transcripts share the same small count while detection itself varies across cell types and conditions. Magnitudes still contribute to some perturbation readouts. Evaluating encoding choices by task therefore gives a clearer account than assuming that any preserved transfer score requires precise expression values.

Beyond cohort means, the white box identifies a shared functional component of gene offsets and a gene-specific readout of a low-dimensional cell state. Depth leads in the controlled models, whereas lineage contributes more in pretrained scGPT. These leading scalars follow the objective and preprocessing and predict responses to thinning, whereas longer training adds dimensions that tie training dynamics to the complexity of the prediction function. Future evaluations should measure how much a model adds beyond baselines built from depth and cell type, alongside its expression-prediction accuracy.

Interpretability requires a separate comparison between what can be decoded and what contributes through a proposed mechanism. Detection projections, gene-frequency controls, attention marginals, random directions and untrained networks provide accessible baselines for that comparison, while channel-uninformative models show that embedding-path attribution can credit a channel that carries no information. These controls complement observations that attention can recover curated regulatory edges {[}\hyperlink{ref:3}{3}{]} while adding little over gene-level features in perturbation-target prediction {[}\hyperlink{ref:22}{22}{]}.

Depth also reflects biological variation, as mitochondrial and ribosomal fractions couple measurement quality to metabolic and translational state. Distinguishing regulatory mechanisms from that shared cell state calls for interventional data assessed against calibrated noise {[}\hyperlink{ref:10}{10}--\hyperlink{ref:12}{12}, \hyperlink{ref:29}{29}, \hyperlink{ref:30}{30}{]}, and future work should test how training to convergence and interventional objectives change the cell state beyond depth and lineage.

\clearpage

\section*{Methods}

\subsection*{Cohorts, splitting and normalisation}

\begin{figure}[p]
\phantomsection\label{edfig:1}
\centering
\includegraphics[width=\textwidth,height=0.68\textheight,keepaspectratio]{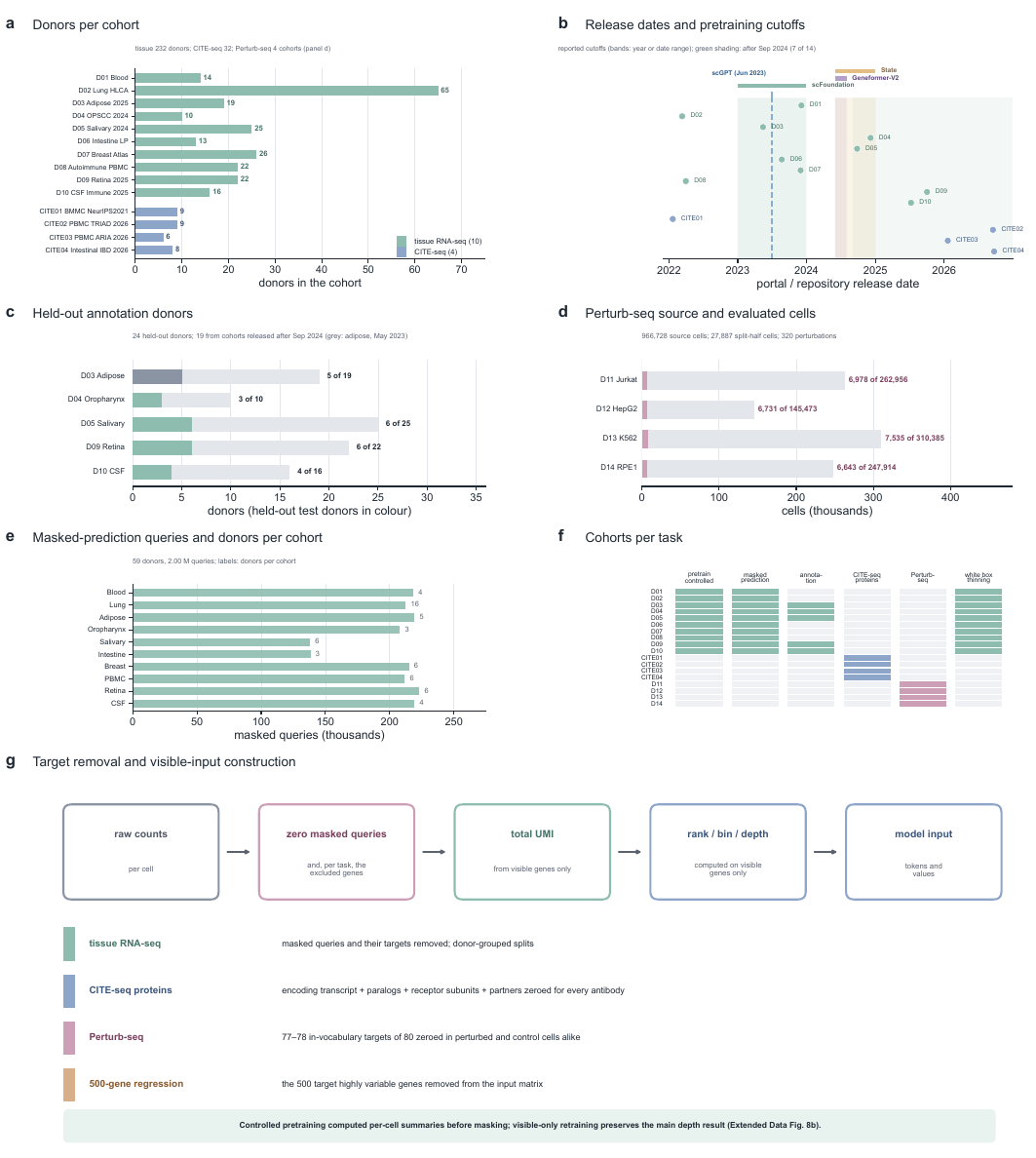}
\vspace{4pt}
\caption*{\footnotesize {\sffamily\bfseries Extended Data Figure 1 \textbar{} Cohort timing and donor splits define the evaluation populations.} \textbf{a}, Donor inventories across tissue transcriptomes and paired transcript--surface-protein cohorts; bars show donor counts by modality, and perturbation cohorts appear in \textbf{d}. \textbf{b}, Repository release dates relative to model pretraining cutoffs. Points mark cohorts, lines dated cutoffs, bands cutoff windows and green shading the later-release subset. \textbf{c}, Donor allocation across cell-type annotation cohorts, with coloured bars identifying held-out donors within cohort totals and grey distinguishing adipose tissue in the temporal sensitivity analysis. \textbf{d}, Perturb-seq populations across 320 perturbations. Pale and coloured bars distinguish source and split-half evaluation cells, respectively. \textbf{e}, Masked-prediction evaluation coverage across the ten tissue cohorts, with bar lengths encoding query counts and labels giving donor counts across 59 held-out donors. \textbf{f}, Cohort-by-task assignment matrix, where coloured tiles mark participation and colours identify assay modalities. \textbf{g}, Schematic of target removal and visible-input construction. Arrows order preprocessing, task-specific rows identify excluded transcripts, and the footer distinguishes controlled pretraining summaries from visible-only rebuilding (Methods).}
\end{figure}

\begin{figure}[p]
\phantomsection\label{edfig:2}
\centering
\includegraphics[width=\textwidth,height=0.68\textheight,keepaspectratio]{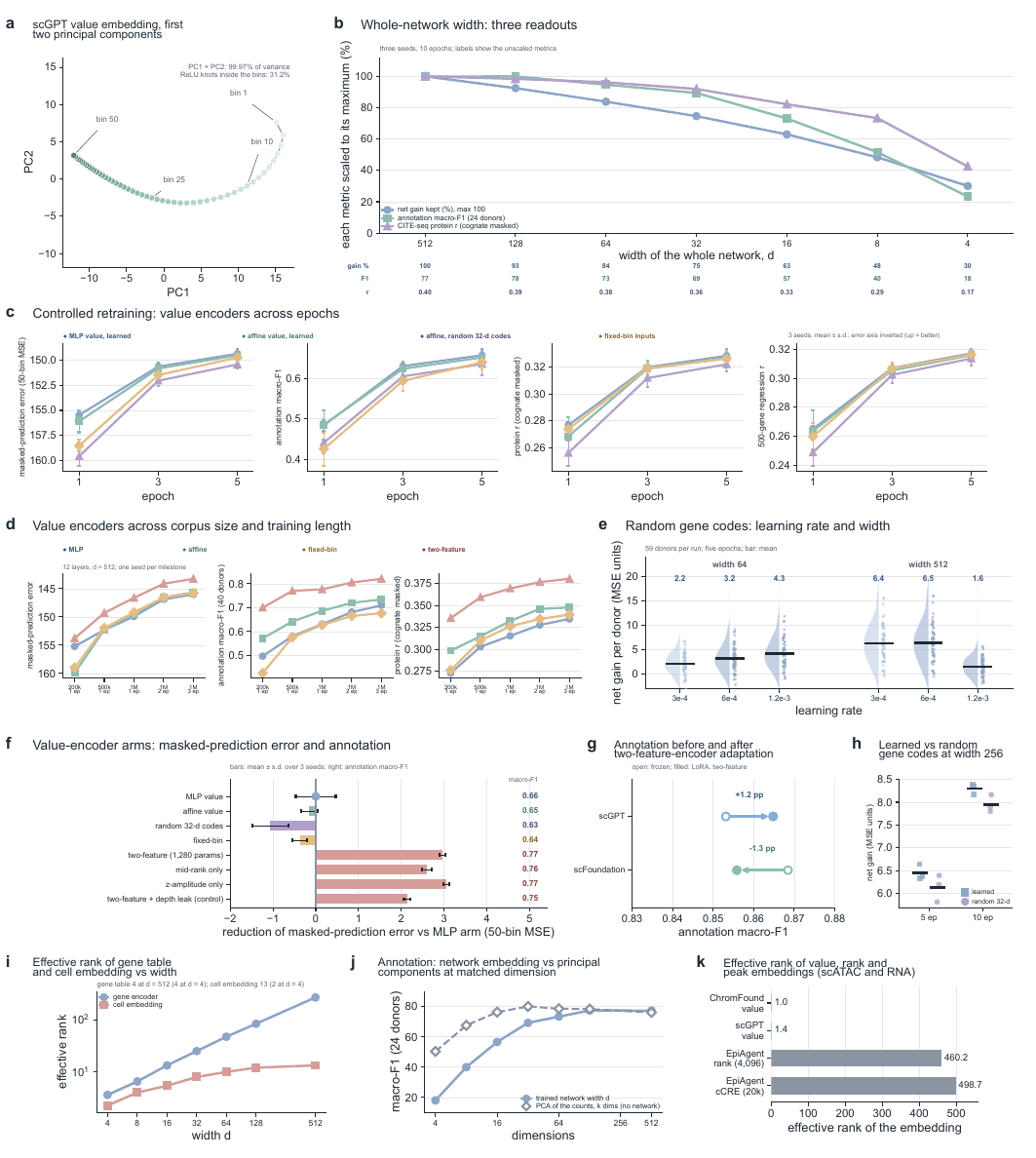}
\vspace{4pt}
\caption*{\footnotesize {\sffamily\bfseries Extended Data Figure 2 \textbar{} Value-encoder comparisons separate scalar encoding from backbone capacity.} \textbf{a}, Principal-component projection of pretrained scGPT value embeddings, coloured by input bin with geometric summary annotations. \textbf{b}, Whole-network width scans across masked prediction, cell-type annotation and surface-protein prediction; curves are maximum-scaled while labels retain original units. \textbf{c}, Controlled-Transformer training trajectories by value encoder, with bars giving three-run mean $\pm$ standard deviation and lower error plotted upwards. \textbf{d}, Value-encoder scaling across corpus sizes and training durations, with symbols distinguishing encoders. \textbf{e}, Masked-prediction net gain for random-code models across learning rates and widths; dots represent 59 donors per run and ticks mark means. \textbf{f}, Two-feature value-encoder ablations. Bars give three-run error drops $\pm$ standard deviation on masked prediction, and text labels give macro-F1. \textbf{g}, Cell-type annotation in pretrained scGPT and scFoundation before (open circles) and after (filled circles, paired by arrows) two-feature-encoder adaptation. \textbf{h}, Net gain for learned versus random gene codes at matched width. Dots represent individual runs and ticks mark means. \textbf{i}, Effective rank of the gene table and cell embedding across widths, plotted as three-run mean $\pm$ standard deviation bands. \textbf{j}, Cell-type annotation across 24 donors from network embeddings versus matched-dimension principal-component projections, distinguished by line style. \textbf{k}, Effective rank of value, rank and peak-identity embeddings in chromatin-accessibility and RNA models (Methods).}
\end{figure}

RNA-based evaluation covered 18 human single-cell cohorts spanning three assay modalities (\hyperref[edfig:1]{Extended Data Fig. 1a,d} and Supplementary Table 1). Ten multi-donor droplet single-cell or single-nucleus RNA-sequencing studies came from CZ CELLxGENE Discover {[}\hyperlink{ref:31}{31}{]}. They cover peripheral blood after SARS-CoV-2 infection and mRNA vaccination {[}\hyperlink{ref:32}{32}{]}, integrated lung core {[}\hyperlink{ref:33}{33}{]}, subcutaneous white adipose tissue {[}\hyperlink{ref:34}{34}{]}, oropharyngeal squamous cell carcinoma {[}\hyperlink{ref:35}{35}{]} and minor salivary gland in Sjögren disease {[}\hyperlink{ref:36}{36}{]}. They also include paediatric ileal lamina propria {[}\hyperlink{ref:37}{37}{]}, primary breast tumour {[}\hyperlink{ref:38}{38}{]}, autoimmune and immunodeficiency peripheral blood {[}\hyperlink{ref:39}{39}{]}, developing fetal retina {[}\hyperlink{ref:40}{40}{]} and leptomeningeal cerebrospinal fluid immune cells {[}\hyperlink{ref:41}{41}{]}.

Four multi-donor cellular indexing of transcriptomes and epitopes by sequencing (CITE-seq) studies {[}\hyperlink{ref:42}{42}{]} from NCBI Gene Expression Omnibus cover bone marrow mononuclear cells, trisomy-8 autoinflammatory peripheral blood, lecanemab amyloid-related imaging abnormality peripheral blood {[}\hyperlink{ref:43}{43}{]} and inflammatory bowel disease intestinal mucosa. Pooled CRISPR-interference (CRISPRi) Perturb-seq covers Jurkat and HepG2 {[}\hyperlink{ref:30}{30}{]}, K562 and RPE1 {[}\hyperlink{ref:29}{29}{]}. Supplementary Table 1 gives repository accessions, dataset identifiers, portal release dates, donor counts and cell counts; \hyperref[edfig:1]{Extended Data Fig. 1} summarises task assignments and preprocessing.

We evaluated public pretrained scGPT whole-human (33 million cells, 12 layers, width 512) {[}\hyperlink{ref:1}{1}{]}, scFoundation 50M-0.1B-res (50 million cells, 19,264 genes) {[}\hyperlink{ref:2}{2}{]}, Geneformer-V2 gc104M (12 layers, width 768, 4,096-token context) {[}\hyperlink{ref:3}{3}, \hyperlink{ref:4}{4}{]}, and Arc State/State embedding model SE-100M (State) (8 layers, width 1,024, 2,048-slot context) {[}\hyperlink{ref:5}{5}{]}. State initialised gene features with pretrained 5,120-dimensional ESM-2 protein embeddings {[}\hyperlink{ref:44}{44}{]}.

Tissue and CITE-seq fitting, validation and evaluation used disjoint biological donors, whereas Perturb-seq used perturbation split halves. Downstream normalisations, ranks, quantile boundaries and depth summaries used unmasked transcripts. Exceptions were the positive-only scFoundation protocol, which retained pre-masking depth, and scGPT calibration with full-cell quantile boundaries; Supplementary Note 2 describes their visible-only sensitivity checks. Controlled pretraining computed per-cell summaries before query masking; evaluation rebuilt them from visible genes. The visible-input pretraining controls below test this difference.

For each CITE-seq antibody, we removed its encoding transcript and curated paralogs, receptor subunits and binding partners before computing visible counts per ten thousand ($\text{CP10k}_{\text{vis}}$). Perturb-seq evaluation used 27,887 split-half cells from 966,728 source cells. We removed all in-vocabulary perturbation targets identically from perturbed and non-targeting cells before normalisation: 77 to 78 of the 80 targets per cohort, 310 total. Expression-regression inputs excluded the 500 target highly variable genes.

\subsection*{Pretrained encoders, interventions and probes}

We extracted representations with dropout disabled (\hyperref[fig:1]{Figs. 1b} and 2, and \hyperref[edfig:2]{Extended Data Figs. 2} and 4). We centred each scalar value-embedding curve across input levels and applied singular-value decomposition, then projected onto its first $k \in \{1, 2, 4, 8\}$ directions. Comparators were the original encoder, an ordinary-least-squares affine value encoder and a constant. Evaluations used lung tissue and K562 cells.

For scGPT, expressed genes were encoded in 50 within-cell quantile bins, up to 1,199 genes plus the classification token at $-2.0$. The affine value encoder replaced the post-normalisation value-encoder output over $v\in[1,49]$ while keeping the classification token unchanged. Clamped input set every detected-gene value to 25.5; value shuffling permuted bins among detected genes. Identity permutation sampled vocabulary genes of matched count while retaining the value multiset. Cell representations were $L_2$-normalised classification-token states of width 512.

scFoundation received counts over 19,264 genes, with both depth tokens set to $\log_{10}(N_{\text{vis}})$, and returned pooled 3,072-dimensional representations. The affine fit covered positive values $v\in[0.05,6.0]$. Clamped input set positive values to 1.0 while retaining depth tokens; the fixed-depth control set both tokens to 3.5. Value shuffling permuted positive values within cells; identity permutation shuffled nonzero gene identities.

Geneformer-V2 inputs were visible CP10k counts divided by published gene-median factors and sorted, with up to 4,094 gene tokens between boundary tokens. Representations were $L_2$-normalised, mean-pooled penultimate-layer states of width 768. Controls omitted median scaling, randomised detected-gene order for clamped input, reversed rank order, permuted identities or used random initialisation.

State mapped positive unique molecular identifier (UMI) counts to 19,790 protein-coding genes, transformed them through $\log(1+\text{UMI})$ to relative weights $100w_g$, and sorted them into 2,047 slots, padding with unexpressed genes. Its 10-bin soft-count embedding produced $L_2$-normalised classification-token states of width 1,024. The affine fit covered $c\in[0.015,0.60]$. Clamped input set positive counts to 1 before weighting; a fixed-cardinality control held positive slot weights constant across cells. Value shuffling permuted weights among expressed slots.

Net gain is the improvement over the cohort-mean predictor: the score increase for higher-is-better measures or the error reduction for masked prediction. For altered input $A$, retention is $[G(A)-G(I)]/[G(O)-G(I)]$, where $G$ is net gain, $I$ is identity-permuted input and $O$ is original input.

Probes were refitted for each input. Cell-type annotation used multinomial logistic regression; expression features were standardised using fit cells and divided by $\sqrt d$, with regularisation selected on validation cells from $\{0.1, 0.5, 1, 5, 10, 50\}$. Label budgets were two labelled cells per class (two-shot), then 5, 10, 25, 50, 100 and 200 cells per class and the full split, with five subsampling replicates. Full-split annotation used a common five-cohort protocol; few-shot Geneformer-V2 and State scores used paired intervention runs. Temporal sensitivity excluded adipose tissue and retained 19 of the 24 held-out donors.

CITE-seq prediction used donor-disjoint $L_2$-regularised ridge regression of centred-log-ratio protein abundance, with donor-grouped cross-validation across 32 donors and 463 non-isotype channels, 461 with nonzero staining. Partial fine-tuning initialised linear heads from pretrained-representation probes {[}\hyperlink{ref:45}{45}{]}. AdamW {[}\hyperlink{ref:46}{46}{]} then trained the value module, top two Transformer blocks and readout across two CITE-seq cohorts (4 held-out donors, 264 channels) and eight held-out RNA-sequencing donors at full labels and ten labels per class (Supplementary Table 5). Perturbation consistency was the cosine between uncentred shifts of $L_2$-normalised embeddings, averaged over ten random split halves per perturbation.

Annotation fine-tuning on adipose and oropharyngeal carcinoma donors (full training splits and 10 labelled cells per class) compared original, affine and clamped inputs in scGPT and scFoundation, including clamped input with fixed depth. These RNA-sequencing donors differed from the CITE-seq fine-tuning population, which supplied the protein comparison in \hyperref[edfig:4]{Extended Data Fig. 4c}.

\subsection*{Chromatin-accessibility models}

We analysed single-cell chromatin accessibility in bone marrow mononuclear cells (BMMC; GSE194122; 20,000 cells, 10 donors), brain (GSE312668; 18,000 cells, 9 donors), cerebellum (GSE331288; 21,660 cells, 10 donors) and DOGMA (GSE156478; 16,000 cells, 4 libraries). Three-fold grouped cross-validation used a stratified subsample of at most 2,100 cells per cohort, grouping by donor in BMMC, brain and cerebellum. DOGMA groups are its four libraries (LLL\_CTRL, LLL\_STIM, DIG\_CTRL and DIG\_STIM), not independent donors. Unlike the RNA transfer analyses above, each accessibility probe was fitted on intact training-fold embeddings and evaluated unchanged on test-fold interventions, with performance scored by macro-F1. We compared EpiAgent classification-token and mean embeddings, term frequency--inverse document frequency (TF-IDF) with latent semantic indexing, and ChromFound. Interventions binarised counts, permuted non-zero values within each cell or permuted peak coordinates within 20 cohort-level feature-detection-frequency strata using a fixed random seed. Paired BMMC and DOGMA RNA supplied scGPT controls (Supplementary Table 1D).

The share of non-zero entries equal to one or two was computed on called peaks. Value-embedding effective rank was derived from singular values of the value or rank embedding (\hyperref[edfig:2]{Extended Data Fig. 2k}). Controlled BMMC models included a binary autoencoder, an autoencoder with a linear value input, a Poisson autoencoder with a depth adversary and a joint RNA--ATAC model, using 15,241 training cells, 4,759 held-out-donor test cells and three independent runs (\hyperref[edfig:10]{Extended Data Fig. 10p}).

\subsection*{Count sparsity and detection-explained variance}

We used raw UMI counts from 21,000 evaluation cells in adipose, oropharyngeal carcinoma, retina and cerebrospinal fluid {[}\hyperlink{ref:13}{13}, \hyperlink{ref:14}{14}, \hyperlink{ref:47}{47}, \hyperlink{ref:48}{48}{]}. Salivary-gland pseudo-counts were excluded. Single- and double-molecule frequencies were computed over nonzero entries and averaged equally across cohorts. For gene $g$, $Y_g=\log(1+\text{CP10k}_g)$ and $I_g=\mathbf1\{C_g>0\}$. Detection-explained variance was $\operatorname{Var}(\mathbb E[Y_g\mid I_g])/\operatorname{Var}(Y_g)$, from $\operatorname{Var}(Y_g)=\mathbb E[\operatorname{Var}(Y_g\mid I_g)]+\operatorname{Var}(\mathbb E[Y_g\mid I_g])$. We also computed it using held-out-donor cells only.

\subsection*{Noise floors and masked-output calibration}

Within-cell rank quantiles assign equal positive counts to a contiguous interval of integer bins $\{a+1, \dots, b\}$. Uniform tie-breaking has conditional variance $\sigma_{\mathrm{tie}}^2(a,b)=((b - a)^2 - 1)/12$. Averaging over masked queries gives the tie floor $L_{\mathrm{tie}}$; subtracting it from the out-of-cohort gene-mean error $G$ gives the post-tie window $W_{\mathrm{tie}}$. These are squared-bin quantities. We report cohort-equal, query-weighted and mean-of-cohort-ratio aggregations separately (\hyperref[edfig:5]{Extended Data Fig. 5e}).

For the technical-sampling sensitivity {[}\hyperlink{ref:47}{47}, \hyperlink{ref:48}{48}{]}, we held other counts fixed and resampled one positive masked query from a zero-truncated Poisson distribution, $\mathbb P(C=k\mid C>0;\lambda)=(\lambda^k/k!)/(e^\lambda-1)$. The upper truncation was $\lambda + 10\sqrt\lambda + 10$, with tail mass below $10^{-11}$. Integrating tie-breaking over recomputed boundaries gave $L^\star_{cg}=\mathbb E[m_2(C_{cg})\mid C_{cg}>0]-(\mathbb E[\mu_{\mathrm{bin}}(C_{cg})\mid C_{cg}>0])^2$, and $W^\star=G-L^\star$. Expression rates used empirical-Bayes neighbour smoothing with $k_{\mathrm{nn}}\in\{10, 100\}$ and 20 global-prior pseudo-counts. We also measured whether single-query resampling changed a visible bin, against a 5\% threshold; this distinguishes the sampling assumption from the exact tie calculation.

Post-hoc scGPT prediction used 25\% positive masked queries and 50-bin targets. scFoundation used either mixed queries (30\% positive and 3\% zero) with visible normalisation or positive-only queries (25\% positive) retaining full-cell depth. We evaluated calibration both without target-cohort fitting and using 3,000 target-cohort training cells plus 750 validation cells from donors disjoint from the held-out set. For scGPT, candidate features combined expression-decoder and cell-context-decoder predictions, empirical-Bayes training gene means, gene-by-detection-count lookups, visible gene count and log visible transcript depth. Regularised per-gene and cross-validated ridge combinations fitted the target bins.

We calibrated scFoundation by combining decoder predictions and detection-rate de-attenuation, $\mathrm{pred}/P_{\mathrm{det}}$, with out-of-cohort gene means, visible detection and log visible depth. Detection rates came from out-of-cohort marginals or $k=100$ binary-detection neighbours; empirical-Bayes per-gene ridge blending combined features. Marginal-only rescaling tested dependence on neighbour lookup. Positive-only scores and mixed-query scores used different targets and normalisations and are presented separately. scGPT boundary sensitivity recomputed quantiles using visible genes only before repeating calibration. Supplementary Note 2 describes the assumptions of the noise estimates; Supplementary Table 2 gives exact comparisons.

\subsection*{Controlled pretraining and dimensionality}

Controlled Transformers (Transformers trained from scratch under controlled conditions) used six pre-normalisation layers, width 256, 8 heads, up to 1,024 tokens, 20,080 protein-coding genes and AdamW {[}\hyperlink{ref:46}{46}{]}, trained on 199,936 cells from training donors of all ten tissue cohorts. Three independent runs shared data order, masks and schedules across conditions. The rank-target objective used 50 within-cell rank bins. Visible tokens attended to visible keys, whereas masked queries attended to visible keys and themselves so that no other token read masked values.

The MLP value encoder was a two-layer rectified-linear network with layer normalisation. Its affine comparator was $a+(v/50)b$, $a,b\in\mathbb R^{256}$, variance-matched at initialisation. Learned gene tables were compared with random 32-dimensional Gaussian codes, row-normalised to length $\sqrt{256}$ and projected through a shared orthogonally initialised $32\times256$ layer with layer normalisation. Both had trainable per-gene output biases.

The width scan used $d\in\{64, 128, 256, 512\}$, with the same construction scaled to $d$ (\hyperref[fig:1]{Fig. 1g} and \hyperref[edfig:2]{Extended Data Fig. 2e,h}). Random-code five-epoch runs used three replicates at every width; ten-epoch runs used three at width 256 and one elsewhere. Learned-code runs used three at width 256 and one at widths 64 and 512. At width 512, learning-rate multipliers $0.5\times$, $1\times$ and $2\times$ tested optimisation dependence (Supplementary Table 2).

Calibration used 3,000 training-donor cells from the target cohort and in-cohort gene means. Without target-cohort calibration, predictions were evaluated against out-of-cohort gene means.

\begin{figure}[p]
\phantomsection\label{edfig:3}
\centering
\includegraphics[width=\textwidth,height=0.68\textheight,keepaspectratio]{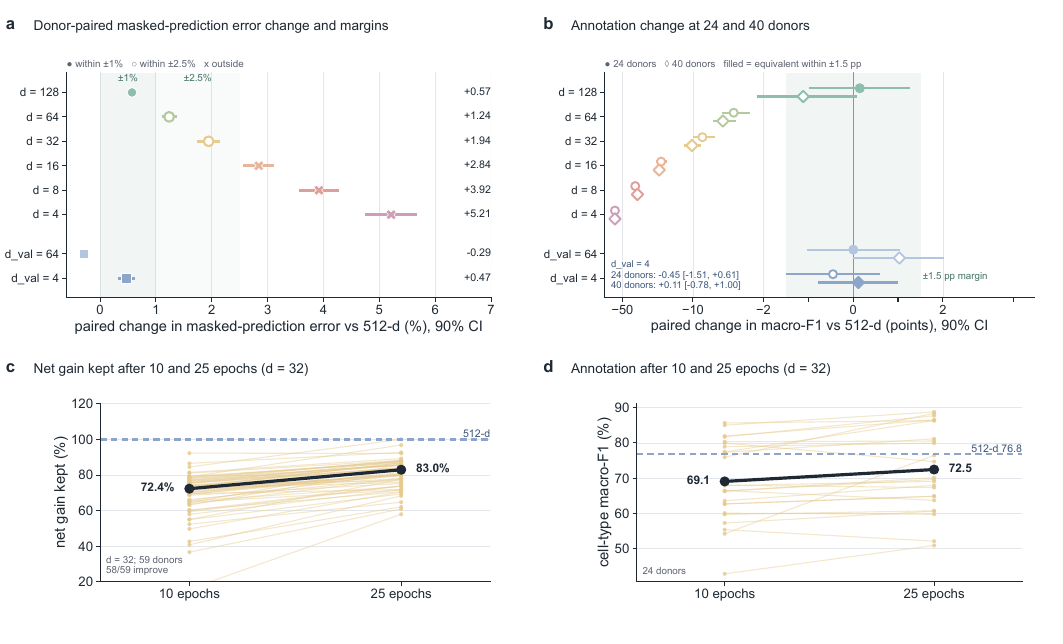}
\vspace{4pt}
\caption*{\footnotesize {\sffamily\bfseries Extended Data Figure 3 \textbar{} Equivalence tests distinguish value-track narrowing from whole-network narrowing.} \textbf{a}, Donor-paired masked-prediction error differences between narrowed models and their wide references across 59 held-out donors. Bars show two-sided 90\% paired-donor \textit{t} intervals for whole networks (circles) and value tracks (squares); filled markers pass $\pm$1\%, open markers pass $\pm$2.5\%, and crosses fail both margins. \textbf{b}, Cell-type annotation contrasts under the 24-donor (circles) and 40-donor (diamonds) protocols. Bars show two-sided 90\% paired-donor \textit{t} intervals, shading marks $\pm$1.5 points, and filled symbols denote equivalence within that margin. \textbf{c}, Longer training of a narrow whole network on masked prediction across held-out donors; thin lines pair individual donors, the thick line joins pooled net gains, and the dashed line marks the wide reference. \textbf{d}, Matched training-duration evaluation on cell-type annotation across 24 held-out donors. Thin lines pair donor scores, the thick line joins means, and the dashed line marks the wide model (Methods).}
\end{figure}

\begin{figure}[p]
\phantomsection\label{edfig:4}
\centering
\includegraphics[width=\textwidth,height=0.68\textheight,keepaspectratio]{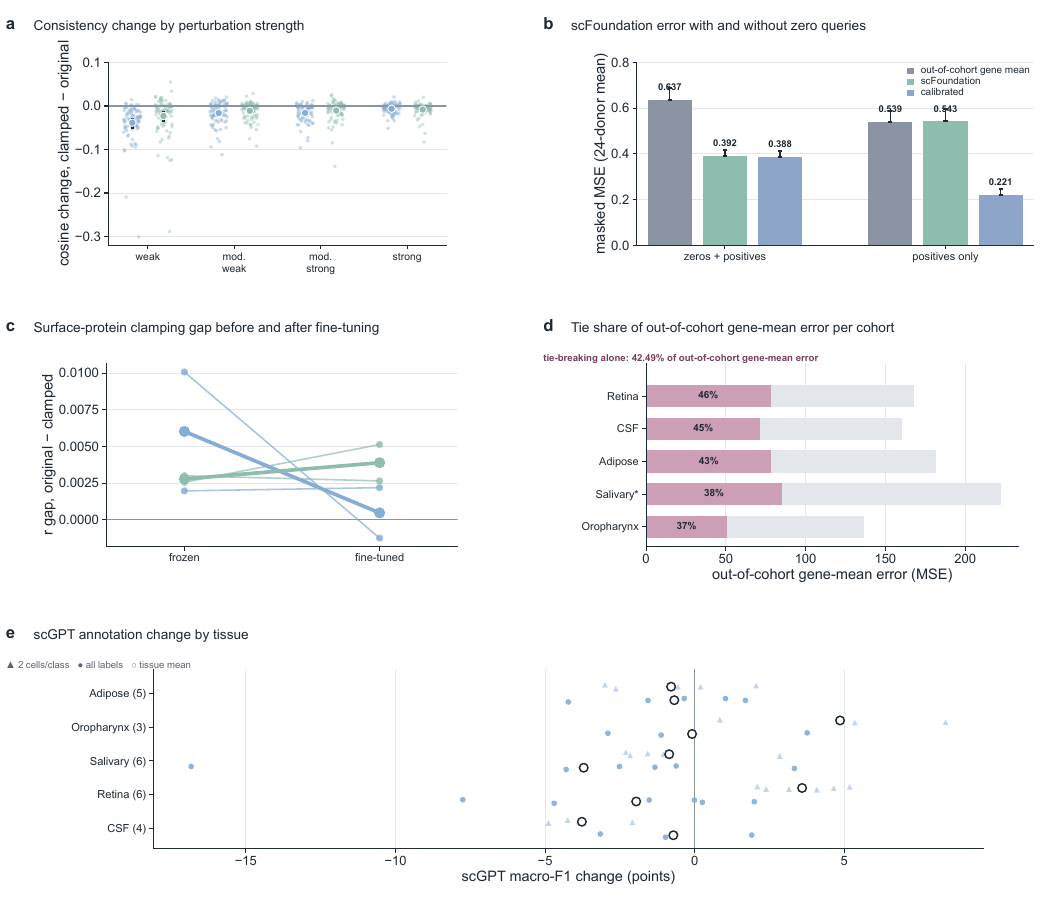}
\vspace{4pt}
\caption*{\footnotesize {\sffamily\bfseries Extended Data Figure 4 \textbar{} Clamping effects depend on the readout and evaluation population.} \textbf{a}, Clamping-induced change in perturbation-response consistency across perturbation-strength quartiles. Dots denote individual targets among 320 perturbations; summary points show mean cosine changes with 95\% perturbation-bootstrap intervals. \textbf{b}, scFoundation masked-prediction error under mixed zero--positive and positive-only query protocols across 24 donors. Bars distinguish out-of-cohort gene means, raw predictions and calibrated predictions; upward error bars show standard errors of donor means. \textbf{c}, Original-minus-clamped surface-protein correlation gaps before and after partial fine-tuning across four held-out donors, with thin lines connecting donor gaps and thick lines connecting cohort means by model colour. \textbf{d}, Cohort-level partition of rank-bin out-of-cohort gene-mean error. Coloured bar segments show analytical tie variance within total error, and remaining segments show error beyond ties. \textbf{e}, Tissue-grouped pretrained-scGPT cell-type annotation contrasts; triangles mark two labelled cells per class, circles all labels, and rings tissue means, with donor counts labelled (Methods).}
\end{figure}

\begin{figure}[p]
\phantomsection\label{edfig:5}
\centering
\includegraphics[width=\textwidth,height=0.68\textheight,keepaspectratio]{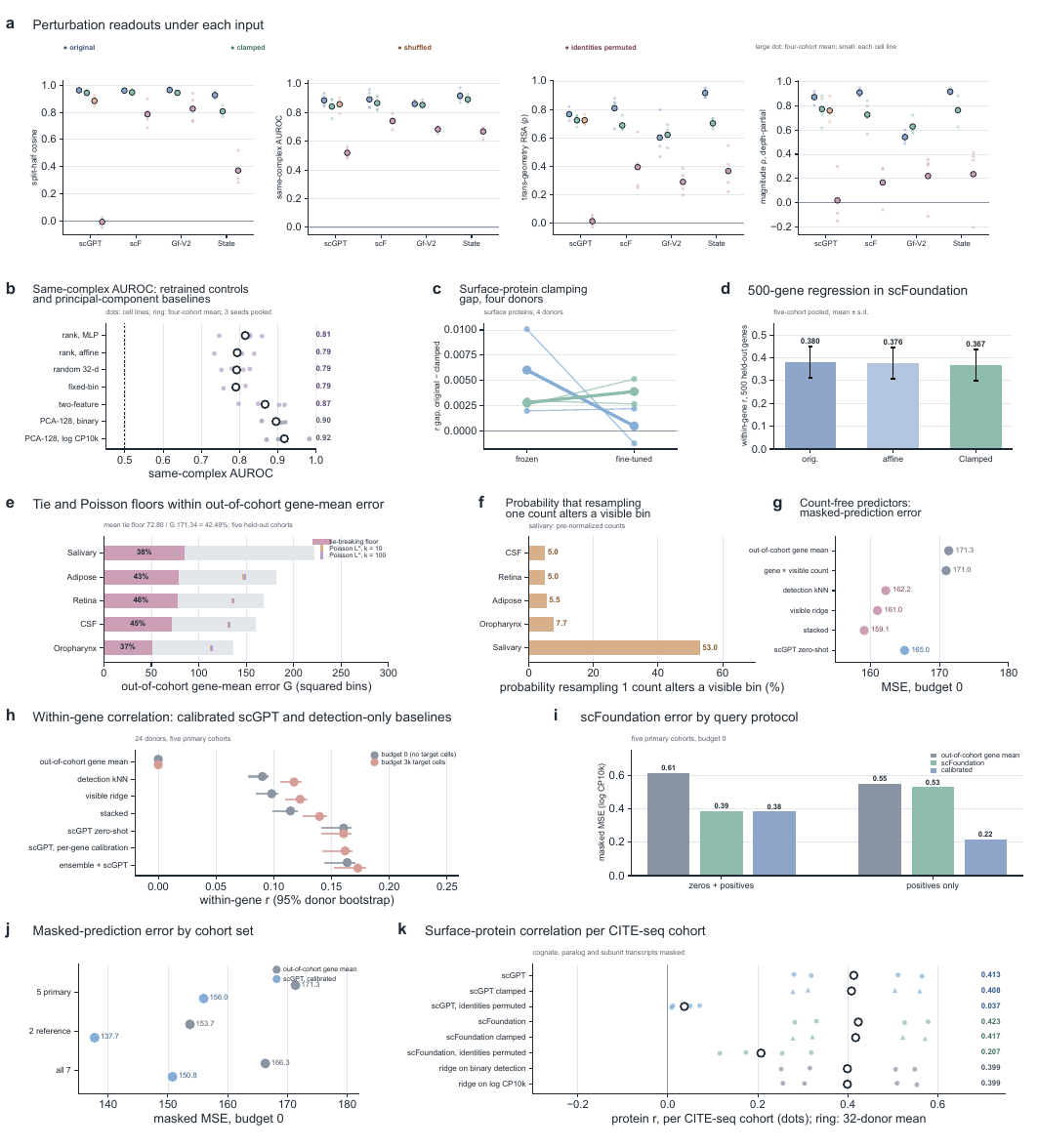}
\vspace{4pt}
\caption*{\footnotesize {\sffamily\bfseries Extended Data Figure 5 \textbar{} Count noise and calibration shape masked-prediction and transfer readouts.} \textbf{a}, Input-intervention profiles across perturbation consistency, complex recovery, response geometry and depth-adjusted magnitude; small dots denote cell lines and large dots denote four-cohort means. \textbf{b}, Protein-complex recovery for controlled and principal-component baselines. Dots mark cell lines, and rings mark mean area under the receiver operating characteristic curve (AUROC). \textbf{c}, Surface-protein clamping gaps before and after fine-tuning across four donors, with thin lines denoting cohorts and thick lines means. \textbf{d}, Held-out-gene prediction across scFoundation input variants (five-cohort mean within-gene correlation $\pm$ standard deviation). \textbf{e}, Out-of-cohort gene-mean error alongside analytical tie variance (bars) and Poisson noise floors under alternative neighbourhood sizes (ticks). \textbf{f}, Visible-bin instability under single-query count resampling across cohorts, with salivary counts pre-normalised. \textbf{g}, Aggregate masked-prediction error of count-free predictors versus pretrained scGPT evaluated without target-cohort fitting. \textbf{h}, Within-gene correlation of calibrated pretrained scGPT across fitting budgets. Points give estimates and bars give 95\% stratified donor-bootstrap intervals. \textbf{i}, Calibrated scFoundation error by query protocol in log counts per 10,000 (CP10k). \textbf{j}, Masked-prediction error across temporal cohort subsets, coloured for out-of-cohort gene means versus calibrated scGPT. \textbf{k}, Surface-protein correlation across model inputs and count baselines with cognate-related transcripts masked; dots mark cohorts and rings show 32-donor means (Methods).}
\end{figure}

\begin{figure}[p]
\phantomsection\label{edfig:6}
\centering
\includegraphics[width=\textwidth,height=0.68\textheight,keepaspectratio]{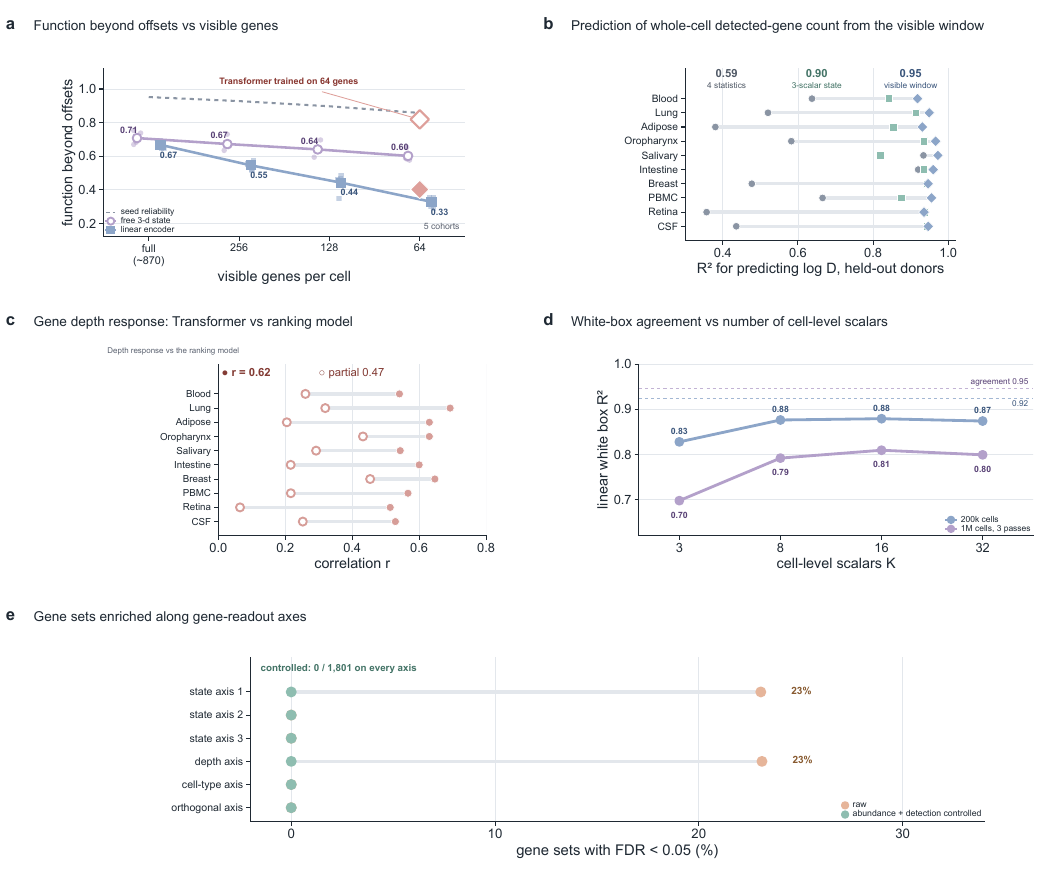}
\vspace{4pt}
\caption*{\footnotesize {\sffamily\bfseries Extended Data Figure 6 \textbar{} Sparse context limits linear recovery of a low-dimensional cell state.} \textbf{a}, White-box fits across progressively subsampled visible contexts in five cohorts. Small points denote cohorts; curves distinguish linear encoders, free states and independent-model reliability, while diamonds mark models trained with sparse context. \textbf{b}, Prediction of whole-cell detected-gene count from visible-input features on held-out donors across ten cohorts, distinguishing window features, cell state and summary statistics with pooled agreement annotated. \textbf{c}, Gene-specific depth responses versus a multinomial ranking model. Filled and open points distinguish raw correlations from correlations adjusted for abundance and detection across cohort rows. \textbf{d}, Linear white-box agreement across numbers of cell-level scalars and training scales; lines identify training conditions, and dashed horizontal references mark independent-model agreement. \textbf{e}, Gene-set enrichment along gene-readout axes before and after abundance control. Paired points show significant-set shares among 1,801 sets at a false-discovery-rate threshold of 0.05 (Methods).}
\end{figure}

Whole-network narrowing trained every component at $d\in\{512, 128, 64, 32, 16, 8, 4\}$ for 10 epochs on 200,000 cells, with three runs. Longer narrow whole-network runs are shown in \hyperref[edfig:3]{Extended Data Fig. 3c,d}. The two-track Transformer retained width 512 for identity, attention routing and cell embeddings, while values stayed at $d_{\text{val}}\in\{512, 64, 4\}$ through every layer and the masked head. The value track contains 20,488,196 parameters at width 512 and 2,080 at width 4; total parameters are 35,500,554 and 15,014,438, respectively. Figure 1i retains component-specific denominators: whole-network backbone/value parameters for the whole-network scan and value-stream parameters for the two-track scan, each relative to its own width-512 reference. Value-track learning rates were $6\times10^{-4}$ at width 512 and $3\times10^{-3}$ at width 4, with 8 and 4 heads, respectively.

Width scans express masked-prediction net gain relative to the width-512 model of the same architecture. Donor-paired relative error was assessed over 59 donors with a $\pm$1\% equivalence margin. Intervals in \hyperref[edfig:3]{Extended Data Fig. 3a,b} are two-sided 90\% Student's t intervals on paired donor differences. Annotation used linear probes on 24 donors from five cohorts and a separate 40-donor protocol, with a $\pm$1.5-point margin; the value-track comparison meets this margin on the latter protocol. Its 24-donor 90\% interval is {[}$-$1.5077, +0.6051{]} points and narrowly fails equivalence; the 40-donor interval is {[}$-$0.7754, +1.0033{]} and passes.

The 15-arm value-encoder bottleneck sweep used three independent runs per arm, 200,000 cells, 20,080 genes and five epochs, at hidden width 512 (768 for the scFoundation-sized block). Blocks used rank-target or continuous log CP10k targets; bottleneck widths were $k\in\{1,2,4,8,16,32\}$ and full width, alongside an affine value encoder. Gates required masked-prediction mean squared error within $\pm$2.5\%, net gain within $\pm$2.5 percentage points and annotation within $-$1 point of the block's full-width encoder. Other downstream probes used $-$0.010 non-inferiority for expression and surface-protein correlation, overall and within cell type or lineage. Two-one-sided tests used $\alpha=0.05$ and 1,000 bootstrap draws. Effective rank counted singular directions of MLP value-encoder output needed for 99\% variance. The one-dimensional MLP bottleneck passed the masked-prediction and downstream criteria and exceeded the upper gain margin (103.7\% of full-width net gain).

\subsection*{Two-feature value encoder}

The two-feature encoder used the rank-two affine map $a+q_{\mathrm{mid}}b_{\mathrm{rank}}+z_{\mathrm{amp}}b_{\mathrm{amp}}$, where $q_{\mathrm{mid}}$ is the tie-free mid-rank of a count among the cell's detected genes and $z_{\mathrm{amp}}$ is log count standardised within the cell. We excluded raw depth from the classification token ({[}CLS{]}) and anchored the readout head to visible moments (mean, standard deviation, minimum positive value and log depth). The controlled six-layer, width-256 Transformer used 20,080 genes, three independent runs, five epochs, ten cohorts and 59 held-out donors. The two-feature encoder had 1,280 value-encoder parameters versus 66,816 for the MLP value encoder. Mid-rank-only and amplitude-only arms each had 768 parameters; a depth-leak control supplied depth through {[}CLS{]}. A common continuous target supported donor-paired contrasts against quantile affine, mid-rank-only and other ablation arms.

The graft replaced pretrained scGPT's 264,704-parameter value encoder with 2,560 parameters, unfroze layers 8--11 and trained for two epochs (3,124 steps) on a donor-disjoint 200,000-cell corpus. Scaling used a 12-layer, width-512 model and 1,000,000 cells (Supplementary Table 4).

\subsection*{Sparse-panel predictors and attention controls}

The controlled 1,024-gene positive-expression panel used donor-disjoint peripheral blood, 1,999 test cells and 5,230 queries, averaging 17.96 visible genes per cell. Gain share compared the training gene mean with the mean risk of the three reference Transformers trained for 12 passes. Non-neural predictors received visible identities and bins plus masked query identities. They combined detected-count-conditioned Gaussian means, gradient-boosted closure trees over visible and query summaries, and binary-detection Jaccard neighbours ($k=200$); incremental contributions were averaged over both orderings.

Architecture-matched Deep Sets {[}\hyperlink{ref:49}{49}{]} pooled visible genes and masked identities without self-attention, following the same 48-pass cosine schedule as the Transformers with evaluation at passes 12, 24 and 48. Further controls pooled visible genes alone, blocked masked keys, or kept masked keys and values at normalised initial embeddings. Positive-only panel queries reveal detection through their identity; full-vocabulary retraining tested the consequence.

\begin{figure}[p]
\phantomsection\label{edfig:7}
\centering
\includegraphics[width=\textwidth,height=0.68\textheight,keepaspectratio]{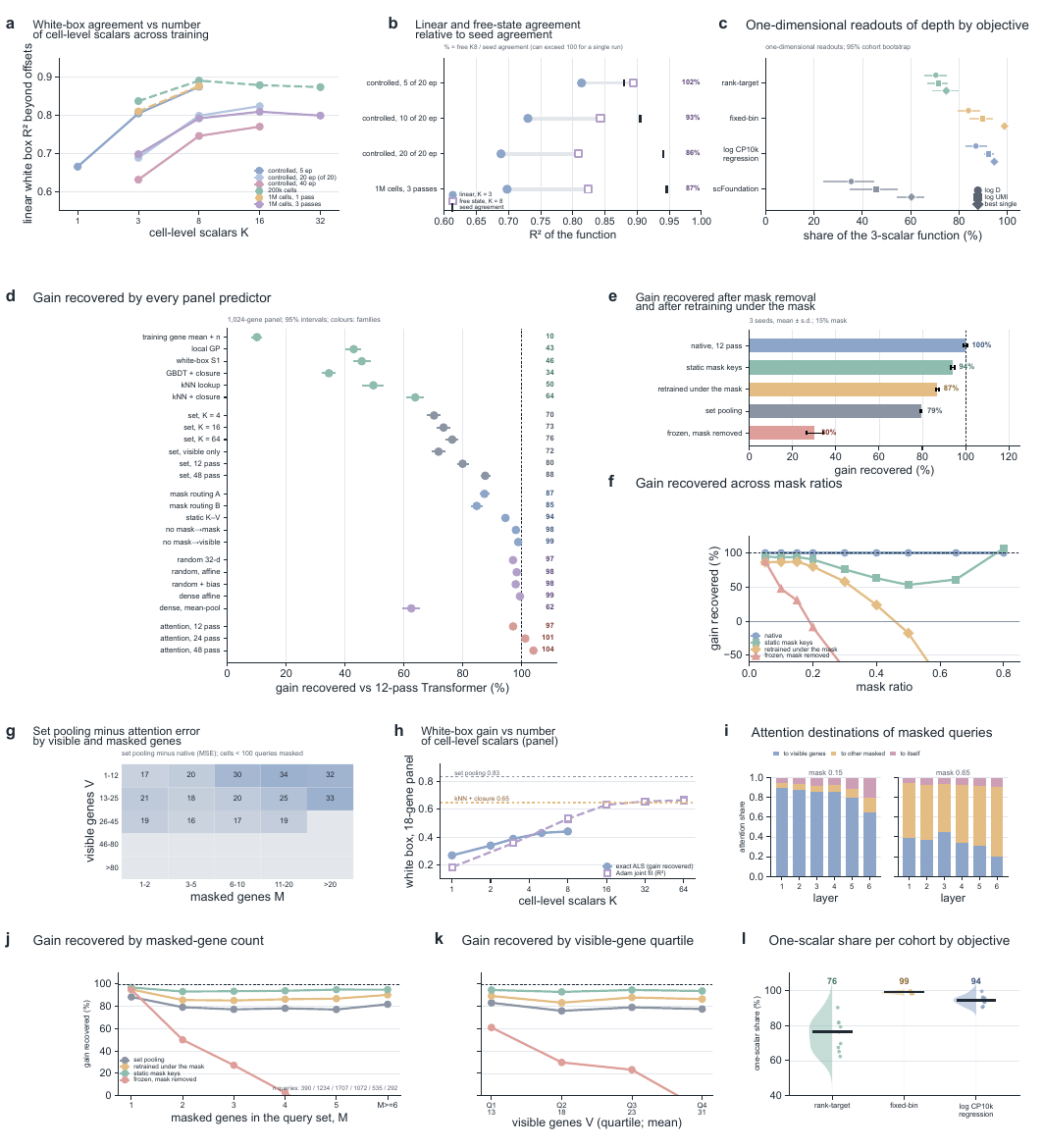}
\vspace{4pt}
\caption*{\footnotesize {\sffamily\bfseries Extended Data Figure 7 \textbar{} Sparse-panel controls distinguish attention routing from state dimensionality.} \textbf{a}, White-box dimensionality curves across training schedules and corpus sizes. \textbf{b}, Model agreement for linear versus free-state white-box fits across configurations, with ticks marking independent-run agreement. \textbf{c}, Single-covariate depth shares by training objective, with bars giving 95\% cohort-bootstrap intervals. \textbf{d}, Gain recovered on the sparse panel across predictor families; colours group predictor families and bars give 95\% run-by-cell bootstrap intervals. \textbf{e}, Gain recovered after post-hoc attention-mask removal versus retraining under the mask, plotted as three-run mean $\pm$ standard deviation. \textbf{f}, Routing-variant sweeps across query-mask ratios; curves trace gain recovered relative to the reference Transformer. \textbf{g}, Set-pooling minus attention error differences across visible- and masked-gene counts, with grey excluding bins below 100 queries. \textbf{h}, Sparse-panel white-box performance by number of cell-level scalars. Solid curves trace alternating-least-squares gain recovered, while dashed curves trace joint-fit distillation agreement. \textbf{i}, Layer-wise partition of masked-query attention among visible genes, other queries and self-attention. \textbf{j}, Gain recovered by routing variants across masked-query-count strata, with query counts labelled per stratum. \textbf{k}, Gain recovered for the same variants across visible-context quartiles, labelled with quartile bounds and mean visible-gene counts. \textbf{l}, Single-scalar contributions across nine cohorts by objective; dots denote cohorts and ticks mark medians (Methods).}
\end{figure}

\begin{figure}[p]
\phantomsection\label{edfig:8}
\centering
\includegraphics[width=\textwidth,height=0.68\textheight,keepaspectratio]{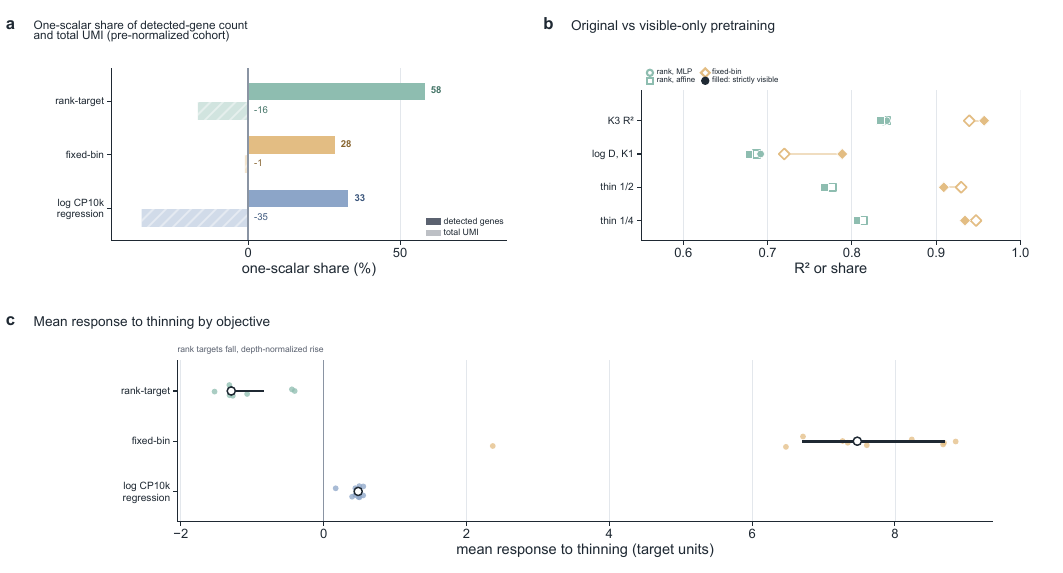}
\vspace{4pt}
\caption*{\footnotesize {\sffamily\bfseries Extended Data Figure 8 \textbar{} Objective and input construction determine which depth covariate describes the response.} \textbf{a}, Depth-predictor shares in the pre-normalised salivary-gland cohort across training objectives. Solid bars show detected-gene-count contributions and pale hatched bars show total unique molecular identifier (UMI) contributions, each expressed relative to the cell-state increment beyond gene offsets. \textbf{b}, Original versus visible-only pretraining, where open and filled markers distinguish input construction, connectors pair conditions, and marker shape separates rank-target from fixed-bin log counts per 10,000 (CP10k) encoders; rows report white-box agreement, detected-gene-count contribution and thinning-response agreement at both doses (Methods). \textbf{c}, Mean thinning responses across ten cohorts under rank-target, fixed-bin log CP10k and continuous-regression objectives. Coloured dots represent cohorts, rings show medians with 95\% cohort-bootstrap intervals, and the vertical line marks zero response in native target units for the half- and quarter-depth interventions in the schematic (Methods).}
\end{figure}

\begin{figure}[p]
\phantomsection\label{edfig:9}
\centering
\includegraphics[width=\textwidth,height=0.68\textheight,keepaspectratio]{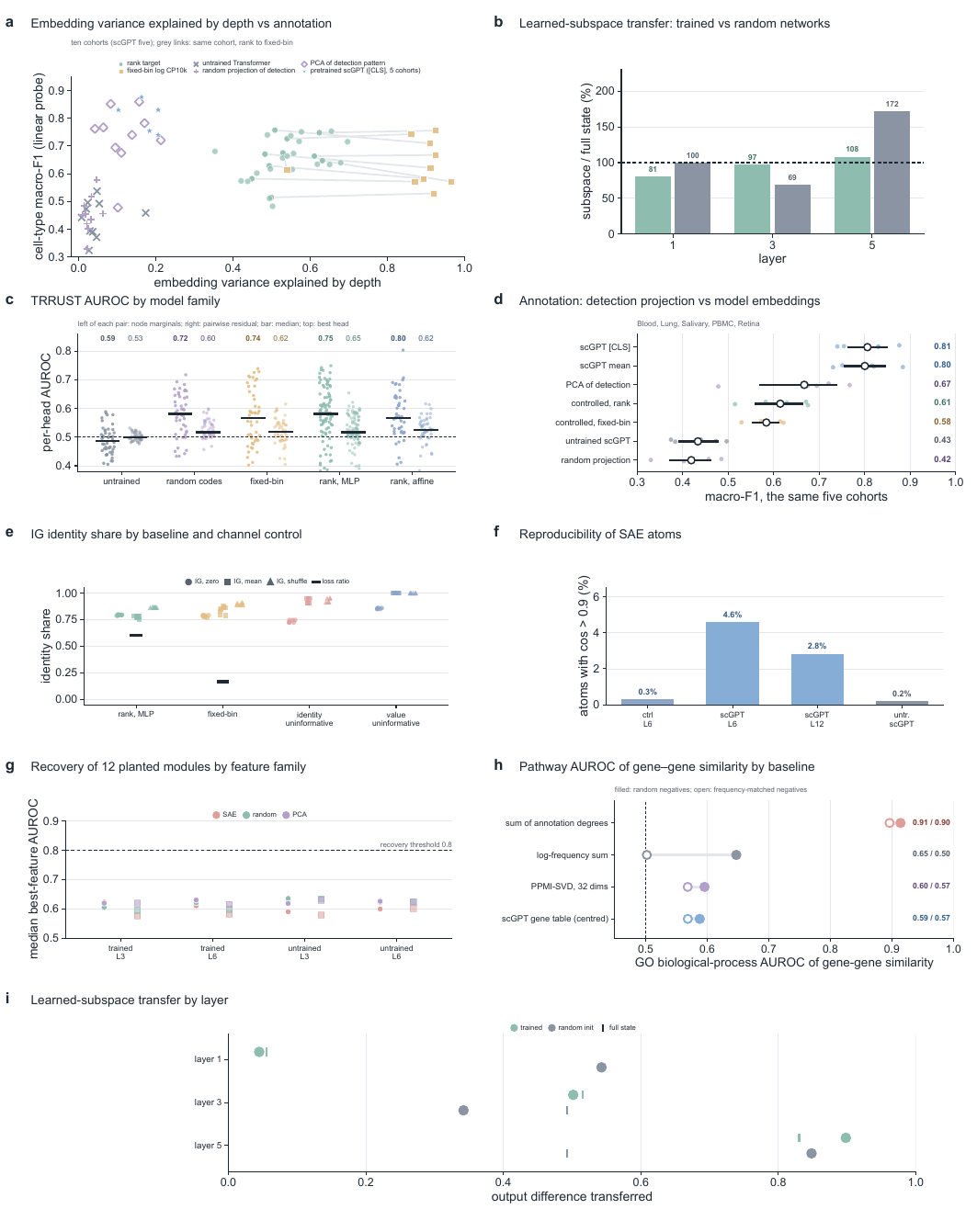}
\vspace{4pt}
\caption*{\footnotesize {\sffamily\bfseries Extended Data Figure 9 \textbar{} Matched nulls explain much of the structure recovered by interpretability readouts.} \textbf{a}, Depth-explained embedding variance versus cell-type annotation across ten cohorts (five for pretrained scGPT), with symbols identifying models and links pairing objectives. \textbf{b}, Learned-subspace swap transfer in trained and randomly initialised networks, expressed relative to each network's full-state swap. \textbf{c}, TRRUST regulatory-interaction recovery by attention component. Dots mark heads, and ticks mark median area under the receiver operating characteristic curve (AUROC) for paired marginals and residuals. \textbf{d}, Cell-type annotation across five cohorts from binary-detection projections versus model embeddings, with rings giving means and 95\% cohort-bootstrap intervals. \textbf{e}, Integrated Gradients (IG) attribution across channel-informative and channel-uninformative controls. Shapes identify baselines and black ticks mark defined intervention-loss identity shares; ratios with a zero denominator are omitted. \textbf{f}, Reproducibility of sparse autoencoder (SAE) atoms across independent fits, plotted as the share exceeding cosine similarity 0.9. \textbf{g}, Recovery of twelve planted modules across feature families under value-informative and value-null conditions (shapes), coloured by family relative to the AUROC 0.8 threshold (dashed line). \textbf{h}, Pathway recovery from gene-similarity baselines using random (filled) versus frequency-matched (open) negatives. \textbf{i}, Layer-wise learned-subspace transfer in trained and randomly initialised networks, with dots marking subspace swaps and ticks marking full-state references (Methods).}
\end{figure}

\begin{figure}[p]
\phantomsection\label{edfig:10}
\centering
\includegraphics[width=\textwidth,height=0.68\textheight,keepaspectratio]{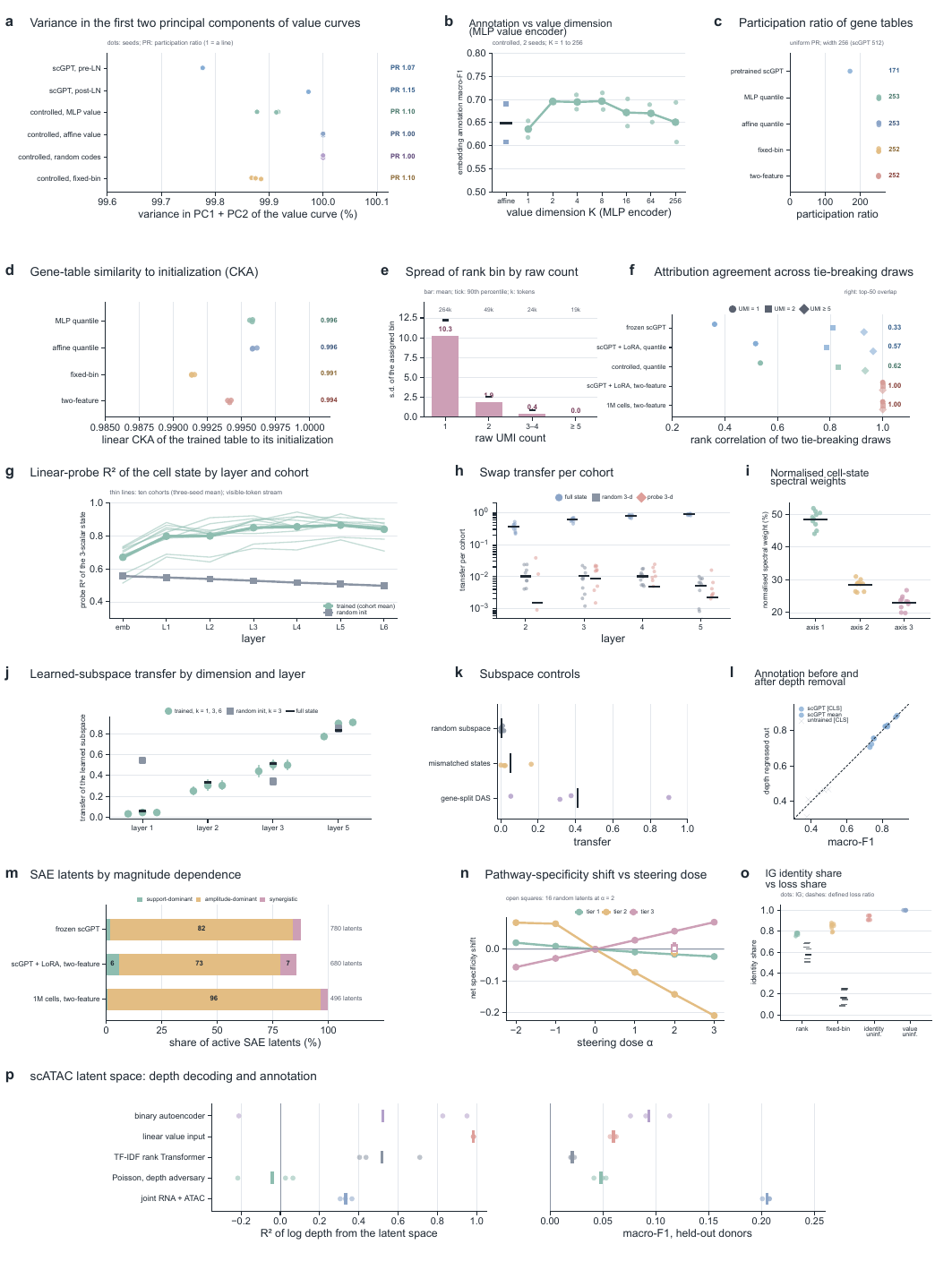}
\vspace{4pt}
\caption*{\footnotesize {\sffamily\bfseries Extended Data Figure 10 \textbar{} Geometry and intervention controls distinguish readable structure from functional use.} \textbf{a}, Scalar value-curve geometry across models, plotted as variance shares and participation ratios (dots). \textbf{b}, MLP value-encoder annotation across dimensions (two runs). \textbf{c}, Gene-table effective rank measured by participation ratio across pretrained, controlled and randomly initialised models (dots). \textbf{d}, Centred kernel alignment of gene tables to initialisation. \textbf{e}, Quantile-bin dispersion across unique molecular identifier (UMI) strata, with means (bars) and ninetieth percentiles (ticks). \textbf{f}, Gradient-times-input attribution agreement across tie-breaking draws by UMI stratum (shapes) and top-50 overlap. \textbf{g}, Layer-wise linear-probe decoding of cell state across cohorts and pooled model summaries. \textbf{h}, Cohort-level output-difference transfer under swap interventions on a logarithmic axis, with ticks marking pooled slopes. \textbf{i}, Normalised cell-state spectral weights across cohorts. \textbf{j}, Distributed alignment search transfer across subspace dimensions and layers, with 95\% bootstrap intervals. \textbf{k}, Subspace-swap controls across conditions (dots) and means (ticks). \textbf{l}, Cohort-level cell-type annotation before and after training-fold depth removal, plotted against the equality diagonal. \textbf{m}, Active sparse autoencoder (SAE) latents partitioned into support-dominant, amplitude-dominant and synergistic tiers. \textbf{n}, Pathway-specificity shifts under latent steering (means $\pm$ standard deviation, 16 random latents, dose two). \textbf{o}, Integrated Gradients (IG) identity shares (dots) and defined intervention-loss shares (ticks); undefined ratios are omitted while IG points remain, as the two measures use different endpoints. \textbf{p}, Depth decoding and annotation in chromatin-accessibility models across three runs (dots) and means (ticks; Methods).}
\end{figure}

Depth variants stacked $L\in\{1, 3, 5\}$ visible-context layers before one readout or self-attention layer. Mask-routing controls removed mask-to-visible or mask-to-mask attention, either post hoc or during retraining (\hyperref[edfig:7]{Extended Data Fig. 7}).

Post-hoc routing interventions removed masked and off-diagonal visible edges from already trained panel models. Retrained controls used the identical attention mask during fitting. The mask-ratio comparison included updated attention, static masked keys and values, retrained sparsity and set pooling. Error differences between set pooling and attention were tabulated jointly by masked-query count and visible-gene count. Cells with fewer than 100 queries were omitted. Further summaries grouped queries by masked-gene count or visible-count quartile, and repeated the comparison for query UMI at least 6 (\hyperref[edfig:7]{Extended Data Fig. 7e--g,j,k}).

Panel white-box fits compared exact alternating least squares with an Adam joint fit across numbers of cell-level scalars. Gain share and distillation agreement used the same panel reference and Transformer outputs. Multiple random initialisations and initialisation from the Adam solution checked sensitivity to starting state. Full-context subsampling removes visible evidence from an existing full-vocabulary task; the sparse positive panel changes the vocabulary and restricts queries to detected genes.

\subsection*{Prediction score and function agreement}

For Transformer $T$, predictor $S$ and reference $P$, distillation $R^2=1-\sum(T-S)^2/\sum(T-P)^2$ over test queries. Gain share is $[\mathbb E(Y-P)^2-\mathbb E(Y-S)^2]/[\mathbb E(Y-P)^2-\mathbb E(Y-T)^2]$. If $T$ and $S$ are conditional expectations under nested information, with $S=\mathbb E[T\mid\text{information of }S]$ and $P$ nested in $S$, squared-error orthogonality makes the two quantities equal. Calibration was checked through gain divided by deviation energy, which equals one for a calibrated predictor. Full-vocabulary retrieval compares score with output agreement.

\subsection*{White box and cell covariates}

The reference Transformer $T$ averaged three controlled full-vocabulary Transformers with learned gene codes and an affine value encoder, evaluated after five epochs of a ten-epoch schedule. Fitting used training donors within each cohort; evaluation used held-out donors. We define the function as the model's function beyond cohort means. The reference $P_g$ was the fit-cell mean bin, shrunk with five pseudo-counts toward the cohort-wide mean.

The low-rank linear white box was $P_g+b_g+\sum_{k=1}^{K}v_{gk}u_{ck}$. Alternating least squares fitted free cell states and gene readouts to Transformer outputs, minimising squared error plus $\lambda(\lVert U\rVert^2+\lVert V\rVert^2)$, with $\lambda=3$, a truncated singular-value initialisation and at most 40 iterations. Linear-kernel ridge regression then mapped visible features to cell state, $u_c=W\phi(c)$. Five-fold cross-validation on fit cells selected ridge penalty from $\{0.1, 1, 10, 10^2, 10^3, 10^4\}$ by query-level prediction error.

Features comprised visible binary detection over the vocabulary, standardised visible bins $(q-25.5)/15$, and four summaries: $\log(1+n_{\text{vis}})$, bin mean, bin standard deviation and fraction of bins at least 40. Test predictions used only these visible features, with cohort results pooling sums of squared errors. Offsets alone corresponded to $K=0$; shuffling permuted test features within cohorts. Reproducibility compared each run with the mean of the other two. Pretrained scGPT used the same fitting procedure on its own visible genes and bins in five donor-disjoint cohorts.

Covariate fits replaced $\phi$ while retaining gene readouts. Covariates were whole-cell log detected-gene count $\log D$, one-hot coarse cell type, visible summaries, log total UMI, mitochondrial fraction and ribosomal fraction. Single covariates used $K=1$ and combinations $K=3$. The share of the white-box increment was $(R_Z^2-R_{K=0}^2)/(R_{\phi,K=3}^2-R_{K=0}^2)$; the fraction of the full function beyond offsets was $(R^2-R_{K=0}^2)/(1-R_{K=0}^2)$.

For enrichment, cohort gene readouts were stacked into three varimax-rotated pooled singular vectors, alongside projections of gene--depth correlation, cell-type effects orthogonal to depth and the remaining direction onto the same span. Scores were tested raw and after regression on mean bin, detection rate, bin standard deviation and log mean count. The 1,801 gene sets included Hallmark, Reactome sets with at least five vocabulary genes and curated quality, stress, cell-cycle and lineage programmes. Weighted running-sum tests used 150 random sets per logarithmic size bin and Benjamini--Hochberg correction within axis.

The free-state control retained fit-cell gene readouts and fitted each test state on half its queries, scoring the other half and reversing the split. This free-state control complemented the visible-input encoder and independent-model agreement. Single-model reliability was the uncentred correlation of independent deviations from $P$, whereas sparse-context reliability used the Spearman--Brown correction $r_3=3r_1/(1 + 2 r_1)$.

\subsection*{Rank arithmetic and state geometry}

Gene depth responses were slopes of Transformer deviations or targets on standardised $\log D$. A multinomial simulation used each cell's total UMI and fit-cohort mean gene proportions, with three draws per cell, 50-bin quantiles, uniform tie-breaking, a 1,024-gene window and 15\% masking. We correlated simulated target slopes with Transformer slopes across cohort--gene pairs. Ridge regression of $\log D$ on visible features measured window information about whole-cell detection.

For state plots, if $w_g$ is gene $g$'s query share and $\sqrt w V=PSQ^\top$, coordinates $SQ^\top u_c$ have length equal to the prediction's root-mean-square change. Rotation aligned the first axis with least-squares $\log D$ and the second with the largest orthogonal direction. Cell-type separation used $\eta^2$. Query-level count, bin, prediction and depth summaries were averaged over depth deciles. Window-ranked controls recomputed targets among the same selected window genes, retaining architecture, masks, order and schedule.

\subsection*{Thinning and objective controls}

Count thinning (binomial thinning of visible counts to one half or one quarter) used 1,500 held-out cells per cohort. Zeroed genes left the input and visible bins were recomputed. Queries and targets stayed fixed. A white box fitted to untouched training cells predicted each response as $v_g^\top W[\phi(c_p)-\phi(c)]$.

Agreement used uncentred response $R^2$ pooled across cohorts, while split-half free states quantified the response fraction within gene-readout space. Thinning jointly altered detected-gene count, library size and bins.

The depth direction was the least-squares map from untouched test states to $\log_{10}D$ within cohort. Its displacement share was summed squared projection divided by total squared displacement. We report the cohort median. The thinned whole-cell detection estimate scaled the original count by the surviving fraction of visible genes. Pretrained scGPT used visible-only quantiles, unchanged queries and matched tie-breaking draws, with its own fit-cell white box and target means.

Objective controls trained the same six-layer width-256 Transformer on 200,000 cells for five epochs with three replicates. Fixed-bin log CP10k targets used full-cell log CP10k assigned to 50 global bins, with edges at the 2nd to 98th percentiles of a 20,000-cell subsample. Visible values used those bins. After thinning, values were rebuilt using visible library size scaled by the retained count share and the same global edges. Auxiliary target readouts used stop-gradient hidden states.

The log CP10k regression Transformer used an affine value encoder rescaled within the visible set and projected two logarithmic features into the classification token: detected-gene count and visible CP10k sum. During pretraining, the detected-gene feature used 0.85 times whole-cell detection, whereas evaluation used visible detection. Separate interventions thinned values with this feature fixed, changed only this feature, or changed both; the CP10k-sum feature tracked the values. Pretraining bins and normalisation otherwise used full-cell summaries before masking. Visible-input controls recomputed rank and fixed-bin inputs after masking during training, retaining targets, masks, order and schedule.

We recovered total UMI as the median of $c\,10^4/(e^y-1)$ across detected genes, where $c$ was raw count and $y$ stored log CP10k. Salivary-gland pseudo-count totals carry little library-size information. We therefore excluded that cohort from comparisons of UMI and detection, reporting it separately. Unique contributions measured the increment from the two-covariate fit over each single-covariate fit, relative to the white-box increment.

scFoundation held out 30\% of detected genes and 3\% of measured zeros, using visible-library log CP10k and two visible-depth tokens. Fits and tests each used 1,500 cells from seven cohorts; depth comparisons excluded salivary gland. Positive-query targets retained full-cell normalisation. Visible features comprised detection, standardised values and log visible library size. White-box, covariate and free-state analyses followed the procedures above; a sensitivity analysis included sampled zeros.

\subsection*{Sparse context, scale and training duration}

Context subsampling retained $n\in\{64, 128, 256\}$ random visible genes and refitted the white box using retained features. An in-distribution control trained three models with a 64-gene visible cap, preserving queries and replacing dropped tokens with padding identities.

Scaling used a 12-layer width-512 Transformer with 8 heads and the same target, window and mask, trained on a 1,000,000-cell corpus combining the 200,000-cell corpus with 29 external datasets (excluding evaluation donors). We compared checkpoints after 7,812 steps (one pass, three runs) and 23,436 steps (three passes) against the smaller corpus at matched steps. One-pass controls annealed at 7,812 steps. A second three-pass run and ten-epoch controlled checkpoints tested training duration. For continual adaptation of pretrained scGPT, we trained the affine value encoder, top four layers and readout for two epochs while freezing lower layers and the gene table.

The one-pass million-cell checkpoint matched five passes over the controlled corpus in optimisation steps. Annealed one-pass controls separated stopping time from final learning rate. Three-pass model agreement compared MLP and affine value encoders, whereas controlled-model agreement compared independent runs of the same configuration (\hyperref[fig:3]{Fig. 3h} and \hyperref[edfig:7]{Extended Data Fig. 7a}).

\subsection*{Interpretability readouts and controls}

Integrated Gradients {[}\hyperlink{ref:50}{50}{]} in the primary additive-embedding and synthetic audits used 24 midpoint integration steps; the additional channel-control runs used 16. Baselines were zero vectors, a mean identity embedding over the evaluation batch with a mid-range value embedding, and independently shuffled visible identities and values. The scalar endpoint was the sum of masked predictions. Identity share was visible-identity $L_1$ attribution divided by visible identity-plus-value $L_1$ attribution; query-identity attribution was stored separately. Added identity and value vectors receive the same path gradient, multiplied by their respective displacements from baseline. Channel-uninformative controls randomised identities on every forward pass or clamped values before encoding. IG therefore integrated along embedding paths that the discrete preprocessing never produces. Separate step-sensitivity analyses used the settings in Supplementary Table 5.

Because within-cell identity permutation and value shuffling are inverse permutations, their expected losses coincide for order-invariant readouts and their loss ratio cannot identify channel use. Figure 5d reports the identity-permutation versus clamped-input loss ratio, $s_{\rm loss}=\Delta L_{\rm id}^{+}/(\Delta L_{\rm id}^{+}+\Delta L_{\rm constval}^{+})$, where $x^+=\max(x,0)$. Ratios whose denominator was zero at float32 precision were omitted. Step-sensitivity checks are in Supplementary Note 3.

Embedding analyses used ten donor-disjoint cohorts for controlled models and five for scGPT. Depth was a polynomial in log detection and log UMI; annotation used standardised linear probes. For scGPT, five-fold cross-fitting estimated covariate-explained variance at every layer, and donor-grouped ridge classification assessed annotation before and after training-fold depth removal.

Deletion analyses used 19,200 controlled-model deletions and 4,000 scGPT deletions, trained and untrained, with classification-token and mean pooling. Nuisance regression predicted embedding-shift norm from log gene frequency, bin and visible-gene count. Attention over 2,400 cells and the 3,000 most expressed genes was decomposed into node marginals and residuals, $\bar A_{ij}=m_i+m_j+R_{ij}$. TRRUST, a curated database of human transcriptional regulatory interactions, supplied 715 positive pairs and frequency-matched negatives. We reported head-averaged and per-head area under the receiver operating characteristic curve (AUROC).

The real-data TopK sparse autoencoders used 1,024 features, 32 active features and two fits per layer, compared with 1,024 random unit directions passed through a rectified linear unit. Controlled fits used 2,500 Adam steps and scGPT fits 2,000, batch size 4,096 and learning rate 0.001. We centred inputs per channel and divided them by a single global standard deviation. Decoder atoms were renormalised after every step. The synthetic-module audit instead used 512 features, 16 active features and one 2,000-step fit per model/layer, with matched 512 random directions. Gene-table controls included annotation-degree sum, gene frequency, five gene-level scalars and positive pointwise-mutual-information factorisation of binary co-occurrence. Supplementary Table 3 gives conditions and readouts; \hyperref[edfig:9]{Extended Data Figs. 9} and 10 show probe, swap, dictionary and attribution controls.

Swap interventions (interchange interventions) compared swapping the full visible-token state with swaps restricted to a random subspace or the linear probe directions for the white-box cell state, across layers 2--5. Output-difference transfer was compared with linear decoding accuracy, using the same target cell state. We assessed learned subspaces in trained and randomly initialised networks and evaluated distributed alignment search at dimensions $k=1,3,6$ across layers 1, 2, 3 and 5. Random-subspace, mismatched-state and gene-split evaluations tested transfer specificity, with the last scored on held-out genes (\hyperref[fig:5]{Fig. 5c} and \hyperref[edfig:9]{Extended Data Figs. 9b,i} and 10j,k). These controls distinguish reading a state linearly from replacing it along particular directions.

Dictionary comparisons evaluated controlled layers 3 and 6, pretrained scGPT layers 6 and 12, and untrained reference layers. Feature enrichment tested top genes against GO and Reactome terms alongside frequency- and annotation-degree-matched shuffles. Decoder-atom reproducibility used cosine > 0.9 across independent fits. On the synthetic benchmark, feature recovery was the median across planted modules of the best single-feature AUROC; trained and untrained models were compared on value-informative and value-null data, with an AUROC 0.8 recovery threshold (\hyperref[edfig:9]{Extended Data Fig. 9g}).

Counterfactual dictionary readouts grouped active latents into support-dominant, amplitude-dominant and synergistic tiers. The evaluated models were pretrained scGPT, two-feature encoder scGPT with low-rank adaptation and the million-cell two-feature encoder model. Dose-response plots compared latent steering with mean $\pm$ s.d. for 16 random latents at $\alpha=2$ (\hyperref[edfig:10]{Extended Data Fig. 10m,n}). Gradient-times-input attribution stability compared two tie-breaking draws using Spearman correlation within UMI strata 1, 2 and at least 5, alongside top-50 Jaccard overlap (\hyperref[edfig:10]{Extended Data Fig. 10f}). These are separate evaluations from the synthetic-module recovery test.

The synthetic benchmark generated negative-binomial counts for 600 genes across six cell types, with 40 transcription factors and 12 modules of 40 non-transcription-factor genes. In the value-null condition, detection depended on activity, whereas positive counts depended only on gene baseline and depth. Models used 100,000 training cells, 6,000 test cells, two independent runs and five epochs. These conditions supplied planted regulators and modules for deletion and dictionary controls. Supplementary Note 4 specifies the generator distributions and coefficients.

The enhancer--promoter benchmark used the EngreitzLab CRISPR\_comparison EPCrisprBenchmark GRCh38 tables: 10,356 K562 pairs and 4,378 pairs from five held-out cell types, totalling 14,734 valid pairs with 661 positives. Labels were {\small\texttt{Regulated}}; rows required valid connections, finite coordinates and autosomal or sex-chromosome membership. Distance was the supplied {\small\texttt{distanceToTSS}}. We scored inverse distance and ChromFound token cosine by AUROC overall and after distance matching. Within each source panel, positives in a fixed random order were matched to up to five unused negatives within 0.05 log10-distance units (distance floored at 100 bp), falling back to 0.12 when no candidate existed. Positives without candidates were omitted; negatives were not reused. Supplementary Table 3 reports point estimates across all evaluated scoring rules.

\subsection*{Statistical analysis}

Donor-level annotation and CITE-seq contrasts used 10,000 stratified donor-bootstrap draws and paired two-sided sign-flip tests. Geneformer-V2 and State summary retentions and four-cohort Perturb-seq means were point estimates; scGPT and scFoundation donor-level comparisons had paired intervals. Multi-run retraining reports arithmetic mean and standard deviation of individual risks. Panel contrasts used 4,000 two-level run-by-cell bootstrap draws.

Unless specified otherwise, function-analysis intervals were 95\% percentile intervals from 4,000 cohort resamples, recomputing pooled values with deviation-energy weights and pairing cohorts for contrasts. Supplementary Note 2 separates assumption-free tie-breaking variance from model-dependent Poisson estimates.

\subsection*{Data availability}

The single-cell and single-nucleus RNA-sequencing tissue cohorts were obtained from CZ CELLxGENE Discover {[}\hyperlink{ref:31}{31}{]}; dataset and collection identifiers, portal release dates and cohort definitions are listed in Supplementary Table 1A. CITE-seq data are available from NCBI Gene Expression Omnibus (GEO) under accessions GSE194122, GSE272583, GSE316096 and GSE331003 (Supplementary Table 1B). The four Perturb-seq cohorts were obtained from the published Replogle et al. {[}\hyperlink{ref:29}{29}{]} and Nadig et al. {[}\hyperlink{ref:30}{30}{]} resources (Supplementary Table 1C). Chromatin-accessibility and paired multiome data are available from GEO under accessions GSE194122, GSE312668, GSE331288 and GSE156478 (Supplementary Table 1D). The enhancer--promoter benchmark used the GRCh38 training-K562 and held-out-five-cell-type tables from the EngreitzLab CRISPR\_comparison EPCrisprBenchmark resource, as specified in Methods.

Processed summary tables underlying \hyperref[fig:1]{Figs. 1}--5 and \hyperref[edfig:1]{Extended Data Figs. 1}--10 are provided in the Supplementary Tables and the public repository specified below. Original raw datasets remain available from their respective repositories.

\subsection*{Code availability}

Code for model training, evaluation, statistical analysis, synthetic-benchmark generation and figure reproduction is available under the MIT License at \href{https://github.com/zhang-lecheng/scfm-value-dimensions}{{\small\texttt{https:/\allowbreak{}/\allowbreak{}github.\allowbreak{}com/\allowbreak{}zhang-\allowbreak{}lecheng/\allowbreak{}scfm-\allowbreak{}value-\allowbreak{}dimensions}}}.

\subsection*{Acknowledgements}

This work was supported by the National Science and Technology Innovation -- Major Program (No. 2022ZD0115101), the National Science and Technology Major Project of China (No. 2021YFA1301603), the SOE Dean Special Project Fund (SOE-DSPF) Program of Westlake University (Project No. WU2025B006), and the Center of Synthetic Biology and Integrated Bioengineering of Westlake University (Project No. WU2024A004).

\subsection*{Author contributions}

L.Z. and Y.Y. contributed equally.

L.Z. conceived the study, designed and executed the experiments, analysed the data and wrote the initial draft of the manuscript.

Y.Y. contributed to experimental evaluation, technical validation and manuscript revision.

S.Z.L. supervised the project, provided guidance and revised the manuscript. All authors approved the final manuscript.

\subsection*{Competing interests}

The authors declare no competing interests.

\clearpage

\section*{References}

{\small\noindent\hypertarget{ref:1}{}\hangindent=1.8em\hangafter=1\textbf{1.}\enspace Cui, H. et al. scGPT: toward building a foundation model for single-cell multi-omics using generative AI. \textit{Nat. Methods} \textbf{21}, 1470--1480 (2024). \href{https://doi.org/10.1038/s41592-024-02201-0}{{\small\texttt{https:/\allowbreak{}/\allowbreak{}doi.\allowbreak{}org/\allowbreak{}10.\allowbreak{}1038/\allowbreak{}s41592-\allowbreak{}024-\allowbreak{}02201-\allowbreak{}0}}}\par\vspace{2pt}
\noindent\hypertarget{ref:2}{}\hangindent=1.8em\hangafter=1\textbf{2.}\enspace Hao, M. et al. Large-scale foundation model on single-cell transcriptomics. \textit{Nat. Methods} \textbf{21}, 1481--1491 (2024). \href{https://doi.org/10.1038/s41592-024-02305-7}{{\small\texttt{https:/\allowbreak{}/\allowbreak{}doi.\allowbreak{}org/\allowbreak{}10.\allowbreak{}1038/\allowbreak{}s41592-\allowbreak{}024-\allowbreak{}02305-\allowbreak{}7}}}\par\vspace{2pt}
\noindent\hypertarget{ref:3}{}\hangindent=1.8em\hangafter=1\textbf{3.}\enspace Theodoris, C. V. et al. Transfer learning enables predictions in network biology. \textit{Nature} \textbf{618}, 616--624 (2023). \href{https://doi.org/10.1038/s41586-023-06139-9}{{\small\texttt{https:/\allowbreak{}/\allowbreak{}doi.\allowbreak{}org/\allowbreak{}10.\allowbreak{}1038/\allowbreak{}s41586-\allowbreak{}023-\allowbreak{}06139-\allowbreak{}9}}}\par\vspace{2pt}
\noindent\hypertarget{ref:4}{}\hangindent=1.8em\hangafter=1\textbf{4.}\enspace Chen, H. et al. Quantized multi-task learning for context-specific representations of gene network dynamics. Preprint at \textit{bioRxiv} \href{https://doi.org/10.1101/2024.08.16.608180}{{\small\texttt{https:/\allowbreak{}/\allowbreak{}doi.\allowbreak{}org/\allowbreak{}10.\allowbreak{}1101/\allowbreak{}2024.\allowbreak{}08.\allowbreak{}16.\allowbreak{}608180}}} (2024).\par\vspace{2pt}
\noindent\hypertarget{ref:5}{}\hangindent=1.8em\hangafter=1\textbf{5.}\enspace Adduri, A. K. et al. Predicting cellular responses to perturbation across diverse contexts with State. \textit{Cell} \textbf{189}, 5914--5931.e20 (2026). \href{https://doi.org/10.1016/j.cell.2026.07.052}{{\small\texttt{https:/\allowbreak{}/\allowbreak{}doi.\allowbreak{}org/\allowbreak{}10.\allowbreak{}1016/\allowbreak{}j.\allowbreak{}cell.\allowbreak{}2026.\allowbreak{}07.\allowbreak{}052}}}\par\vspace{2pt}
\noindent\hypertarget{ref:6}{}\hangindent=1.8em\hangafter=1\textbf{6.}\enspace Boiarsky, R. et al. Deeper evaluation of a single-cell foundation model. \textit{Nat. Mach. Intell.} \textbf{6}, 1443--1446 (2024). \href{https://doi.org/10.1038/s42256-024-00949-w}{{\small\texttt{https:/\allowbreak{}/\allowbreak{}doi.\allowbreak{}org/\allowbreak{}10.\allowbreak{}1038/\allowbreak{}s42256-\allowbreak{}024-\allowbreak{}00949-\allowbreak{}w}}}\par\vspace{2pt}
\noindent\hypertarget{ref:7}{}\hangindent=1.8em\hangafter=1\textbf{7.}\enspace Kedzierska, K. Z., Crawford, L., Amini, A. P. \& Lu, A. X. Zero-shot evaluation reveals limitations of single-cell foundation models. \textit{Genome Biol.} \textbf{26}, 101 (2025). \href{https://doi.org/10.1186/s13059-025-03574-x}{{\small\texttt{https:/\allowbreak{}/\allowbreak{}doi.\allowbreak{}org/\allowbreak{}10.\allowbreak{}1186/\allowbreak{}s13059-\allowbreak{}025-\allowbreak{}03574-\allowbreak{}x}}}\par\vspace{2pt}
\noindent\hypertarget{ref:8}{}\hangindent=1.8em\hangafter=1\textbf{8.}\enspace Wu, J. et al. Biology-driven insights into the power of single-cell foundation models. \textit{Genome Biol.} \textbf{26}, 334 (2025). \href{https://doi.org/10.1186/s13059-025-03781-6}{{\small\texttt{https:/\allowbreak{}/\allowbreak{}doi.\allowbreak{}org/\allowbreak{}10.\allowbreak{}1186/\allowbreak{}s13059-\allowbreak{}025-\allowbreak{}03781-\allowbreak{}6}}}\par\vspace{2pt}
\noindent\hypertarget{ref:9}{}\hangindent=1.8em\hangafter=1\textbf{9.}\enspace Ahlmann-Eltze, C., Huber, W. \& Anders, S. Deep-learning-based gene perturbation effect prediction does not yet outperform simple linear baselines. \textit{Nat. Methods} \textbf{22}, 1657--1661 (2025). \href{https://doi.org/10.1038/s41592-025-02772-6}{{\small\texttt{https:/\allowbreak{}/\allowbreak{}doi.\allowbreak{}org/\allowbreak{}10.\allowbreak{}1038/\allowbreak{}s41592-\allowbreak{}025-\allowbreak{}02772-\allowbreak{}6}}}\par\vspace{2pt}
\noindent\hypertarget{ref:10}{}\hangindent=1.8em\hangafter=1\textbf{10.}\enspace Miller, H. E. et al. Deep learning perturbation models can outperform baselines on calibrated metrics. \textit{Nat. Biotechnol.} (2026). \href{https://doi.org/10.1038/s41587-026-03307-w}{{\small\texttt{https:/\allowbreak{}/\allowbreak{}doi.\allowbreak{}org/\allowbreak{}10.\allowbreak{}1038/\allowbreak{}s41587-\allowbreak{}026-\allowbreak{}03307-\allowbreak{}w}}}\par\vspace{2pt}
\noindent\hypertarget{ref:11}{}\hangindent=1.8em\hangafter=1\textbf{11.}\enspace Vollenweider, M. \& Bühlmann, P. Signal, Bounds, and Baselines: Principles for Evaluating Virtual Cell Perturbation Models. Preprint at \textit{bioRxiv} \href{https://doi.org/10.64898/2026.04.20.719650}{{\small\texttt{https:/\allowbreak{}/\allowbreak{}doi.\allowbreak{}org/\allowbreak{}10.\allowbreak{}64898/\allowbreak{}2026.\allowbreak{}04.\allowbreak{}20.\allowbreak{}719650}}} (2026).\par\vspace{2pt}
\noindent\hypertarget{ref:12}{}\hangindent=1.8em\hangafter=1\textbf{12.}\enspace Wang, X., Kuipers, J., Hugi, F., Platt, R. J. \& Beerenwinkel, N. Reliable single-cell perturbations explain and improve model performance. Preprint at \textit{bioRxiv} \href{https://doi.org/10.64898/2026.08.11.744177}{{\small\texttt{https:/\allowbreak{}/\allowbreak{}doi.\allowbreak{}org/\allowbreak{}10.\allowbreak{}64898/\allowbreak{}2026.\allowbreak{}08.\allowbreak{}11.\allowbreak{}744177}}} (2026).\par\vspace{2pt}
\noindent\hypertarget{ref:13}{}\hangindent=1.8em\hangafter=1\textbf{13.}\enspace Qiu, P. Embracing the dropouts in single-cell RNA-seq analysis. \textit{Nat. Commun.} \textbf{11}, 1169 (2020). \href{https://doi.org/10.1038/s41467-020-14976-9}{{\small\texttt{https:/\allowbreak{}/\allowbreak{}doi.\allowbreak{}org/\allowbreak{}10.\allowbreak{}1038/\allowbreak{}s41467-\allowbreak{}020-\allowbreak{}14976-\allowbreak{}9}}}\par\vspace{2pt}
\noindent\hypertarget{ref:14}{}\hangindent=1.8em\hangafter=1\textbf{14.}\enspace Bouland, G. A., Mahfouz, A. \& Reinders, M. J. T. Consequences and opportunities arising due to sparser single-cell RNA-seq datasets. \textit{Genome Biol.} \textbf{24}, 86 (2023). \href{https://doi.org/10.1186/s13059-023-02933-w}{{\small\texttt{https:/\allowbreak{}/\allowbreak{}doi.\allowbreak{}org/\allowbreak{}10.\allowbreak{}1186/\allowbreak{}s13059-\allowbreak{}023-\allowbreak{}02933-\allowbreak{}w}}}\par\vspace{2pt}
\noindent\hypertarget{ref:15}{}\hangindent=1.8em\hangafter=1\textbf{15.}\enspace Sadalski, I. When Task-Specific Learning Outperforms Transfer Learning: A Benchmark of Gene and Expression Encoding Strategies. Preprint at \textit{bioRxiv} \href{https://doi.org/10.64898/2025.12.05.690830}{{\small\texttt{https:/\allowbreak{}/\allowbreak{}doi.\allowbreak{}org/\allowbreak{}10.\allowbreak{}64898/\allowbreak{}2025.\allowbreak{}12.\allowbreak{}05.\allowbreak{}690830}}} (2025).\par\vspace{2pt}
\noindent\hypertarget{ref:16}{}\hangindent=1.8em\hangafter=1\textbf{16.}\enspace Schlede, S. et al. Raw-count embeddings improve single-cell foundation models. Preprint at \textit{bioRxiv} \href{https://doi.org/10.64898/2026.06.29.735389}{{\small\texttt{https:/\allowbreak{}/\allowbreak{}doi.\allowbreak{}org/\allowbreak{}10.\allowbreak{}64898/\allowbreak{}2026.\allowbreak{}06.\allowbreak{}29.\allowbreak{}735389}}} (2026).\par\vspace{2pt}
\noindent\hypertarget{ref:17}{}\hangindent=1.8em\hangafter=1\textbf{17.}\enspace DenAdel, A. et al. Evaluating the role of pretraining dataset size and diversity on single-cell foundation model performance. \textit{Nat. Methods} \textbf{23}, 1447--1457 (2026). \href{https://doi.org/10.1038/s41592-026-03120-y}{{\small\texttt{https:/\allowbreak{}/\allowbreak{}doi.\allowbreak{}org/\allowbreak{}10.\allowbreak{}1038/\allowbreak{}s41592-\allowbreak{}026-\allowbreak{}03120-\allowbreak{}y}}}\par\vspace{2pt}
\noindent\hypertarget{ref:18}{}\hangindent=1.8em\hangafter=1\textbf{18.}\enspace Verma, T. S. \& Pearl, J. Equivalence and synthesis of causal models. In \textit{Proceedings of the Sixth Conference on Uncertainty in Artificial Intelligence (UAI 1990)}, 220--227 (1990).\par\vspace{2pt}
\noindent\hypertarget{ref:19}{}\hangindent=1.8em\hangafter=1\textbf{19.}\enspace Chickering, D. M. Optimal structure identification with greedy search. \textit{J. Mach. Learn. Res.} \textbf{3}, 507--554 (2002).\par\vspace{2pt}
\noindent\hypertarget{ref:20}{}\hangindent=1.8em\hangafter=1\textbf{20.}\enspace Shimizu, S., Hoyer, P. O., Hyvärinen, A. \& Kerminen, A. A linear non-Gaussian acyclic model for causal discovery. \textit{J. Mach. Learn. Res.} \textbf{7}, 2003--2030 (2006).\par\vspace{2pt}
\noindent\hypertarget{ref:21}{}\hangindent=1.8em\hangafter=1\textbf{21.}\enspace Moriel, N., Shimoni, Y., Rosen-Zvi, M. \& Danziger, M. M. Toward a Coherent Virtual Cell Model: Probing Biological World-Model Coherence in Transcriptomic Foundation Models. In \textit{NeurIPS 2025 Workshop on AI Virtual Cells and Instruments (AI4D3)} \href{https://openreview.net/forum?id=nLy3iIONkt}{{\small\texttt{https:/\allowbreak{}/\allowbreak{}openreview.\allowbreak{}net/\allowbreak{}forum?\allowbreak{}id=\allowbreak{}nLy3iIONkt}}} (2025).\par\vspace{2pt}
\noindent\hypertarget{ref:22}{}\hangindent=1.8em\hangafter=1\textbf{22.}\enspace Kendiukhov, I. Systematic evaluation of single-cell foundation model interpretability: attention-derived edge scores add no incremental value over gene-level features for perturbation-target prediction. \textit{BMC Genomics} \textbf{27}, 634 (2026). \href{https://doi.org/10.1186/s12864-026-12965-8}{{\small\texttt{https:/\allowbreak{}/\allowbreak{}doi.\allowbreak{}org/\allowbreak{}10.\allowbreak{}1186/\allowbreak{}s12864-\allowbreak{}026-\allowbreak{}12965-\allowbreak{}8}}}\par\vspace{2pt}
\noindent\hypertarget{ref:23}{}\hangindent=1.8em\hangafter=1\textbf{23.}\enspace Kendiukhov, I. Sparse autoencoders reveal interpretable cell-type programs in single-cell foundation model representations. \textit{J. Biomed. Inform.} \textbf{180}, 105056 (2026). \href{https://doi.org/10.1016/j.jbi.2026.105056}{{\small\texttt{https:/\allowbreak{}/\allowbreak{}doi.\allowbreak{}org/\allowbreak{}10.\allowbreak{}1016/\allowbreak{}j.\allowbreak{}jbi.\allowbreak{}2026.\allowbreak{}105056}}}\par\vspace{2pt}
\noindent\hypertarget{ref:24}{}\hangindent=1.8em\hangafter=1\textbf{24.}\enspace Kendiukhov, I. Discovery of a Hematopoietic Manifold in scGPT Yields a Method for Extracting Performant Algorithms from Biological Foundation Model Internals. Preprint at \textit{arXiv} \href{https://arxiv.org/abs/2603.10261}{{\small\texttt{https:/\allowbreak{}/\allowbreak{}arxiv.\allowbreak{}org/\allowbreak{}abs/\allowbreak{}2603.\allowbreak{}10261}}} (2026).\par\vspace{2pt}
\noindent\hypertarget{ref:25}{}\hangindent=1.8em\hangafter=1\textbf{25.}\enspace Saxe, A. M., McClelland, J. L. \& Ganguli, S. Exact solutions to the nonlinear dynamics of learning in deep linear neural networks. In \textit{International Conference on Learning Representations} (2014). \href{https://arxiv.org/abs/1312.6120}{{\small\texttt{https:/\allowbreak{}/\allowbreak{}arxiv.\allowbreak{}org/\allowbreak{}abs/\allowbreak{}1312.\allowbreak{}6120}}}\par\vspace{2pt}
\noindent\hypertarget{ref:26}{}\hangindent=1.8em\hangafter=1\textbf{26.}\enspace Gidel, G., Bach, F. \& Lacoste-Julien, S. Implicit regularization of discrete gradient dynamics in linear neural networks. \textit{Adv. Neural Inf. Process. Syst.} \textbf{32}, 3196--3206 (2019).\par\vspace{2pt}
\noindent\hypertarget{ref:27}{}\hangindent=1.8em\hangafter=1\textbf{27.}\enspace Boix-Adserà, E. et al. Transformers learn through gradual rank increase. \textit{Adv. Neural Inf. Process. Syst.} \textbf{36}, 24519--24551 (2023). \href{https://doi.org/10.52202/075280-1066}{{\small\texttt{https:/\allowbreak{}/\allowbreak{}doi.\allowbreak{}org/\allowbreak{}10.\allowbreak{}52202/\allowbreak{}075280-\allowbreak{}1066}}}\par\vspace{2pt}
\noindent\hypertarget{ref:28}{}\hangindent=1.8em\hangafter=1\textbf{28.}\enspace Wang, Y. et al. Graph-structured joint-embedding prediction for single-cell transcriptomics. Preprint at \textit{arXiv} \href{https://arxiv.org/abs/2608.05928}{{\small\texttt{https:/\allowbreak{}/\allowbreak{}arxiv.\allowbreak{}org/\allowbreak{}abs/\allowbreak{}2608.\allowbreak{}05928}}} (2026).\par\vspace{2pt}
\noindent\hypertarget{ref:29}{}\hangindent=1.8em\hangafter=1\textbf{29.}\enspace Replogle, J. M. et al. Mapping information-rich genotype-phenotype landscapes with genome-scale Perturb-seq. \textit{Cell} \textbf{185}, 2559--2575.e28 (2022). \href{https://doi.org/10.1016/j.cell.2022.05.013}{{\small\texttt{https:/\allowbreak{}/\allowbreak{}doi.\allowbreak{}org/\allowbreak{}10.\allowbreak{}1016/\allowbreak{}j.\allowbreak{}cell.\allowbreak{}2022.\allowbreak{}05.\allowbreak{}013}}}\par\vspace{2pt}
\noindent\hypertarget{ref:30}{}\hangindent=1.8em\hangafter=1\textbf{30.}\enspace Nadig, A. et al. Transcriptome-wide analysis of differential expression in perturbation atlases. \textit{Nat. Genet.} \textbf{57}, 1228--1237 (2025). \href{https://doi.org/10.1038/s41588-025-02169-3}{{\small\texttt{https:/\allowbreak{}/\allowbreak{}doi.\allowbreak{}org/\allowbreak{}10.\allowbreak{}1038/\allowbreak{}s41588-\allowbreak{}025-\allowbreak{}02169-\allowbreak{}3}}}\par\vspace{2pt}
\noindent\hypertarget{ref:31}{}\hangindent=1.8em\hangafter=1\textbf{31.}\enspace CZI Cell Science Program et al. CZ CELLxGENE Discover: a single-cell data platform for scalable exploration, analysis and modeling of aggregated data. \textit{Nucleic Acids Res.} \textbf{53}, D886--D900 (2025). \href{https://doi.org/10.1093/nar/gkae1142}{{\small\texttt{https:/\allowbreak{}/\allowbreak{}doi.\allowbreak{}org/\allowbreak{}10.\allowbreak{}1093/\allowbreak{}nar/\allowbreak{}gkae1142}}}\par\vspace{2pt}
\noindent\hypertarget{ref:32}{}\hangindent=1.8em\hangafter=1\textbf{32.}\enspace Ivanova, E. N. et al. mRNA COVID-19 vaccine elicits potent adaptive immune response without the acute inflammation of SARS-CoV-2 infection. \textit{iScience} \textbf{26}, 108572 (2023). \href{https://doi.org/10.1016/j.isci.2023.108572}{{\small\texttt{https:/\allowbreak{}/\allowbreak{}doi.\allowbreak{}org/\allowbreak{}10.\allowbreak{}1016/\allowbreak{}j.\allowbreak{}isci.\allowbreak{}2023.\allowbreak{}108572}}}\par\vspace{2pt}
\noindent\hypertarget{ref:33}{}\hangindent=1.8em\hangafter=1\textbf{33.}\enspace Sikkema, L. et al. An integrated cell atlas of the lung in health and disease. \textit{Nat. Med.} \textbf{29}, 1563--1577 (2023). \href{https://doi.org/10.1038/s41591-023-02327-2}{{\small\texttt{https:/\allowbreak{}/\allowbreak{}doi.\allowbreak{}org/\allowbreak{}10.\allowbreak{}1038/\allowbreak{}s41591-\allowbreak{}023-\allowbreak{}02327-\allowbreak{}2}}}\par\vspace{2pt}
\noindent\hypertarget{ref:34}{}\hangindent=1.8em\hangafter=1\textbf{34.}\enspace Efthymiou, V. et al. Single-nucleus analysis of human white adipose tissue reveals adipocyte subsets with distinct metabolic profiles. \textit{Nat. Commun.} \textbf{17}, 9471 (2026). \href{https://doi.org/10.1038/s41467-026-75426-6}{{\small\texttt{https:/\allowbreak{}/\allowbreak{}doi.\allowbreak{}org/\allowbreak{}10.\allowbreak{}1038/\allowbreak{}s41467-\allowbreak{}026-\allowbreak{}75426-\allowbreak{}6}}}\par\vspace{2pt}
\noindent\hypertarget{ref:35}{}\hangindent=1.8em\hangafter=1\textbf{35.}\enspace Jenkins, B. H. et al. Single cell and spatial analysis of immune-hot and immune-cold tumours identifies fibroblast subtypes associated with distinct immunological niches and positive immunotherapy response. \textit{Mol. Cancer} \textbf{24}, 3 (2025). \href{https://doi.org/10.1186/s12943-024-02191-9}{{\small\texttt{https:/\allowbreak{}/\allowbreak{}doi.\allowbreak{}org/\allowbreak{}10.\allowbreak{}1186/\allowbreak{}s12943-\allowbreak{}024-\allowbreak{}02191-\allowbreak{}9}}}\par\vspace{2pt}
\noindent\hypertarget{ref:36}{}\hangindent=1.8em\hangafter=1\textbf{36.}\enspace Pranzatelli, T. J. F. et al. GZMK+CD8+ T cells target a specific acinar cell type in Sjögren's disease. \textit{Ann. Rheum. Dis.} \textbf{85}, 880--898 (2026). \href{https://doi.org/10.1016/j.ard.2025.08.014}{{\small\texttt{https:/\allowbreak{}/\allowbreak{}doi.\allowbreak{}org/\allowbreak{}10.\allowbreak{}1016/\allowbreak{}j.\allowbreak{}ard.\allowbreak{}2025.\allowbreak{}08.\allowbreak{}014}}}\par\vspace{2pt}
\noindent\hypertarget{ref:37}{}\hangindent=1.8em\hangafter=1\textbf{37.}\enspace Zheng, H. B. et al. Concerted changes in the pediatric single-cell intestinal ecosystem before and after anti-TNF blockade. \textit{eLife} \textbf{12}, RP91792 (2026). \href{https://doi.org/10.7554/eLife.91792}{{\small\texttt{https:/\allowbreak{}/\allowbreak{}doi.\allowbreak{}org/\allowbreak{}10.\allowbreak{}7554/\allowbreak{}eLife.\allowbreak{}91792}}}\par\vspace{2pt}
\noindent\hypertarget{ref:38}{}\hangindent=1.8em\hangafter=1\textbf{38.}\enspace Wu, S. Z. et al. A single-cell and spatially resolved atlas of human breast cancers. \textit{Nat. Genet.} \textbf{53}, 1334--1347 (2021). \href{https://doi.org/10.1038/s41588-021-00911-1}{{\small\texttt{https:/\allowbreak{}/\allowbreak{}doi.\allowbreak{}org/\allowbreak{}10.\allowbreak{}1038/\allowbreak{}s41588-\allowbreak{}021-\allowbreak{}00911-\allowbreak{}1}}}\par\vspace{2pt}
\noindent\hypertarget{ref:39}{}\hangindent=1.8em\hangafter=1\textbf{39.}\enspace Chan Zuckerberg Initiative Single-Cell COVID-19 Consortia. Single cell profiling of COVID-19 patients: an international data resource from multiple tissues. Preprint at \textit{medRxiv} \href{https://doi.org/10.1101/2020.11.20.20227355}{{\small\texttt{https:/\allowbreak{}/\allowbreak{}doi.\allowbreak{}org/\allowbreak{}10.\allowbreak{}1101/\allowbreak{}2020.\allowbreak{}11.\allowbreak{}20.\allowbreak{}20227355}}} (2020).\par\vspace{2pt}
\noindent\hypertarget{ref:40}{}\hangindent=1.8em\hangafter=1\textbf{40.}\enspace Kriukov, E. et al. Unraveling the developmental heterogeneity of human retinal ganglion cells within the developing retina to study the continuity of maturation. \textit{Dev. Biol.} \textbf{528}, 91--104 (2025). \href{https://doi.org/10.1016/j.ydbio.2025.08.025}{{\small\texttt{https:/\allowbreak{}/\allowbreak{}doi.\allowbreak{}org/\allowbreak{}10.\allowbreak{}1016/\allowbreak{}j.\allowbreak{}ydbio.\allowbreak{}2025.\allowbreak{}08.\allowbreak{}025}}}\par\vspace{2pt}
\noindent\hypertarget{ref:41}{}\hangindent=1.8em\hangafter=1\textbf{41.}\enspace Nieto, P. et al. Integrative CSF profiling identifies disease-specific immune responses in leptomeningeal disease. \textit{Cell Rep. Med.} \textbf{7}, 102651 (2026). \href{https://doi.org/10.1016/j.xcrm.2026.102651}{{\small\texttt{https:/\allowbreak{}/\allowbreak{}doi.\allowbreak{}org/\allowbreak{}10.\allowbreak{}1016/\allowbreak{}j.\allowbreak{}xcrm.\allowbreak{}2026.\allowbreak{}102651}}}\par\vspace{2pt}
\noindent\hypertarget{ref:42}{}\hangindent=1.8em\hangafter=1\textbf{42.}\enspace Stoeckius, M. et al. Simultaneous epitope and transcriptome measurement in single cells. \textit{Nat. Methods} \textbf{14}, 865--868 (2017). \href{https://doi.org/10.1038/nmeth.4380}{{\small\texttt{https:/\allowbreak{}/\allowbreak{}doi.\allowbreak{}org/\allowbreak{}10.\allowbreak{}1038/\allowbreak{}nmeth.\allowbreak{}4380}}}\par\vspace{2pt}
\noindent\hypertarget{ref:43}{}\hangindent=1.8em\hangafter=1\textbf{43.}\enspace Johnson, L. A. et al. Clonal expansion of cytotoxic CD8+ T cells in lecanemab-associated ARIA. \textit{Nat. Commun.} \textbf{17}, 2180 (2026). \href{https://doi.org/10.1038/s41467-026-68921-3}{{\small\texttt{https:/\allowbreak{}/\allowbreak{}doi.\allowbreak{}org/\allowbreak{}10.\allowbreak{}1038/\allowbreak{}s41467-\allowbreak{}026-\allowbreak{}68921-\allowbreak{}3}}}\par\vspace{2pt}
\noindent\hypertarget{ref:44}{}\hangindent=1.8em\hangafter=1\textbf{44.}\enspace Lin, Z. et al. Evolutionary-scale prediction of atomic-level protein structure with a language model. \textit{Science} \textbf{379}, 1123--1130 (2023). \href{https://doi.org/10.1126/science.ade2574}{{\small\texttt{https:/\allowbreak{}/\allowbreak{}doi.\allowbreak{}org/\allowbreak{}10.\allowbreak{}1126/\allowbreak{}science.\allowbreak{}ade2574}}}\par\vspace{2pt}
\noindent\hypertarget{ref:45}{}\hangindent=1.8em\hangafter=1\textbf{45.}\enspace Kumar, A., Raghunathan, A., Jones, R., Ma, T. \& Liang, P. Fine-tuning can distort pretrained features and underperform out-of-distribution. In \textit{International Conference on Learning Representations} (2022). \href{https://arxiv.org/abs/2202.10054}{{\small\texttt{https:/\allowbreak{}/\allowbreak{}arxiv.\allowbreak{}org/\allowbreak{}abs/\allowbreak{}2202.\allowbreak{}10054}}}\par\vspace{2pt}
\noindent\hypertarget{ref:46}{}\hangindent=1.8em\hangafter=1\textbf{46.}\enspace Loshchilov, I. \& Hutter, F. Decoupled weight decay regularization. In \textit{International Conference on Learning Representations} (2019). \href{https://arxiv.org/abs/1711.05101}{{\small\texttt{https:/\allowbreak{}/\allowbreak{}arxiv.\allowbreak{}org/\allowbreak{}abs/\allowbreak{}1711.\allowbreak{}05101}}}\par\vspace{2pt}
\noindent\hypertarget{ref:47}{}\hangindent=1.8em\hangafter=1\textbf{47.}\enspace Svensson, V. Droplet scRNA-seq is not zero-inflated. \textit{Nat. Biotechnol.} \textbf{38}, 147--150 (2020). \href{https://doi.org/10.1038/s41587-019-0379-5}{{\small\texttt{https:/\allowbreak{}/\allowbreak{}doi.\allowbreak{}org/\allowbreak{}10.\allowbreak{}1038/\allowbreak{}s41587-\allowbreak{}019-\allowbreak{}0379-\allowbreak{}5}}}\par\vspace{2pt}
\noindent\hypertarget{ref:48}{}\hangindent=1.8em\hangafter=1\textbf{48.}\enspace Sarkar, A. \& Stephens, M. Separating measurement and expression models clarifies confusion in single-cell RNA sequencing analysis. \textit{Nat. Genet.} \textbf{53}, 770--777 (2021). \href{https://doi.org/10.1038/s41588-021-00873-4}{{\small\texttt{https:/\allowbreak{}/\allowbreak{}doi.\allowbreak{}org/\allowbreak{}10.\allowbreak{}1038/\allowbreak{}s41588-\allowbreak{}021-\allowbreak{}00873-\allowbreak{}4}}}\par\vspace{2pt}
\noindent\hypertarget{ref:49}{}\hangindent=1.8em\hangafter=1\textbf{49.}\enspace Zaheer, M. et al. Deep Sets. \textit{Adv. Neural Inf. Process. Syst.} \textbf{30}, 3391--3401 (2017). \href{https://arxiv.org/abs/1703.06114}{{\small\texttt{https:/\allowbreak{}/\allowbreak{}arxiv.\allowbreak{}org/\allowbreak{}abs/\allowbreak{}1703.\allowbreak{}06114}}}\par\vspace{2pt}
\noindent\hypertarget{ref:50}{}\hangindent=1.8em\hangafter=1\textbf{50.}\enspace Sundararajan, M., Taly, A. \& Yan, Q. Axiomatic attribution for deep networks. \textit{Proc. Mach. Learn. Res.} \textbf{70}, 3319--3328 (2017). \href{https://arxiv.org/abs/1703.01365}{{\small\texttt{https:/\allowbreak{}/\allowbreak{}arxiv.\allowbreak{}org/\allowbreak{}abs/\allowbreak{}1703.\allowbreak{}01365}}}\par\vspace{2pt}
}

\clearpage

\section*{Extended Data Figures}

\clearpage

\clearpage

\clearpage

\clearpage

\clearpage

\clearpage

\clearpage

\clearpage

\clearpage

\clearpage

\clearpage
\thispagestyle{plain}
\begin{flushleft}
{\footnotesize\sffamily\bfseries\color{biorxivslate} SUPPLEMENTARY INFORMATION}\\[6pt]
{\LARGE\sffamily\bfseries\color{biorxivnavy} Do single-cell foundation models need 512 dimensions for one number?}\\[10pt]
{\normalsize\sffamily Lecheng Zhang$^{1,\dagger}$, Yuhao Yang$^{1,2,\dagger}$ and Stan Z. Li$^{1,*}$}\\[6pt]
{\footnotesize\color{biorxivslate} $^{1}$\,AI Research and Innovation Laboratory, School of Engineering, Westlake University, Hangzhou, Zhejiang 310030, China\\ $^{2}$\,SDU-ANU Joint Science College, Shandong University, Weihai, Shandong 264209, China\\[3pt]$^{\dagger}$\,These authors contributed equally to this work. \quad $^{*}$\,Correspondence: \href{mailto:stan.zq.li@westlake.edu.cn}{\texttt{stan.zq.li@westlake.edu.cn}}}
\end{flushleft}
\vspace{4pt}

{\footnotesize\color{biorxivslate}\noindent Reference numbers, figures and Extended Data figures refer to the main manuscript.\par\vspace{4pt}}

\section*{Supplementary Methods}

\subsection*{Supplementary Note 1 | Checkpoints and temporal subsets}

Methods identifies the public checkpoints and architectures used for each model family. The temporal comparisons below use the availability of the underlying data.

A post-cutoff cohort is one released after a model's pretraining cutoff. Comparisons use portal release dates rather than publication dates (Supplementary Table 1). The temporal sensitivity subset excludes adipose because its deposit predates its publication. Supplementary Table 4 distinguishes the primary and temporal donor subsets.

\subsection*{Supplementary Note 2 | Noise-floor assumptions and calibration sensitivity}

The analytical tie variance follows the exact interval occupied by equal counts and needs no count-generating model. The Poisson sensitivity analysis instead resamples one query while holding other counts fixed; it assumes technical sampling and permits induced changes to visible bins. Methods gives the estimators and calibration protocols; Supplementary Table 2 reports numerical comparisons, including visible-only quantile-boundary sensitivity. Full-cell input summaries in controlled pretraining and visible-only rebuilding at evaluation are distinguished in Methods, Thinning and objective controls.

\subsection*{Supplementary Note 3 | Attention masking and attribution sensitivity}

Post-hoc attention-edge deletion removes edges from a trained model. Retraining under the same mask tests whether the remaining connections can learn the task (Extended Data Fig. 7e). Top-ranked and random-edge deletion compare the effects of removing selected edges and removing arbitrary edges.

Integrated Gradients (IG) integrates the path gradient from a baseline to the input. Step sensitivity tests numerical integration; changing the baseline tests path dependence. Channel-uninformative controls specify which discrete inputs contain signal. They do not prescribe an embedding-path attribution. The main audit compares IG for the sum of predictions with positive-part MSE increases from identity permutation and value clamping. These quantities have different endpoints, interventions and baselines. Supplementary Table 5 reports integration settings and completeness errors.

\subsection*{Supplementary Note 4 | Synthetic generator specification}

The synthetic generator simulates 600 genes, 40 transcription factors (TFs), 12 modules and six equiprobable cell types using fixed random seeds. A permutation assigns 480 non-TF genes to modules of 40; the remaining 80 have no module loading. TF $t$ belongs to module $t\bmod12$. Each non-TF gene draws zero, one or two candidate regulators with probabilities 0.35, 0.40 and 0.25, without replacement; a within-module edge is skipped with probability 0.5 and retained edges have coefficient $\pm$0.6 with equal probability. Each cell type upregulates three randomly selected modules to 1 and sets the other module means to $-$0.5.

Cell-specific module activities equal these means plus independent $N(0,0.5^2)$ noise. TF activity is half its module activity plus independent $N(0,1)$ noise. Gene log means sum a fixed $N(-2.5,1.5^2)$ baseline, a module loading of 0.7 where assigned, and regulatory coefficients times TF activity; TF expression also receives 0.8 times its own activity. Cell depth is $\exp[N(\log3,0.5^2)]$. Value-informative counts are negative binomial with this depth-scaled mean and dispersion 3. In the value-null regime, detection is Bernoulli with probability $1-\exp(-\lambda_{cg})$ from the activity-dependent mean, while positive counts are $1+\mathrm{Poisson}(d_c\exp b_g)$. Positive magnitudes thus retain gene baseline and depth but not the planted module/TF activity conditional on those quantities.

Independent fixed random seeds generate 100,000 training cells and 6,000 test cells before filtering, distinct from the model-training runs. Cells with fewer than eight detected genes are removed, and tokenisation uses 50 within-cell quantile bins with uniform random tie-breaking.

\section*{Supplementary Tables}

\section*{Supplementary Table 1 | Human single-cell evaluation cohorts, repository accessions and portal release timestamps}

{\footnotesize\color{darkgray}\noindent Portal public release dates are reported alongside publication years to compare data availability with model pretraining cutoffs (scGPT: June 2023; scFoundation: 2023; Geneformer-V2: August 2024; State: pre-2025). Collection identifiers and publication DOIs identify the public datasets {[}31{]}.\par\vspace{4pt}}

\subsection*{A. Multi-donor droplet single-cell / single-nucleus RNA-sequencing tissue cohorts (10 cohorts; 232 donors; 52,500 evaluated cells)}

\begingroup
\scriptsize
\setlength{\tabcolsep}{3.5pt}
\renewcommand{\arraystretch}{1.18}
\begin{longtable}[]{@{}>{\raggedright\arraybackslash}p{(\textwidth - 16\tabcolsep) * \real{0.0772}}
  >{\raggedright\arraybackslash}p{(\textwidth - 16\tabcolsep) * \real{0.1643}}
  >{\raggedright\arraybackslash}p{(\textwidth - 16\tabcolsep) * \real{0.1705}}
  >{\raggedright\arraybackslash}p{(\textwidth - 16\tabcolsep) * \real{0.1270}}
  >{\raggedright\arraybackslash}p{(\textwidth - 16\tabcolsep) * \real{0.0510}}
  >{\raggedleft\arraybackslash}p{(\textwidth - 16\tabcolsep) * \real{0.0525}}
  >{\raggedright\arraybackslash}p{(\textwidth - 16\tabcolsep) * \real{0.1738}}
  >{\raggedright\arraybackslash}p{(\textwidth - 16\tabcolsep) * \real{0.1738}}@{}}
\toprule
{\sffamily\bfseries Cohort ID} & {\sffamily\bfseries Descriptive tissue / disease cohort} & {\sffamily\bfseries Repository \& dataset UUID} & {\sffamily\bfseries Collection UUID} & {\sffamily\bfseries Initial portal release} & {\sffamily\bfseries Donors (held-out)} & {\sffamily\bfseries Primary role} & {\sffamily\bfseries Verified publication \& DOI} \\
\midrule
\endhead
\bottomrule
\endlastfoot
D01 & Peripheral blood following SARS-CoV-2 infection and mRNA vaccination & CZ CELLxGENE {\small\texttt{242c6e7f-\allowbreak{}9016-\allowbreak{}4048-\allowbreak{}af70-\allowbreak{}d631f5eea188}} & {\small\texttt{ecb739c5-\allowbreak{}fe0d-\allowbreak{}4b48-\allowbreak{}81c6-\allowbreak{}217c4d64eec4}} & 2023-12-04 & 14 & Reference / controlled retraining / 1,024-gene panel & Ivanova, E. N. et al. \textit{iScience} \textbf{26}, 108572 (2023). {\small\texttt{10.\allowbreak{}1016/\allowbreak{}j.\allowbreak{}isci.\allowbreak{}2023.\allowbreak{}108572}} {[}32{]} \\
\addlinespace[2pt]
D02 & Integrated human lung cell atlas (core) & CZ CELLxGENE {\small\texttt{066943a2-\allowbreak{}fdac-\allowbreak{}4b29-\allowbreak{}b348-\allowbreak{}40cede398e4e}} & {\small\texttt{6f6d381a-\allowbreak{}7701-\allowbreak{}4781-\allowbreak{}935c-\allowbreak{}db10d30de293}} & 2022-03-11 & 65 & Reference / lung tissue value-rank evaluation / controlled retraining & Sikkema, L. et al. \textit{Nat. Med.} \textbf{29}, 1563--1577 (2023). {\small\texttt{10.\allowbreak{}1038/\allowbreak{}s41591-\allowbreak{}023-\allowbreak{}02327-\allowbreak{}2}} {[}33{]} \\
\addlinespace[2pt]
D03 & Subcutaneous white adipose tissue & CZ CELLxGENE {\small\texttt{9d8e5dca-\allowbreak{}03a3-\allowbreak{}457d-\allowbreak{}b7fb-\allowbreak{}844c75735c83}} & {\small\texttt{6b701826-\allowbreak{}37bb-\allowbreak{}4356-\allowbreak{}9792-\allowbreak{}ff41fc4c3161}} & 2023-05-15 (rev. 2026-08) & 19 (5 held-out) & Primary 5-cohort set (excluded in 19-donor post-September-2024 sensitivity, Supplementary Table 4) & Efthymiou, V. et al. \textit{Nat. Commun.} \textbf{17}, 9471 (2026). {\small\texttt{10.\allowbreak{}1038/\allowbreak{}s41467-\allowbreak{}026-\allowbreak{}75426-\allowbreak{}6}} {[}34{]} \\
\addlinespace[2pt]
D04 & HPV-negative and HPV-positive oropharyngeal squamous cell carcinoma & CZ CELLxGENE {\small\texttt{b6b5ea88-\allowbreak{}e092-\allowbreak{}46e4-\allowbreak{}b9c5-\allowbreak{}e93b52d5c195}} & {\small\texttt{3c34e6f1-\allowbreak{}6827-\allowbreak{}47dd-\allowbreak{}8e19-\allowbreak{}9edcd461893f}} & 2024-12-05 & 10 (3 held-out) & Post-cutoff evaluation \& controlled retraining & Jenkins, B. H. et al. \textit{Mol. Cancer} \textbf{24}, 3 (2025). {\small\texttt{10.\allowbreak{}1186/\allowbreak{}s12943-\allowbreak{}024-\allowbreak{}02191-\allowbreak{}9}} {[}35{]} \\
\addlinespace[2pt]
D05 & Minor salivary gland in Sjögren disease & CZ CELLxGENE {\small\texttt{df1edb87-\allowbreak{}e512-\allowbreak{}43ae-\allowbreak{}b5f4-\allowbreak{}cb179cfc2bb4}} & {\small\texttt{21bbfaec-\allowbreak{}6958-\allowbreak{}46bc-\allowbreak{}b1cd-\allowbreak{}1535752f6304}} & 2024-09-25 & 25 (6 held-out) & Post-cutoff evaluation \& controlled retraining & Pranzatelli, T. J. F. et al. \textit{Ann. Rheum. Dis.} \textbf{85}, 880--898 (2026). {\small\texttt{10.\allowbreak{}1016/\allowbreak{}j.\allowbreak{}ard.\allowbreak{}2025.\allowbreak{}08.\allowbreak{}014}} {[}36{]} \\
\addlinespace[2pt]
D06 & Paediatric ileal lamina propria & CZ CELLxGENE {\small\texttt{6ceeaa86-\allowbreak{}9ceb-\allowbreak{}4582-\allowbreak{}b390-\allowbreak{}6d4dd6ff0572}} & {\small\texttt{a064d795-\allowbreak{}3b77-\allowbreak{}44ae-\allowbreak{}9fb5-\allowbreak{}beb57c567443}} & 2023-08-22 & 13 & Controlled retraining (10-cohort corpus) & Zheng, H. B. et al. \textit{eLife} \textbf{12}, RP91792 (2026). {\small\texttt{10.\allowbreak{}7554/\allowbreak{}eLife.\allowbreak{}91792}} {[}37{]} \\
\addlinespace[2pt]
D07 & Primary human breast cancer atlas & CZ CELLxGENE {\small\texttt{9fddb063-\allowbreak{}056d-\allowbreak{}4202-\allowbreak{}8b8a-\allowbreak{}4b0ee531d3ce}} & {\small\texttt{dea97145-\allowbreak{}f712-\allowbreak{}431c-\allowbreak{}a223-\allowbreak{}6b5f565f362a}} & 2023-11-30 & 26 & Controlled retraining (10-cohort corpus) & Wu, S. Z. et al. \textit{Nat. Genet.} \textbf{53}, 1334--1347 (2021). {\small\texttt{10.\allowbreak{}1038/\allowbreak{}s41588-\allowbreak{}021-\allowbreak{}00911-\allowbreak{}1}} {[}38{]} \\
\addlinespace[2pt]
D08 & Autoimmune and immunodeficiency peripheral blood & CZ CELLxGENE {\small\texttt{c2a461b1-\allowbreak{}0c15-\allowbreak{}4047-\allowbreak{}9fcb-\allowbreak{}1f966fe55100}} & {\small\texttt{eb735cc9-\allowbreak{}d0a7-\allowbreak{}48fa-\allowbreak{}b255-\allowbreak{}db726bf365af}} & 2022-03-31 & 22 & Controlled retraining (10-cohort corpus) & CZI Single-Cell COVID-19 Consortia. \textit{medRxiv} (2020). {\small\texttt{10.\allowbreak{}1101/\allowbreak{}2020.\allowbreak{}11.\allowbreak{}20.\allowbreak{}20227355}} {[}39{]} \\
\addlinespace[2pt]
D09 & Developing human fetal retina & CZ CELLxGENE {\small\texttt{aecc2743-\allowbreak{}4ff2-\allowbreak{}4b55-\allowbreak{}ac8c-\allowbreak{}a1ff76fc4bb4}} & {\small\texttt{c11009b8-\allowbreak{}b113-\allowbreak{}4a99-\allowbreak{}9890-\allowbreak{}78b2f9df9d79}} & 2025-10-01 & 22 (6 held-out) & Post-cutoff evaluation \& controlled retraining & Kriukov, E. et al. \textit{Dev. Biol.} \textbf{528}, 91--104 (2025). {\small\texttt{10.\allowbreak{}1016/\allowbreak{}j.\allowbreak{}ydbio.\allowbreak{}2025.\allowbreak{}08.\allowbreak{}025}} {[}40{]} \\
\addlinespace[2pt]
D10 & Leptomeningeal cerebrospinal fluid immune cells & CZ CELLxGENE {\small\texttt{0bf30b02-\allowbreak{}db97-\allowbreak{}4560-\allowbreak{}b9cd-\allowbreak{}0fe3588892af}} & {\small\texttt{573e2e06-\allowbreak{}8af0-\allowbreak{}4d96-\allowbreak{}bfdd-\allowbreak{}7d64a4bb9c21}} & 2025-07-08 & 16 (4 held-out) & Post-cutoff evaluation \& controlled retraining & Nieto, P. et al. \textit{Cell Rep. Med.} \textbf{7}, 102651 (2026). {\small\texttt{10.\allowbreak{}1016/\allowbreak{}j.\allowbreak{}xcrm.\allowbreak{}2026.\allowbreak{}102651}} {[}41{]} \\
\addlinespace[2pt]
\end{longtable}
\endgroup

\subsection*{B. Multi-donor cellular indexing of transcriptomes and epitopes by sequencing (CITE-seq) surface-proteomics cohorts (4 cohorts; 32 held-out donors; 24,000 cells; 463 non-isotype antibody channels, 461 stained)}

\begingroup
\scriptsize
\setlength{\tabcolsep}{3.5pt}
\renewcommand{\arraystretch}{1.18}
\begin{longtable}[]{@{}>{\raggedright\arraybackslash}p{(\textwidth - 14\tabcolsep) * \real{0.0772}}
  >{\raggedright\arraybackslash}p{(\textwidth - 14\tabcolsep) * \real{0.2457}}
  >{\raggedright\arraybackslash}p{(\textwidth - 14\tabcolsep) * \real{0.0866}}
  >{\raggedright\arraybackslash}p{(\textwidth - 14\tabcolsep) * \real{0.0757}}
  >{\raggedleft\arraybackslash}p{(\textwidth - 14\tabcolsep) * \real{0.0444}}
  >{\raggedleft\arraybackslash}p{(\textwidth - 14\tabcolsep) * \real{0.1256}}
  >{\raggedright\arraybackslash}p{(\textwidth - 14\tabcolsep) * \real{0.3348}}@{}}
\toprule
{\sffamily\bfseries Cohort ID} & {\sffamily\bfseries Descriptive tissue / disease cohort} & {\sffamily\bfseries NCBI GEO accession} & {\sffamily\bfseries GEO public date} & {\sffamily\bfseries Donors} & {\sffamily\bfseries Non-isotype antibody channels (stained)} & {\sffamily\bfseries Verified source / publication} \\
\midrule
\endhead
\bottomrule
\endlastfoot
CITE01 & Human bone marrow mononuclear cells & {\small\texttt{GSE194122}} & 2022-01-24 & 9 & 134 (134) & Luecken, M. D. et al. \textit{NeurIPS Datasets and Benchmarks} (2021); NCBI GEO {\small\texttt{GSE194122}} \\
\addlinespace[2pt]
CITE02 & Trisomy-8 autoinflammatory peripheral blood & {\small\texttt{GSE272583}} & 2026-09-17 & 9 & 130 (130) & Manthiram, K. et al. Deep immunologic profiling of trisomy-8 autoinflammatory disease; NCBI GEO {\small\texttt{GSE272583}} (2026) \\
\addlinespace[2pt]
CITE03 & Lecanemab ARIA peripheral blood & {\small\texttt{GSE316096}} & 2026-01-20 & 6 & 130 (130) & Johnson, L. A. et al. \textit{Nat. Commun.} \textbf{17}, 2180 (2026). {\small\texttt{10.\allowbreak{}1038/\allowbreak{}s41467-\allowbreak{}026-\allowbreak{}68921-\allowbreak{}3}} {[}43{]} \\
\addlinespace[2pt]
CITE04 & Inflammatory bowel disease intestinal mucosa & {\small\texttt{GSE331003}} & 2026-09-23 & 8 & 69 (67) & Hesser, L. A. et al. Intestinal T-bet+ plasmablasts in epithelial barrier dysfunction and IBD; NCBI GEO {\small\texttt{GSE331003}} (2026) \\
\addlinespace[2pt]
\end{longtable}
\endgroup

{\footnotesize\color{darkgray}\noindent Channel counts reflect the 463 non-isotype channels (461 with nonzero staining) evaluated across donor-grouped cross-validation.\par\vspace{4pt}}

\subsection*{C. Pooled perturbation-response cell-line cohorts (4 cohorts; 320 perturbations; 966,728 total source cells / 27,887 evaluated split-half cells; 310 in-vocabulary target genes masked)}

CRISPR-interference (CRISPRi) Perturb-seq cohorts support evaluation of split-half perturbation-associated embedding-shift consistency.

\begingroup
\footnotesize
\setlength{\tabcolsep}{4.5pt}
\renewcommand{\arraystretch}{1.18}
\begin{longtable}[]{@{}>{\raggedright\arraybackslash}p{(\textwidth - 10\tabcolsep) * \real{0.0472}}
  >{\raggedright\arraybackslash}p{(\textwidth - 10\tabcolsep) * \real{0.2502}}
  >{\raggedleft\arraybackslash}p{(\textwidth - 10\tabcolsep) * \real{0.1755}}
  >{\raggedleft\arraybackslash}p{(\textwidth - 10\tabcolsep) * \real{0.1688}}
  >{\raggedright\arraybackslash}p{(\textwidth - 10\tabcolsep) * \real{0.3483}}@{}}
\toprule
{\sffamily\bfseries Cohort ID} & {\sffamily\bfseries Cell line} & {\sffamily\bfseries Total source cells (evaluated split-half cells)} & {\sffamily\bfseries Perturbations evaluated (in-vocabulary target genes masked)} & {\sffamily\bfseries Verified publication \& DOI} \\
\midrule
\endhead
\bottomrule
\endlastfoot
D11 & Jurkat T lymphoblast & 262,956 (6,978) & 80 (78) & Nadig, A. et al. \textit{Nat. Genet.} \textbf{57}, 1228--1237 (2025). {\small\texttt{10.\allowbreak{}1038/\allowbreak{}s41588-\allowbreak{}025-\allowbreak{}02169-\allowbreak{}3}} {[}30{]} \\
\addlinespace[2pt]
D12 & HepG2 hepatocellular carcinoma & 145,473 (6,731) & 80 (78) & Nadig, A. et al. \textit{Nat. Genet.} \textbf{57}, 1228--1237 (2025). {\small\texttt{10.\allowbreak{}1038/\allowbreak{}s41588-\allowbreak{}025-\allowbreak{}02169-\allowbreak{}3}} {[}30{]} \\
\addlinespace[2pt]
D13 & K562 chronic myelogenous leukemia (essential-scale) & 310,385 (7,535) & 80 (77) & Replogle, J. M. et al. \textit{Cell} \textbf{185}, 2559--2575.e28 (2022). {\small\texttt{10.\allowbreak{}1016/\allowbreak{}j.\allowbreak{}cell.\allowbreak{}2022.\allowbreak{}05.\allowbreak{}013}} {[}29{]} \\
\addlinespace[2pt]
D14 & RPE1 retinal pigment epithelium (essential-scale) & 247,914 (6,643) & 80 (77) & Replogle, J. M. et al. \textit{Cell} \textbf{185}, 2559--2575.e28 (2022). {\small\texttt{10.\allowbreak{}1016/\allowbreak{}j.\allowbreak{}cell.\allowbreak{}2022.\allowbreak{}05.\allowbreak{}013}} {[}29{]} \\
\addlinespace[2pt]
\end{longtable}
\endgroup

\subsection*{D. Chromatin-accessibility cohorts and evaluation groups}

\begingroup
\footnotesize
\setlength{\tabcolsep}{4.5pt}
\renewcommand{\arraystretch}{1.18}
\begin{longtable}[]{@{}>{\raggedright\arraybackslash}p{(\textwidth - 8\tabcolsep) * \real{0.1524}}
  >{\raggedright\arraybackslash}p{(\textwidth - 8\tabcolsep) * \real{0.1044}}
  >{\raggedright\arraybackslash}p{(\textwidth - 8\tabcolsep) * \real{0.2563}}
  >{\raggedright\arraybackslash}p{(\textwidth - 8\tabcolsep) * \real{0.4768}}@{}}
\toprule
{\sffamily\bfseries Cohort} & {\sffamily\bfseries Accession} & {\sffamily\bfseries Curated cells / grouping} & {\sffamily\bfseries Evaluation definition} \\
\midrule
\endhead
\bottomrule
\endlastfoot
BMMC multiome & GSE194122 & 20,000 cells, 10 donors & Donor-grouped three-fold evaluation; paired RNA control \\
\addlinespace[2pt]
Brain scATAC & GSE312668 & 18,000 cells, nine donors & Donor-grouped three-fold evaluation \\
\addlinespace[2pt]
Cerebellum multiome & GSE331288 & 21,660 cells, 10 donors & Donor-grouped three-fold evaluation \\
\addlinespace[2pt]
DOGMA trimodal & GSE156478 & 16,000 cells, four libraries & Library-grouped three-fold evaluation; paired RNA control; library labels do not establish donor independence \\
\addlinespace[2pt]
\end{longtable}
\endgroup

Figure 2k uses at most 2,100 stratified cells per cohort and fixed intact-trained probes. The separate controlled BMMC models use a donor-disjoint train/test split (three held-out test donors; 15,241 training and 4,759 test cells).

\section*{Supplementary Table 2 | Backbone width scan, scFoundation zero-vs-positive decomposition and controlled gene-panel mechanisms}

\subsection*{A. Backbone width scan \& width-512 learning-rate check (ten-cohort 3,000-cell target-calibrated masked quantile error)}

\begingroup
\footnotesize
\setlength{\tabcolsep}{4.5pt}
\renewcommand{\arraystretch}{1.18}
\begin{longtable}[]{@{}>{\raggedright\arraybackslash}p{(\textwidth - 10\tabcolsep) * \real{0.2462}}
  >{\raggedright\arraybackslash}p{(\textwidth - 10\tabcolsep) * \real{0.2322}}
  >{\raggedleft\arraybackslash}p{(\textwidth - 10\tabcolsep) * \real{0.1150}}
  >{\raggedleft\arraybackslash}p{(\textwidth - 10\tabcolsep) * \real{0.1261}}
  >{\raggedright\arraybackslash}p{(\textwidth - 10\tabcolsep) * \real{0.2705}}@{}}
\toprule
{\sffamily\bfseries Configuration \& width} & {\sffamily\bfseries Epochs \& seeds} & {\sffamily\bfseries Masked MSE ($\text{bin}^2$)} & {\sffamily\bfseries Recovery vs learned 256-D (\%)} & {\sffamily\bfseries Paired contrast / note} \\
\midrule
\endhead
\bottomrule
\endlastfoot
Learned gene embeddings, width 64 & 5 epochs, $n = 1$ seed & 150.06 & 75.33\% & Single-seed run \\
\addlinespace[2pt]
Random 32-D codes + bias, width 64 & 5 epochs ($n = 3$) / 10 epochs ($n = 1$) & 150.70 / 149.50 & 64.52\% / 66.13\% & Width capacity bottleneck \\
\addlinespace[2pt]
Random 32-D codes + bias, width 128 & 5 epochs ($n = 3$) / 10 epochs ($n = 1$) & 150.09 / 148.24 & 74.68\% / 82.86\% & Monotonic width recovery \\
\addlinespace[2pt]
Learned gene embeddings, width 256 (reference) & 5 epochs / 10 epochs ($n = 3$) & 148.60 / 146.94 & 100.00\% / 100.00\% & Baseline reference per epoch \\
\addlinespace[2pt]
Random 32-D codes + bias, width 256 & 5 epochs / 10 epochs ($n = 3$) & 148.89 / 147.26 & 95.04\% / 95.75\% & Equivalent to learned at width 256 \\
\addlinespace[2pt]
Random 32-D codes + bias, width 512 & 5 epochs ($n = 3$) / 10 epochs ($n = 1$) & 148.99 / 147.62 & 93.38\% / 90.97\% & Plateaus below learned 512-D \\
\addlinespace[2pt]
Learned gene embeddings, width 512 & 5 epochs, $n = 1$ seed & 147.97 & 110.55\% & Matched run (148.59 vs. 147.97) random minus learned +0.62 {[}0.50, 0.75{]} \\
\addlinespace[2pt]
Random 32-D codes, width 512 LR check & 5 epochs, $n = 1$ ($0.5\times, 1\times, 2\times$) & 148.74 / 148.59 / 153.22 & Not applicable & Optimisation LR fails to close width-512 gap \\
\addlinespace[2pt]
\end{longtable}
\endgroup

\subsection*{B. Controlled gene-panel three-seed mechanisms \& scFoundation zero-vs-positive MSE}

\begingroup
\footnotesize
\setlength{\tabcolsep}{4.5pt}
\renewcommand{\arraystretch}{1.18}
\begin{longtable}[]{@{}>{\raggedright\arraybackslash}p{(\textwidth - 8\tabcolsep) * \real{0.3062}}
  >{\raggedright\arraybackslash}p{(\textwidth - 8\tabcolsep) * \real{0.1640}}
  >{\raggedleft\arraybackslash}p{(\textwidth - 8\tabcolsep) * \real{0.2135}}
  >{\raggedright\arraybackslash}p{(\textwidth - 8\tabcolsep) * \real{0.3062}}@{}}
\toprule
{\sffamily\bfseries Comparison group \& condition} & {\sffamily\bfseries Seeds / scope} & {\sffamily\bfseries Value or gain share} & {\sffamily\bfseries Paired contrast {[}95\% CI{]}} \\
\midrule
\endhead
\bottomrule
\endlastfoot
48-pass trajectories (passes 12, 24, 48): Attention & $n = 3$ independent runs & 96.26\% / 99.75\% / 104.54\% & Attn minus Set lead: +16.8 {[}13.8, 19.8{]} pp (p12) \\
\addlinespace[2pt]
48-pass trajectories (passes 12, 24, 48): Deep Sets & $n = 3$ independent runs & 79.44\% / 84.66\% / 87.34\% & +15.1 {[}11.8, 18.4{]} pp (p24); +17.2 {[}15.0, 19.4{]} pp (p48) \\
\addlinespace[2pt]
Static membership: Deep Sets visible vs vis+masked & $n = 3$ seeds, 12 passes & 70.94\% $\rightarrow$ 79.33\% & Membership gain +8.39 {[}6.74, 10.12{]} pp \\
\addlinespace[2pt]
Static membership: Attention visible-only vs static-KV & $n = 3$ seeds, 12 passes & 86.94\% $\rightarrow$ 94.11\% & Membership gain +7.17 {[}5.44, 8.97{]} pp \\
\addlinespace[2pt]
Static membership difference (Set minus Attention) & $n = 3$ seeds, 4,000-draw boot & Not applicable & 1.21 {[}-0.96, 3.35{]} pp (comparable; full-vocab readable mask -0.163 pp) \\
\addlinespace[2pt]
Visible-only encoder depth ($L=1, 3, 5$; 1-layer baseline) & $n = 3$ seeds, 12 passes & 81.16\% / 82.66\% / 80.97\% (1L 81.65\%) & Deep minus shallow ($L=5$ minus $L=1$) -0.20 {[}-1.48, 1.06{]} pp; ($L=3$ minus $L=1$) +1.50 {[}0.19, 2.80{]} pp \\
\addlinespace[2pt]
Panel non-neural predictors (visible+masked identities) & 1,999 test cells, 5,230 queries & 10.15\% + 19.59\% + 34.89\% = 64.63\% & Unmodified Transformer dynamic share 5.89 pp above static KV \\
\addlinespace[2pt]
Panel random 32-D codes (unbiased $n=2$ / biased $n=3$) & $n = 2$ vs $n = 3$ seeds, 12 passes & 98.49\% ($n=2$) / 97.89\% ($n=3$) & Reference Transformer 12-pass MSE 61.68; gene mean 146.53 \\
\addlinespace[2pt]
Post-hoc attention masking vs retrained sparsity & $n = 3$ seeds & -111.58 pp (post-hoc) vs -15.42 pp (retrained) & Demonstrates off-manifold vulnerability of post-hoc edge zeroing \\
\addlinespace[2pt]
scFoundation 24-donor MSE (no target fitting / +3,000 cells) & Mixed zero+pos vs positive-only & Uncond 0.637 $\rightarrow$ 0.388 (raw 0.392); cal3k 0.369 & Pos-only 0.539 (gene mean), 0.543 (raw), 0.221 (out-cohort), 0.146 (3k cal; mean 0.371) \\
\addlinespace[2pt]
scGPT analytical noise decomposition \& boundary sensitivity & 5 cohorts, full-cell vs vis-only & Out-cohort $G$ 171.34, tie floor 72.8 (42.49\%) & $W^\star/W_{\text{tie}}$ 51.74\% (calibrated scGPT capture 30.11\%); vis-only shift +0.075 (+0.048\%) / -0.054 (-0.036\%) \\
\addlinespace[2pt]
\end{longtable}
\endgroup

{\footnotesize\color{darkgray}\noindent Notes: Panel intervals use 4,000-draw two-level run-by-cell bootstraps. Full-vocabulary lookups in Fig. 3b compare prediction gain with output agreement.\par\vspace{4pt}}

\section*{Supplementary Table 3 | Interpretability readouts and matched controls}

Enhancer--promoter rows report overall and distance-matched AUROC in separate columns.

\begingroup
\footnotesize
\setlength{\tabcolsep}{4.5pt}
\renewcommand{\arraystretch}{1.18}
\begin{longtable}[]{@{}>{\raggedright\arraybackslash}p{(\textwidth - 10\tabcolsep) * \real{0.1519}}
  >{\raggedright\arraybackslash}p{(\textwidth - 10\tabcolsep) * \real{0.2257}}
  >{\raggedright\arraybackslash}p{(\textwidth - 10\tabcolsep) * \real{0.1608}}
  >{\raggedright\arraybackslash}p{(\textwidth - 10\tabcolsep) * \real{0.2257}}
  >{\raggedright\arraybackslash}p{(\textwidth - 10\tabcolsep) * \real{0.2257}}@{}}
\toprule
{\sffamily\bfseries Interpretability readout} & {\sffamily\bfseries Model / condition} & {\sffamily\bfseries Reported or raw metric} & {\sffamily\bfseries Matched null or construction-defined control} & {\sffamily\bfseries Primary confound or verdict} \\
\midrule
\endhead
\bottomrule
\endlastfoot
Additive-embedding IG (e\_id vs e\_val) & Fixed-bin log CP10k identity-uninformative control (ID info = 0) & IG ID 0.759--0.892 & Identity is randomised before embedding; defined intervention-loss identity share is 0 & IG varies embeddings after randomisation; its share is not an information fraction \\
\addlinespace[2pt]
Additive-embedding IG (e\_id vs e\_val) & Rank-target value-uninformative control (Val info = 0) & Zero-base IG Val 14.5\% & Values are clamped before embedding; within-cell identity permutation and clamping both give zero loss change, so the loss ratio is undefined & Zero-baseline IG can credit displacement of a constant embedding; retain the attribution and omit the undefined ratio \\
\addlinespace[2pt]
Additive-embedding IG ({\small\texttt{e\_\allowbreak{}id}} vs {\small\texttt{e\_\allowbreak{}val}}) & Pretrained scGPT (33 million pretraining cells) (expression / cell-context decoders) & IG ID 0.68--0.82 & No channel ground truth: within-cell identity permutation equals within-cell value shuffling in expectation; constant clamping is off-manifold & No conclusion on scGPT's channel use \\
\addlinespace[2pt]
Value-MLP geometry \& LayerNorm & Pretrained scGPT (33 million pretraining cells) (before / after normalisation) & PC1+PC2 = 99.78\% / 99.97\% & In-domain knots 160/512 (31.2\%) & Sparse ReLU knots + spherical LN arc; shifts IG norm ratio \\
\addlinespace[2pt]
Residual-stream TopK SAE & Controlled layers L3 / L6 (trained, 2 seeds) & Enriched 9.0\% / 12.6\%; reproducibility 0.10\% & Random directions + ReLU 39.1\% / 39.8\% (0.0\% SVD) & Random ReLU beats SAE 3--4$\times$; SVD invisibility is tautological \\
\addlinespace[2pt]
Residual-stream TopK SAE & Pretrained scGPT (33 million pretraining cells) L6 / L12 (2 seeds) & Enriched 40.4\% / 35.1\%; reproducibility 4.59\%/2.83\% & Random directions + ReLU 43.3\% / 50.5\%; PCA+ 44.1\% & Random ReLU and PCA+ exceed trained SAE; >95\% atoms non-reproducible \\
\addlinespace[2pt]
Static gene table (GO\_BP / Reactome) & Pretrained scGPT (33 million pretraining cells) ($d=512$) & GO\_BP 0.595, React 0.643, frequency-matched negatives 0.563 & PPMI-SVD 0.602/0.653 (frequency-matched negatives 0.567); five-scalar baseline 0.572; deg sum 0.914 & Matched by binary co-occurrence \& 5-scalar confounds; PR drops 52.5\% under scaling \\
\addlinespace[2pt]
Static gene table (similarity to initialisation) & Controlled nonlinear rank-target model ($E$ vs $E-\alpha E_0$) & Full $E$: PR 252.6, GO\_BP 0.504 & Residual $R$: PR 9.0, GO\_BP 0.547 ($\text{CKA}=0.996$) & Raw table dominated by random initialisation $E_0$ \\
\addlinespace[2pt]
Attention-derived GRN (TRRUST) & Controlled nonlinear, affine and random-code models & Raw 0.659 / 0.649 / 0.664 & Marginal $m_i+m_j$ 0.637--0.647 (peak head 0.804); med head $R_{ij}$ 0.513--0.526; cov 0.638 & Dominated by node marginal hub sinks \& degree bias (0.996) \\
\addlinespace[2pt]
Cell embeddings (10 cohorts) & Controlled models vs detection PCA256 and untrained models & Controlled-model F1 0.637--0.662 ($R^2_{\text{depth}} 53.7$--87.3\%) & PCA256 F1 0.733; Untrained 0.438 vs RandProj 0.453; scGPT 0.806 (five cohorts, different protocol) & Depth dominates controlled-model variance; binary PCA beats 200k Transformers \\
\addlinespace[2pt]
Single-gene token deletion (ISP) & Controlled models / scGPT / synthetic benchmark & Trained nuisance $R^2$ 26.5\%--49.2\% ($\rho=0.37$--0.68) & Untrained $R^2$ 93.2\%--95.9\% (mean pooling), 47.3\% ({[}CLS{]}); synthetic TF AUROC 0.684 (unadjusted) vs freq 0.520 & Genuine trans signal in synthetic test, but requires covariate adjustment \\
\addlinespace[2pt]
Surgical masking \& subspace patching & Controlled panel \& 5-layer Transformer & Post-hoc edge zeroing $-$111.58 pp; untrained DAS 172\% & Retrained mask $-$15.42 pp; probe-subspace swap 0.23\% & Off-manifold activation shift \& scalar-readout patching illusion \\
\addlinespace[2pt]
Enhancer--promoter scoring & InverseDistance & Overall AUROC 0.877 & Distance-matched AUROC 0.524 & CRISPRi benchmark; matching defined in Methods \\
\addlinespace[2pt]
Enhancer--promoter scoring & ABC proxy & Overall AUROC 0.918 & Distance-matched AUROC 0.735 & CRISPRi benchmark; matching defined in Methods \\
\addlinespace[2pt]
Enhancer--promoter scoring & ChromatinActivity & Overall AUROC 0.782 & Distance-matched AUROC 0.798 & CRISPRi benchmark; matching defined in Methods \\
\addlinespace[2pt]
Enhancer--promoter scoring & Multiome raw correlation & Overall AUROC 0.687 & Distance-matched AUROC 0.643 & CRISPRi benchmark; matching defined in Methods \\
\addlinespace[2pt]
Enhancer--promoter scoring & Multiome residualised correlation & Overall AUROC 0.579 & Distance-matched AUROC 0.560 & CRISPRi benchmark; matching defined in Methods \\
\addlinespace[2pt]
Enhancer--promoter scoring & EpiAgent static cCRE cosine & Overall AUROC 0.479 & Distance-matched AUROC 0.471 & CRISPRi benchmark; matching defined in Methods \\
\addlinespace[2pt]
Enhancer--promoter scoring & ChromFound token cosine & Overall AUROC 0.868 & Distance-matched AUROC 0.533 & CRISPRi benchmark; matching defined in Methods \\
\addlinespace[2pt]
\end{longtable}
\endgroup

\section*{Supplementary Table 4 | Exact values for statements in the results}

Values rounded or omitted in the Results text. Cohort subsets and denominators distinguish related comparisons. K$n$ denotes an $n$-dimensional value bottleneck in the sweep rows and a white box with $n$ cell-level scalars in the white-box rows; K0 is gene offsets alone. MSE denotes mean squared error; CP10k, counts per ten thousand; UMI, unique molecular identifier; AUROC, area under the receiver operating characteristic curve; PCA, principal-component analysis. Cell-type annotation F1 scores below use macro-averaging. The classification token is denoted {[}CLS{]}.

SAE denotes sparse autoencoder; DAS, distributed alignment search; CKA, centred kernel alignment; ReLU, rectified linear unit; SVD, singular-value decomposition; PPMI, positive pointwise mutual information; TF, transcription factor; GO-BP, Gene Ontology biological process. TRRUST is a curated database of human transcriptional regulatory interactions. PR denotes participation ratio, a spectral effective-dimension measure. ISP denotes in-silico perturbation, implemented here as deleting one gene token and measuring the cell-embedding shift. D denotes detected-gene count; cCRE, candidate cis-regulatory element.

\begingroup
\footnotesize
\setlength{\tabcolsep}{4.5pt}
\renewcommand{\arraystretch}{1.18}
\begin{longtable}[]{@{}>{\raggedright\arraybackslash}p{(\textwidth - 6\tabcolsep) * \real{0.0872}}
  >{\raggedright\arraybackslash}p{(\textwidth - 6\tabcolsep) * \real{0.4071}}
  >{\raggedright\arraybackslash}p{(\textwidth - 6\tabcolsep) * \real{0.4957}}@{}}
\toprule
{\sffamily\bfseries Section} & {\sffamily\bfseries Statement in the text} & {\sffamily\bfseries Exact values} \\
\midrule
\endhead
\bottomrule
\endlastfoot
2.1 & Value curve nearly planar & PC1+PC2 variance: scGPT 99.97\%, scFoundation 96.56\% (50 input bins) \\
\addlinespace[2pt]
2.1 & Straight-line replacement of pretrained value encoders & scGPT and scFoundation downstream transfer changes little; scFoundation's straight line keeps 44.76\% of masked gain on K562 \\
\addlinespace[2pt]
2.1 & Affine vs MLP value encoder, masked prediction & affine error higher by 0.157 squared quantile bins (donor-paired, 59 donors; about 0.1\% of 149.35) \\
\addlinespace[2pt]
2.1 & Affine vs nonlinear, annotation & affine +0.77 macro-F1 points (95\% interval $-$1.4 to 2.9; 40 donors; donor-equal, three-seed mean) \\
\addlinespace[2pt]
2.1 & Affine vs nonlinear, surface proteins & +0.0018 Pearson r (32 donors) \\
\addlinespace[2pt]
2.1 & Random gene codes and width & random codes at width 256: 95.0\% (three seeds; 95.8\% at ten epochs); width 64: 64.5\% (66.1\% at ten epochs); learned 64-d table 75.3\%; width 512: learned 110.6\%, random 93.4\% \\
\addlinespace[2pt]
2.1 & Whole-network narrowing & relative net gain 92.5\% (d=128), 74.6\% (d=32), 30.1\% (d=4); annotation 77 to 18 macro-F1; donor-paired error +1.9\% at d=32 \\
\addlinespace[2pt]
2.1 & Two-track, four-dimensional value track & 95\% net gain (94.5--96.3\% per seed); 64-d track 104\%; donor-paired error +0.47\% (90\% interval +0.32 to +0.62\%); annotation $-$0.45 points (79.7 vs 80.2 macro-F1; 90\% interval $-$1.51 to +0.61; 24 donors; equivalence fails narrowly) and +0.11 points (90\% interval $-$0.78 to +1.00; 40 donors; equivalence passes) \\
\addlinespace[2pt]
2.1 & Bottleneck sweep (rank-target objective) & relative net gain: nonlinear K1 103.7\% (error and downstream gates pass; gain equivalence fails above the upper margin), K2 95.6\%, K4 100.5\%, K8 101.1\%, K16 102.1\%, K32 104.1\% (K32 $-$3.3 annotation points in D01), affine 89.5\% (annotation $-$1.3 to $-$2.1 points) \\
\addlinespace[2pt]
2.1 & Bottleneck sweep, scFoundation-sized and log CP10k blocks & scFoundation K8 96.2\%; log CP10k: affine 99.1\%, K2 101.1\%, K8 100.9\%; log CP10k affine: $-$1.7 six-cohort annotation points, $-$0.013 surface-protein r \\
\addlinespace[2pt]
2.2 & Annotation retention under clamping & scGPT 101.0\% (two labelled cells per class, two-shot) / 97.7\% (all labels); scFoundation 100.3\% / 98.4\%; Geneformer-V2 107.6\% / 100.0\%; 24 donors, 19 from cohorts released after September 2024 \\
\addlinespace[2pt]
2.2 & Mean change under clamping across label budgets & scGPT +0.5, $-$0.5, +0.1, $-$1.1, $-$1.8, $-$2.1, $-$2.3, $-$1.7; scFoundation +0.1, $-$0.6, +0.7, $-$0.3, $-$1.4, $-$0.9, $-$1.0, $-$1.1 macro-F1 points (2, 5, 10, 25, 50, 100, 200, all labels); single donors $-$16.8 to +8.4 (scGPT) \\
\addlinespace[2pt]
2.2 & Shuffling values (scGPT) & macro-F1 59.52 to 54.87 (two-shot), 83.30 to 78.56 (all labels) \\
\addlinespace[2pt]
2.2 & State positive control & two-shot retention 89.4\% \\
\addlinespace[2pt]
2.2 & Surface-protein retention & scGPT 98.6\%, scFoundation 97.3\%, Geneformer-V2 96.7\%, State 94.3\%; per-donor r change: clamping $-$0.005 to $-$0.006, shuffling $-$0.008 to $-$0.010, permuted identities $-$0.22 to $-$0.38 \\
\addlinespace[2pt]
2.2 & Perturbation consistency and complexes & split-half retention scGPT 98.0\%, scFoundation 92.5\%; same-complex AUROC retention 88.3\%, 83.0\% \\
\addlinespace[2pt]
2.2 & Raw counts & 64.4\% of nonzero entries single molecules (63.3\% pooled), 80.6\% at most two; detection 83.2\% of per-gene variance; 21,000 cells, four cohorts \\
\addlinespace[2pt]
2.2 & Tie-breaking floor & 42.49\% of gene-mean error (scGPT quantile bins, five cohorts) \\
\addlinespace[2pt]
2.2 & scFoundation zeros vs positives & MSE unconditional 0.392 vs 0.637 (gene means); positive-only 0.543 vs 0.539; calibrated positive-only 0.221, unconditional 0.388 \\
\addlinespace[2pt]
2.3 & Calibration and score--function diagonal & gain/deviation-energy ratio 1.01--1.10 (controlled Transformers); panel tree-closure + neighbour predictor 63.8\% of gain, 64.6\% of function; full-vocabulary kNN retrieval 81\% of gain, 22\% of function (relative to the training gene mean) \\
\addlinespace[2pt]
2.3 & White box, controlled Transformer & gene offsets K0 0.294; K3 0.805 (95\% interval across cohorts 0.77--0.84) = 95\% of seed agreement 0.843; K2 0.743, K8 0.875; shuffled test inputs $-$0.233; other controlled Transformers 81--93\% of their reproducibility \\
\addlinespace[2pt]
2.3 & White box, pretrained scGPT & K0 0.612, K3 0.794 to 47\% beyond offsets; controlled Transformer on the same five cohorts 76\% \\
\addlinespace[2pt]
2.3 & Leading cell-level scalar & log detected-gene count alone 68\% (62--73\% across cohorts); + cell type 85\% (80--89\%); + quality covariates 96\%; scGPT cell type + D 82\%; 136,135 of 200,000 training cells exceed 1,024 detected genes; visible window predicts log detected-gene count, $R^2$ 0.95 \\
\addlinespace[2pt]
2.3 & Rank arithmetic & multinomial model vs Transformer gene responses r 0.62 (0.48 after controlling detection rate and abundance); window-ranked targets: D share 68\% vs 68\% (difference 0.6 points, 95\% interval $-$0.4 to 1.6) \\
\addlinespace[2pt]
2.3 & Pathway programmes & 0 of 1,801 gene sets enriched after abundance control \\
\addlinespace[2pt]
2.3 & Scale and training length & 1M cells, one pass: K3 0.810, D share 67\%; three passes: +70\% gain over cohort means, value encoders agree r 0.95, K3 0.70, K8 0.79, free state 0.71 / 0.82; annealed one-pass run 0.79--0.82; controlled 20-epoch schedule K3 0.81, 0.73, 0.69 after 5, 10, 20 epochs; D share 64--67\% \\
\addlinespace[2pt]
2.3 & Sparse context & 64 visible genes: free cell state with three dimensions 0.60 (0.71 at full context), seed reproducibility 0.86, linear encoder 0.67 to 0.33; trained on 64 genes: free state 0.82, seed reproducibility 0.95, linear encoder 0.40; panel (17.96 visible genes): linear class 0.41 (K3), saturating near 0.66; attention vs deep sets +16.8 points (12 passes), +17.2 (48 passes) \\
\addlinespace[2pt]
2.4 & Thinning, rank-target objective & response size 0.74 (one half) and 1.34 (one quarter) $\times$ r.m.s. function; white-box $R^2$ 0.78 / 0.81 (slopes 0.88 / 0.99) over 2.0 million queries; per cohort 0.66--0.87 / 0.74--0.88; 84\% of the change in the span of the gene readouts \\
\addlinespace[2pt]
2.4 & Movement along the depth axis & depth axis reads log10 D with $R^2$ 0.91 (lung); 72\% of squared displacement in the median cohort (58--84\% across ten cohorts); state shift $-$0.20 vs estimated log10 D change $-$0.21 (one half), $-$0.32 vs $-$0.43 (one quarter) \\
\addlinespace[2pt]
2.4 & Objective controls & fixed-bin log CP10k: one cell-level scalar 99\% of white-box gain (rank-target objective 75\%); total UMI 90\%; D beyond UMI +0.1 points, UMI beyond D +5.9; thinning: one free scalar $\times$ one readout 97\% of the change, one-scalar white box $R^2$ 0.92 / 0.94; rank-target objective D 70\%, UMI 71\%, D--UMI correlation 0.95--0.99 (nine cohorts); pre-normalised cohort D 58\%, UMI $-$17\%; scFoundation UMI 46\%, D 35\%, cell type 49\% \\
\addlinespace[2pt]
2.4 & scGPT thinning & $R^2$ 0.47 / 0.56; 57--63\% of the response in the span of its gene readouts \\
\addlinespace[2pt]
2.4 & Log CP10k model with depth features & one cell-level scalar 95\% of gain; UMI 92\%, D 87\%; detected-gene feature vs value thinning: uncentred cosine 0.96 at half the length; additivity $R^2$ 0.998 \\
\addlinespace[2pt]
2.4 & Genes read depth & gene response vs level r $-$0.84 to $-$0.94 (controlled), $-$0.68 (scGPT); across objectives r 0.70--0.94; rank-target vs scGPT r 0.82 \\
\addlinespace[2pt]
2.5 & Cell embeddings & depth polynomial $R^2$: 54\% (rank-target objectives), 87\% (fixed-bin log CP10k targets); annotation macro-F1 0.64--0.66 vs binary-detection PCA 0.73; untrained 0.44 vs random projection 0.45; scGPT depth 17\%, annotation 0.81 vs detection projection 0.67 (five cohorts) \\
\addlinespace[2pt]
2.5 & Attribution & identity-control IG identity share 75--95\% across zero, mean and shuffled baselines; clamped-value control zero-baseline IG value share about 15\%; defined intervention-loss ratios and IG use different measurement targets; zero-denominator loss ratios are omitted; scGPT MLP value encoder: 160 of 512 knots in range, PC1+PC2 99.78\% (before layer normalisation), 99.97\% (after layer normalisation) \\
\addlinespace[2pt]
2.5 & Attention & single-head marginal AUROC 0.80; median pairwise residual 0.51--0.53; head-averaged 0.63--0.66; fixed random codes 0.66; co-expression 0.64; TRRUST degree sum 0.996 (715 pairs) \\
\addlinespace[2pt]
2.5 & Token deletion & nuisance $R^2$: untrained 93--96\% (mean pooling), 47\% ({[}CLS{]}); trained 27--49\%; $\lvert\rho\rvert$ with log frequency 0.37--0.68; synthetic TF AUROC 0.68 vs frequency 0.52 \\
\addlinespace[2pt]
2.5 & State patching and ablations & layer-5 probe $R^2$ 0.87 (0.51 at initialisation); probe-subspace swap 0.23\%, random subspace 0.53\%, full state 89\%; untrained learned-subspace transfer 172\%; post-hoc edge zeroing $-$111.58 points vs retrained mask $-$15.42 \\
\addlinespace[2pt]
2.5 & Dictionaries and gene tables & random ReLU directions 39.1\% vs SAE 9.0\% enriched (layer 3, controlled); <5\% atoms reproduce; gene table vs initialisation linear CKA 0.991--0.996; scGPT gene table GO-BP AUROC 0.595 vs PPMI-SVD 0.602; degree sum 0.914 \\
\addlinespace[2pt]
Discussion & Visible-input pretraining (three seeds per arm, identical masks) & MLP value encoder, rank-target objective, covariate pipeline: white box with a three-dimensional cell state $R^2$ 0.837 (original) vs 0.841 (visible-input); log detected-gene count share 69.5\% vs 69.9\%; log detected-gene count + cell type 83.8\% vs 84.1\%. Affine rank-target objective, thinning: white-box $R^2$ (K3) 0.777 vs 0.767 (one half), 0.814 vs 0.806 (one quarter); one cell-level scalar (K1) 0.660 vs 0.653, 0.681 vs 0.673 \\
\addlinespace[2pt]
Discussion, 2.4 & Visible-input pretraining, fixed-bin log CP10k model & one cell-level scalar 98.8\% vs 99.7\% of the gain from the three-dimensional cell state; log detected-gene count 77.1\% vs 81.1\%; log UMI 78.6\% vs 78.5\% (covariate pipeline); thinning white-box $R^2$ (one cell-level scalar) 0.92 vs 0.91 (one half), 0.94 vs 0.93 (one quarter) \\
\addlinespace[2pt]
2.2 & Two-feature encoder controlled retraining \& ablations & 50-bin masked MSE \& three-seed mean annotation macro-F1 (ensemble): MLP quantile 149.35, 0.657 (0.675); two-feature 146.39, 0.767 (0.786); mid-rank only 146.76, 0.757 (0.777); standardised log magnitude only 146.30, 0.771 (0.796); depth-leak control 147.21, 0.754 (0.778) \\
\addlinespace[2pt]
2.2 & Two-feature encoder graft onto pretrained scGPT (33 million pretraining cells) & Value encoder 264,704 $\rightarrow$ 2,560 parameters; layers 8--11 unfrozen; two epochs, 3,124 steps, donor-disjoint 200,000-cell corpus; annotation macro-F1 0.867 (two-feature encoder) vs 0.867 (original-encoder continuation); masked MSE 258.7 (fixed-bin log CP10k targets) vs 141.2 (two-feature encoder) \\
\addlinespace[2pt]
2.2 & Two-feature encoder scale & 12 layers, width 512, 1,000,000 training cells; annotation macro-F1 0.821 (two-feature encoder) vs 0.710 (MLP value encoder); masked MSE 143.25 vs 146.02, respectively \\
\addlinespace[2pt]
2.2 & Accessibility counterfactuals: EpiAgent classification token & Four-cohort mean annotation macro-F1: intact 0.462; binarised 0.455; non-zero values permuted 0.466; coordinates permuted 0.109 \\
\addlinespace[2pt]
2.2 & Accessibility counterfactuals: term frequency--inverse document frequency (TF-IDF), latent semantic indexing & Four-cohort mean annotation macro-F1: intact 0.245; binarised 0.242; non-zero values permuted 0.240; coordinates permuted 0.077 \\
\addlinespace[2pt]
2.2 & Accessibility count sparsity & 72--88\% of non-zero entries on called peaks equal one or two \\
\addlinespace[2pt]
2.5 & Controlled multiome: binary autoencoder & Depth $R^2$ 0.522; annotation macro-F1 0.093 \\
\addlinespace[2pt]
2.5 & Controlled multiome: autoencoder with linear value input & Depth $R^2$ 0.983; annotation macro-F1 0.060 \\
\addlinespace[2pt]
2.5 & Controlled multiome: Poisson autoencoder with depth adversary & Depth $R^2$ $-$0.042; annotation macro-F1 0.048 \\
\addlinespace[2pt]
2.5 & Controlled joint multiome & Annotation macro-F1 0.205 \\
\addlinespace[2pt]
\end{longtable}
\endgroup

\section*{Supplementary Table 5 | Analyses reported in the text and not plotted}

\begingroup
\footnotesize
\setlength{\tabcolsep}{4.5pt}
\renewcommand{\arraystretch}{1.18}
\begin{longtable}[]{@{}>{\raggedright\arraybackslash}p{(\textwidth - 6\tabcolsep) * \real{0.2383}}
  >{\raggedright\arraybackslash}p{(\textwidth - 6\tabcolsep) * \real{0.3538}}
  >{\raggedright\arraybackslash}p{(\textwidth - 6\tabcolsep) * \real{0.3978}}@{}}
\toprule
{\sffamily\bfseries Analysis} & {\sffamily\bfseries Protocol location} & {\sffamily\bfseries Scope} \\
\midrule
\endhead
\bottomrule
\endlastfoot
Readout-moment isolation & Methods, Two-feature value encoder & Common continuous target with visible-moment readout; constructive variants appear in Extended Data Fig. 2f. \\
\addlinespace[2pt]
Partial fine-tuning, annotation & Methods, Pretrained encoders, interventions and probes & Eight held-out donors from two cohorts; full labels and 10 labels per class. Fine-tuned clamped-minus-original donor-mean macro-F1 differences (points; stratified donor-bootstrap 95\% intervals): full labels, scGPT $-$0.613 {[}$-$1.901, +0.670{]}, scFoundation +1.229 {[}$-$1.011, +4.262{]}; 10 labels/class, scGPT $-$1.795 {[}$-$3.612, +0.185{]}, scFoundation $-$1.937 {[}$-$3.388, $-$0.824{]}. No equivalence test is claimed. Protein comparison: Extended Data Fig. 4c. \\
\addlinespace[2pt]
Inference-time edge deletion & Supplementary Note 3 & Panel and scGPT comparisons removed symmetric top-$k$ or random-$k$ off-diagonal edges per head and query, $k\in\{1,2,3,5,8,10\}$; random controls used 20 Monte Carlo draws per cell. Intervals used 95\% cell bootstraps from 2,000 resamples. \\
\addlinespace[2pt]
Integrated Gradients step sensitivity & Supplementary Note 3; Methods, Interpretability readouts and controls & 10--500 integration steps on 16 lung cells; 20 steps on 256 panel cells; zero-vector, corpus-mean and constant-zero-expression baselines. Completeness errors: panel 0.015--0.041; scGPT 0.0017--0.011 at 500 steps vs 1.3--1.66 at 10--20 zero-vector steps. \\
\addlinespace[2pt]
Constructive cis rescue & Controlled multiome analysis & Cis-graph with sequence gate improves within-cell-type correlation by +0.0014 over concatenation; zero-shot enhancer--promoter AUROC 0.539. \\
\addlinespace[2pt]
\end{longtable}
\endgroup

\end{document}